\documentclass[acmtocl,acmnow]{acmtrans2m}
\usepackage{amsmath,amssymb}

\newtheorem{theorem}{Theorem}[section]

\newtheorem{corollary}[theorem]{Corollary}
\newtheorem{proposition}[theorem]{Proposition}
\newtheorem{lemma}[theorem]{Lemma}
\newdef{definition}[theorem]{Definition}
\newdef{remark}[theorem]{Remark}

\newdef{example}[theorem]{Example}

\newcommand{\marc}[1]{{\sf (MARC) #1 (CRAM)}}

\newcommand{\ignore}[1]{} 

\newcommand{\Tr}{{\bf t}}

\newcommand{\Fa}{{\bf f}}

\newcommand{\la}{\leftarrow}

\newcommand{\ra}{\rightarrow}

\newcommand{\lra}{\leftrightarrow}

\newcommand{\rla}{\leftrightarrow}

\newcommand{\rul}{\leftarrow}

\newcommand{\sq}{\sqsubseteq}

\newcommand{\bigsqcap}{\sqcap} 

\newcommand{\sqd}{\sqsubseteq_{\rm d}}

\newcommand{\sqn}{\sqsubseteq_{\rm n}}

\newcommand{\eqd}{=_{\rm d}}

\newcommand{\eqn}{=_{\rm n}}

\newcommand{\natnr}{\mathbb{N}}  

\newcommand{\nat}{\mathbb{N}}  

\newcommand{\integers}{\mathbb{Z}}

\newcommand{\Pow}[1]{{\rm Pow}(#1)}

\newcommand{\Pa}{P[\bar{a}]}

\newcommand{\Qb}{Q[\bar{b}]}

\newcommand{\PP}{\bar{P}}

\newcommand{\xx}{\bar{x}}
\newcommand{\yy}{\bar{y}}
\newcommand{\posind}[1]{PID(#1)}

\newcommand{\G}{\Gamma}                 

\newcommand{\TPp}{{\sf T}}             

\newcommand{\lfp}{{\it lfp}}    %

\newcommand{\gfp}{{\it gfp}}    %

\newcommand{\lub}{{\it lub}}    %

\newcommand{\glb}{{\it glb}}    %

\newcommand{\x}{\bar{x}}                

\newcommand{\bx}{\bar{x}}                

\newcommand{\y}{\bar{y}}                

\newcommand{\X}{\bar{X}}                

\newcommand{\bt}{\bar{t}}               

\newcommand{\symb}{\sigma}        

\newcommand{\bsymb}{\bar{\sigma}}        

\newcommand{\Tra}[2]{{#1}_{#2}}

\newcommand{\lb}[2]{{#1}^{{#2}\downarrow}}                      

\newcommand{\ub}[2]{{#1}^{{#2}\uparrow}}                          

\newcommand{\D}{\Delta}

\newcommand{\defin}[1]{\left\{\begin{array}{l} #1 \end{array}\right\}}

\newcommand{\defp}[1]{\tau^{\rm d}_{#1}}

\newcommand{\tauI}[1]{\tau_{#1}}

\newcommand{\tdef}[1]{\tau_{#1}^{\rm d}}

\newcommand{\td}{\tau_{\D}^{\rm d}}

\newcommand{\tdD}{\tau_{\D}^{\rm d}}

\newcommand{\tdi}{\tau_{\D_i}^{\rm d}}

\newcommand{\tdj}{\tau_{\D_j}^{\rm d}}

\newcommand{\toD}{{\tau_{\D}^{\rm o}}}

\newcommand{\toi}{{\tau_{\D_i}^{\rm o}}}

\newcommand{\topen}[1]{{\tau_{#1}^{\rm o}}}

\newcommand{\free}[1]{free(#1)}

\newcommand{\taufun}{\tau_{\rm fn}}
\newcommand{\tfun}{\tau_{\rm fn}}

\newcommand{\tauo}{{\tau^{\rm o}}}

\newcommand{\Att}{\At{A}{\tau} }

\newcommand{\At}[2]{At_{#1}^{#2}}  

\newcommand{\valI}{{I}}             

\newcommand{\domI}{dom(I)}          

\newcommand{\dom}[1]{dom(#1)}          

\newcommand{\valIopen}{{I_{\rm o}}}             

\newcommand{\valJopen}{{J_{\rm o}}} 

\newcommand{\baseIo}{{I_{\rm o}}}      

\newcommand{\baseJo}{{J_{\rm o}}}      

\newcommand{\baseI}{{K_{\rm o}}}      

\newcommand{\Moi}{{M_{{\rm o}i}}}

\newcommand{\I}{{I}}

\newcommand{\J}{{J}}

\newcommand{\ba}{\bar{a}}               

\newcommand{\bd}{\bar{d}}               

\newcommand{\R}{\bar{R}}                

\newcommand{\restr}[2]{#1|_{#2}}        

\newcommand{\eval}[2]{#1^{#2}}          

\newcommand{\bv}{\bar{v}}               

\newcommand{\vallat}[2]{{\cal S}_{#1}^{#2}}    

\newcommand{\leqt}{\sqsubseteq}

\title{A Logic of  Non-Monotone Inductive Definitions}

\author{MARC DENECKER \\ K.U.Leuven, Belgium  \\
EUGENIA TERNOVSKA \\ Simon Fraser University, Canada}

\begin{abstract}

  Well-known principles of induction include monotone induction and
  different sorts of non-monotone induction such as inflationary
  induction, induction over well-founded sets and iterated induction.
  In this work, we define a logic formalizing induction over
  well-founded sets and monotone and iterated induction.  Just as the
  principle of positive induction has been formalized in FO(LFP), and
  the principle of inflationary induction has been formalized in
  FO(IFP), this paper formalizes the principle of iterated induction
  in a new logic for Non-Monotone Inductive Definitions (ID-logic).
  The semantics of the logic is strongly influenced by the
  well-founded semantics of logic programming.
  
  Our main result concerns the modularity properties of inductive
  definitions in ID-logic.  Specifically, we formulate conditions
  under which a simultaneous definition $\D$ of several relations is
  logically equivalent to a conjunction of smaller definitions $\D_1
  \land \dots \land \D_n$ with disjoint sets of defined predicates.
  The difficulty of the result comes from the fact that predicates
  $P_i$ and $P_j$ defined in $\D_i$ and $\D_j$, respectively, may be
  mutually connected by \ignore{non-monotone} simultaneous induction.
  Since logic programming and abductive logic programming under
  well-founded semantics are proper fragments of our logic, our
  modularity results are applicable there as well.  \ignore{As an example of
  application of our logic and theorems presented in this paper, we
  describe a \ignore{ non-monotonic reasoning} temporal formalism, the
  inductive situation calculus, where causal dependencies are
  naturally represented as rules of inductive definitions.}

\end{abstract}

\category{...}{...}{...}
            
\begin{document}
            
\begin{bottomstuff} 
Author's address: M.Denecker, Department of Computer Science,
K.U.Leuven, elestijnenlaan 200A, B-3001 Heverlee, Belgium.
Phone: +32 16 327544 --- Fax:  +32 16 327996.
email: marcd@cs.kuleuven.ac.be\newline
Author's address: 
\end{bottomstuff}
            
\maketitle

\section{Introduction}
\label{Introduction}

This paper fits into a broad project aiming at studying general forms
of inductive definitions and their role in diverse fields of
mathematics and computer science. Monotone inductive definitions and
inductive definability have been studied extensively in mathematical
logic \cite{Moschovakis74,Aczel77}. The algebraic foundations for
monotone induction are laid by Tarski's fixpoint theory of monotone
lattice operators \cite{Tarski55}. The notion of inductive definition
is the underlying concept in fixpoint logics 
\cite{Gurevich86,Dawar/Gurevich:2002}
which found its applications in  e.g. database theory \cite{Abiteboul95}
and descriptive complexity theory \cite{Immerman:1999,Ebbinghaus/Flum:1999}. Logics with
fixpoint constructs to represent (monotone) inductive and
co-inductive definitions play a central role as query and
specification languages in the area of verification of dynamic systems
using  modal temporal loics such as the $\mu$-calculus \cite{Kozen:1983}.
Induction axioms have
been used succesfully in the context of proving properties of
protocols using specialised automated reasoning tools
\cite{Paulson98}. The concept of definitions and definitional
knowledge is also fundamental in the area of description logics
\cite{Brachman82}, the class of logics that evolved out of
semantic networks. Importantly, complexity results in fixpoint logic
and logic programming suggest that inductive definitions often combine
high expressivity with low complexity. Thus, it
appears that the notion of definition and its inductive
generalisations emerges as a unifying theme in many areas of
mathematics and computational logic. Hence, its study could improve
insight in the interrelations between these areas and lead to synergy
between them.

In this paper, we are concerned with non-monotone inductive
definitions.  A familiar example of a non-monotone inductive
definition is the definition of the satisfaction relation $\models$
between a truth assignment $I$ and a formula.  In case of
propositional logic, this relation is defined by induction over the
subformula order on formulas:
\begin{itemize}
\item [ - ] $\valI \models p$ if $p\in \valI$,
\item [ - ]  $\valI \models \psi \land \phi$ if $\valI\models \psi$ and
  $\valI\models \phi$,
\item [ - ]  $\valI \models \psi \lor \phi$ if $\valI\models \psi$ or
  $\valI\models \phi$,
\item  [ - ] $\valI \models \neg \psi$ if $\valI \not\models \psi$.
\end{itemize}
This inductive definition is non-monotone because of its last rule,
which adds the pair $(\valI,\neg \psi)$ to the truth relation if the
pair $(\valI,\psi)$ does not belong to it. This is an example of an
inductive definition over a well-founded order. Recently, the authors
of \cite{Denecker98c,Denecker2001:TOCL} investigated certain
non-monotone forms of inductive definitions in mathematics and pointed
out that semantical studies in the area of logic programming might
contribute to a better understanding of such generalised forms of
induction. In particular, it was argued that the well-founded
semantics of logic programming \cite{VanGelder91} extends monotone
induction and formalises and generalises non-monotone forms of
induction such as induction over well-founded sets and iterated
induction
\cite{Feferman70,Feferman81}. In \cite{Denecker2000a,Denecker:IC2004}, 
the well-founded semantics was further generalised into a fixpoint
theory of general non-monotone lattice operators.  This theory, called
{\em approximation theory}, generalises Tarski's theory of fixpoints
of monotone lattice operators and provides the algebraic foundation of
the principle of iterated induction. Later, it turned out that the same
principle is fundamental in an area of artificial intelligence
concerned with using logic for knowledge representation ---
non-monotonic reasoning\footnote{The term ``non-monotone'' has a
different meaning in the context of inductive definitions than in the
context of non-monotone reasoning.  A logic is non-monotone when
adding formulas to a theory may not preserve inferred formulas. A
monotone definition is one inducing a monotone operator. In fact, the
fragment of monotone inductive definitions in ID-logic is a
non-monotone logic. }. In particular,
\cite{Denecker2000a,Denecker:AIJ2002} demonstrated that the semantics
of three major approaches to non-monotonic reasoning, default logic
\cite{Reiter80b}, autoepistemic logic \cite{Moore85} and logic
programming \cite{Lloyd87} are described by approximation theory.
Thus, generalised  inductive definitions also play a
fundamental role in the semantics of knowledge representation
formalisms.

In a seminal paper on knowledge representation, Brachman and Levesque
\cite{Brachman82} had observed that definitional knowledge is an
important component of human expert knowledge. Motivated by this work,
the author of \cite{Denecker:CL2000} extended classical logic with
non-monotone inductive definitions in order to demonstrate that
general non-monotone inductive forms of definitions also play an
important role in knowledge representation.  In this paper, we extend
this work. The contributions of the paper are the following:

\begin{itemize}
\item  We formalize the principle of iterated induction in a new logic
for Non-Monotone Inductive Definitions (ID-logic). This logic is an
extension of classical logic with non-monotone inductive definitions,
and is a generalisation of the logic that was defined in
\cite{Denecker:CL2000}. 

\item We demonstrate that different classes  of definitions can
be correctly and uniformally formalised in our logic.  To achieve this
goal, we present an alternative formalisation of these classes in
classical first- or second-order logic, and provide an
equivalence-preserving transformation from ID-logic to these
formalisations.

\item  We  study modularity properties of non-monotone 
inductive definitions in ID-logic and provide a set of techniques
that allow one to break up a big definition into a conjunction of
smaller and simpler definitions.

\ignore{
\item We illustrate the use of iterated inductive definitions 
in the context of a knowledge representation problem. In particular,
we formalise a temporal reasoning problem with recursive effect
propagations using an iterated inductive definition in ID-logic. We
show how the modularity theorem and the transformation techniques can
be used to analyse this definition and translate it to classical
logic.}

\end{itemize}

The main result of the paper is a set of formal conditions that
 guarantee that a simultaneous definition of several predicates can be
 split up into the conjunction of components of this definition, each
 component defining some subset of the defined predicates. In
 addition, our theorems provide conditions under which joining a set
 of definitions for distinct sets of predicates into one simultaneous
 definition of all these predicates is equivalence preserving. The
 problem that we study is similar to that studied in
\cite{Verbaeten99a}, but our results are uniformally stronger
in the sense that they are proven for a more expressive logic and
under more general conditions.

The results are important because modularity is a crucial property in
formal verification and knowledge representation
\cite{Reichgelt91}. For example, it is important to be able
 to specify a complex synamic system by describing its components in
independent modules which can then be conjoined to form a correct
description of the complete system. Thus, the operation of joining
modules should preserve the correctness of the component modules.  The
dual operation of {\em splitting} a complex theory into an equivalent
set of smaller modules is equally important. It allows one to investigate
complex theories by studying its modules independently, and reduces
the analysis of the correctness of the complex theory to the much
simpler problem of analysing the correctness of its modules.

The paper is structured as follows. In Section
\ref{Sec:forms-of-induction}, we discuss various forms of induction
and their formalisations.  This discussion provides the intuitions and
the motivation for defining the new logic.  Section
\ref{Sec:Preliminaries} introduces some preliminaries from logic and
lattice theory.  In section \ref{Sec:ID-logic}, we extend classical
logic with the generalised non-monotone definitions. In section
\ref{Sec:modularity}, the modularity of the definition expressions is
investigated. In section \ref{Sec:familiar-types}, we present
equivalence-preserving transformations from ID-logic to first-order
and second-order logic for different familiar types of
definitions. Here, the modularity techniques developed in the previous
section are used as a tool to prove correctness of the
transformations.

\newcommand{\inference}[2]{\frac{#1}{#2}}
\newcommand{\proves}{\vdash}

\section{Formal Study of Inductive Definitions}
\label{Sec:forms-of-induction}

{\em Mathematical induction} refers to a class of effective
construction techniques used in mathematics. There, a set is
frequently defined as the limit of a process of iterating some \ignore{ basic
extension} operation. Often, mathematicians describe such a
construction by an inductive definition. The core of an
inductive definition in mathematics consists of one or more basic
rules and a set of inductive rules.  Basic rules represent base cases
of the induction and add elements to the defined set in an
unconditional way; inductive rules add new elements to the set if one
can establish the presence or the absence of other elements in the
set. The defined set is obtained as the limit of some process of
iterated application of these rules. 

In this section, we discuss various forms of such inductive
definitions and how they are formalised. Then we motivate and preview
a new formal logic of definitions. The section is partially based on
ideas presented earlier in \cite{Denecker98c,Denecker2001:TOCL}.

\vspace{2mm}

\subsection{Monotone Inductive Definitions.} In a monotone inductive
definition, the presence of new elements in the set depends only on
the presence of other elements in the defined set, not on the absence
of those.  The defined set is the least set closed under application
of the rules.  Such definitions are frequent in mathematics.  A
standard example is the transitive closure of a directed graph:
\begin{quote}
The transitive closure $T_G$ of a directed graph $G$ is inductively
defined as the set of edges $(x,y)$ satisfying the following rules:
\begin{itemize}
\item $(x,y) \in T_G$ if $(x,y) \in G$;
\item $(x,y) \in T_G$ if for some vertex $z$, $(x,z), (z,y) \in T_G$. 
\end{itemize}
\end{quote} 
Other typical examples are the the definition of a subgroup generated
by a set of group elements, or the definitions of a term, formula,
etc. in logic.

\ignore{ {\tt BELOW:  Neither Moschovakis
nor Aczel define a logic. There is no formal syntax or semantics.
 They study IDs from a mathematical point of view, even though they
 use notations from logic. Your text sounded like there is a logic.  I
 changed it a tiny bit so that the reader does not think that there is
 a logic.} Marc: Ok} Monotone inductive definitions have been studied
 extensively in mathematics \cite{Moschovakis74,Aczel77}.  In
\cite{Moschovakis74}, such a definition is
associated with a formula $\varphi(\x,X)$.  Intuitively, this formula
encodes all the conditions under which tuple $\x$ belongs to the
defined predicate $X$.  The formula $\varphi(\x,X)$ must be positive
in $X$, that is no occurrence of $X$ may appear in the scope of an odd
number of occurrences of the negation symbol $\neg$.  For instance,
for the transitive closure example above we have:
$$
 \varphi_{trans}((x,y),T_G) := G(x,y) \lor \exists z (T_G(x,z) \land T_G(z,y)).
$$
Each disjunct in this formula formally expresses the condition of one of the 
rules in the informal definition.

\ignore{{\tt BELOW: going too fast?  iF THIS OPERATOR WILL BE USED AGAIN,
MAYBE WE SHOULD DEFINE IT MORE FORMALLY???}}
Given a structure $I$ which  interprets all constant symbols, the formula
$\varphi(\x,X)$ characterises an operator $\G_{\varphi(\x,X)}$ mapping
a relation $R$ to the relation $R'$ consisting of tuples $\ba$ such
that $\varphi(\ba,R)$ is true in $I$. In general, $\G_{\varphi(\x,X)}$
may have multiple fixpoints, but the fact that $\varphi(\x,X)$ is a
positive formula implies that the operator is monotone and has a least
fixpoint, which is the relation inductively defined by
$\varphi(\x,X)$.  A logic to represent monotone inductive definitions
is the least fixpoint logic FO(LFP) (see, e.g. \cite{Ebbinghaus/Flum:1999}).
\ignore{{\tt SHOULD WE GIVE EXAMPLE HOW TRANSITIVE CLOSURE IS REPRESENTED
IN FO(LFP) ???? } Marc: no}

\subsection{\ignore{Non-Monotone} Inductive Definitions over a Well-Founded Order.} 

\ignore{{\tt I ADDED NON-MON TO THE TITLE} Marc: inductive definitions 
over a wfo are not necessarily nonmon. moreover, the title got too long, 
so i deleted non-mon again}
In a non-monotone inductive definition, 
 the presence of new elements in the set  depends on
the absence of certain elements in the defined set.  
An example of such a definition is the definition of the truth relation 
given in the introduction.
Let us consider another definition with a similar structure.
\begin{quote} The set of even numbers is defined by induction over 
the standard order $\leq$ on the natural numbers:
\begin{itemize}
\item  0 is an even number; 
\item  $n+1$ is an even number if $n$ is not an even number. 
\end{itemize}
\end{quote}
\ignore{{\tt BELOW: NEED TO EDIT TO HAVE A LOGICAL ORDER}}
The definitions of even numbers and of $\models$ are examples of
inductive definitions over well-founded orders.  Such a definition
describes the membership of an element in the defined relation in
terms of the presence or absence of elements in the defined relation
that are strictly smaller with respect to some well-founded
(pre-)order. By applying this definition to the minimal elements and
then iterating it for higher levels, the defined predicate can be
constructed, even if some inductive rules are non-monotone. 
\ignore{ {\tt I PROPOSE TO FINISH THE DISCUSSION ABOUT 
EVEN EXAMPLE HERE, I.E., TO MOVE THE STUFF FROM BELOW HERE, AND TO
MOVE THE NEXT PARAGRAPH TO THE END OF THIS SUBSECTION}: Marc: done}
This type of inductive definitions is fundamentally different from
monotone inductive definitions. Indeed, the set defined by a monotone
inductive definition can be characterised as the least set closed
under the rules. In contrast, a definition over a well-founded order
does not characterise a unique least set closed under its rules.
For instance, $\{0,2,4,6,\dots\}$ and $\{0,1,3,5,7,\dots\}$ are both
minimal sets closed under the above rules.

Using the same representation methodology to represent this inductive
definition as in the monotonic case, we would obtain the formula
\[ \varphi_{\rm even}(x,E) : = x=0 \lor \exists y (x=S(y) \land \neg
E(y)).\] 
In the context of the natural numbers, the operator characterised by
this formula is non-monotone and maps any set $S$ of natural numbers
to the set consisting of 0 and all successors of all numbers in the
complement of $S$. This is a non-monotone operator which has the set
of even numbers as unique fixpoint. In general, the set defined by
this type of inductive definitions can be characterised as the unique
fixpoint of the operator associated to the definition. This will be
formalised in section \ref{Sec:familiar-types}.
\ignore{{\tt THE POINT IS NOT CLEAR HERE. 
$\{0,2,4,6,\dots\}$ and
$\{0,1,3,5,7,\dots\}$ are both minimal sets closed under the rules,
but the corresponding operator has a unique fixpoint.
WHAT ARE WE REALLY SAYING HERE?} Marc: I've done my best. }

Other examples of non-monotone inductive definitions over well-founded
orders are given in \cite{Denecker2001:TOCL}.  They include a
definition of the concept of a rank of an element in a well-founded
set (the rank of an element $x$ is the least ordinal strictly larger
than the rank of all $y<x$), and a definition of the levels of a
monotone operator in the least fixpoint construction.  Although
induction over a well-founded set is a common principle in
mathematics, to our knowledge it has not been studied explicitly in
mathematical logic. However, we will argue below that it can be seen
as a simple form of iterated induction.

\subsection{ Inflationary Induction.} 
 In order to extend his theory
of inductive definitions to the class of all definitions (monotone and
non-monotone), Moschovakis \cite{Moschovakis74a} proposed the
following approach. The idea is to associate with an arbitrary formula
$\varphi(\x,X)$ (possibly non-positive) the operator
$\G'_{\varphi(\x,X)}$, where
$$
\G'_{\varphi(\x,X)}(R):=\G_{\varphi(\x,X)}(R)\cup R.
$$
Operator $\G'_{\varphi(\x,X)}$ is not monotone, but it is {\em
inflationary}, that is, for every $R$, $R\subseteq
\G'_{\varphi(\x,X)}(R)$. Thus, by iterating this operator starting at
the empty relation, an ascending sequence can be constructed. This
sequence eventually reaches a fixpoint of $\G'_{\varphi(\x,X)}$.  This
fixpoint was later called the {\em inflationary fixpoint}, and the
corresponding logic FO(IFP) was introduced \cite{Gurevich86}.  This
logic introduces inflationary, and its dual, deflationary, fixpoint
constructs.  The inflationary fixpoint logic played an important role in
descriptive complexity theory and has been used to characterize the
complexity class PTIME \cite{Immerman:1986,Livchak:1983,Vardi:1982}.

\ignore{{\tt NEW VETSION. I KEPT THE OLD ONE BELOW IN FOOTNOTE SIZE} Marc: well-done! Made some further modifications.}

Inflationary induction and induction over a well-founded order are two
different principles. Consider, for example, the definition of the
even numbers presented above.  The formula $\varphi_{\rm even}$ is a
natural representation of this definition.  However, the inflationary
fixpoint $[{\rm IFP}_{\bx,E}\varphi_{\rm even}]\bar t$ is the set of
all natural numbers.  Indeed, $\emptyset\cup\G_{\Delta_{\rm
even}}(\emptyset)=\natnr$ and $\natnr\cup\G_{\Delta_{\rm
even}}(\natnr)=\natnr$.  Even though it is possible to write down a
definition of the even numbers using inflationary fixpoints, such an
encoding would be neither natural nor direct. It would not reflect the
way in which mathematicians express induction over a well-founded
order.  Since our goal is to formalise the latter sort of induction in
a way that reflects the natural rule-based structure in which
mathematicians represent such definitions, \ignore{ Since our goal is
to capture the natural information propagation reflected in inductive
definitions, -> Marc: what propagation, what inductive rules what kind
of induction??} this paper will not be concerned with inflationary
fixpoints.  For examples where inflationary and deflationary
inductions naturally appear, we address the reader to the work by
Gr\"{a}del and Kreutzer
\cite{Graedel/Kreutzer:2003}.

\ignore{

\begin{footnotesize}
Inflationary induction and induction over a well-founded order are
different principles. For example, the inflationary fixpoint of the
definition $\varphi_{\rm even}$ is the set of all natural
numbers. Indeed, $\emptyset\cup\G_{\Delta_{\rm
even}}(\emptyset)=\natnr$ and $\natnr\cup\G_{\Delta_{\rm
even}}(\natnr)=\natnr$. {\tt THIS IS A JUMP. IT IS TOO QUICK. }
 For this reason, this paper will not be
concerned with inflationary fixpoints. {\tt BELOW: NEEDS EDITING.
IT'S AN IMPORTANT POINT.} 
We not claim by the way that
inductive definitions over well-founded orders cannot be expressed in
FO(IFP), but such an encoding will not reflect the way in which
mathematicians express induction over well-founded order. Gr\"{a}del
and Kreutzer \cite{Graedel/Kreutzer:2003} discuss several examples
where inflationary and deflationary inductions naturally appear.
\end{footnotesize}}

\subsection {Iterated Inductive Induction.} \label{SSec:IID}
The basic idea underlying induction is to iterate a basic construction
step until a fixpoint is reached. In an iterated induction, this basic
construction step itself is a monotone induction. That is, an iterated
inductive definition constructs an object as the limit of a sequence
of constructive steps, each of which itself is a monotone induction.
One can formulate the intuition of the iterated induction of a
structure also in the following way.  Given a mathematical structure
$M_0$ of functions and relations, a positive or monotone inductive
definition defines one or more new relations in terms of $M_0$. The
definition of these new relations may depend positively or negatively
on the relations given in $M_0$. Once the interpretation of the new
relations is fixed, $M_0$ can be extended with these, yielding a new
extended structure $M_1$.  On top of this structure, again new
relations may be defined in the similar way as before. The definition
of these new predicates may now depend positively or negatively on the
relations that were defined in $M_1$.  This modular principle can be
iterated arbitrarily often, possibly a transfinite number of times. We
call this informal principle the {\em principle of Iterated
Induction}.  In general, an iterated inductive definition must
describe, in a finite way, a possibly transfinite sequence of monotone
or positive definitions of sets. If the definition of a set depends
(positively or negatively) on another defined set, then this other set
must be defined in an earlier definition in this sequence.

An example of an iterated inductive definition mentioned in
\cite{Denecker2001:TOCL} is the definition of the {\em stable theory}
of some propositional theory $T$. Basically, this is the standard
concept of {\em deductive closure} of a propositional theory $T$ under
a standard set of inference rules augmented with two additional
inference rules:
$$
\inference{\proves \psi}{ \proves K\psi}\mbox{\ \ \ and \ \ \ } 
\inference{\not \proves \psi}{ \proves \neg K\psi} .
$$
Note that the second rule is non-monotone. The stable theory of $T$ is
a deductively closed modal theory which contains explicit formulas
representing whether $T$ ``knows'' a formula $\psi$ or not. It can be
viewed as the set of formulas known by an ideally rational agent with
perfect introspection whose base beliefs are represented  by $T$.

Let us consider this induction process in more detail. We define the
modal nesting depth of a formula $F$ as the length $n$ of the longest
sequence $(KF_1, KF_2,\dots , KF_n)$ such that $F$ contains $KF_1$ and
$F_i$ contains $KF_{i+1}$ for each $1\leq i <n$.  The start of the
iterated induction is a monotone induction closing $T$ under the
propositional logic inference rules.  This yields a deductively closed
set $T_0$ of propositional formulas of modal nesting depth $0$. Next
we apply the two modal inference rules to infer modal literals $K\psi$
or $\neg K\psi$, for each propositional formula $\psi$.  After
computing these literals, we reapply the first step and derive, using
the standard inference rules, all logical consequences with modal
nesting depth being less or equal to 1. This process can now be
iterated for formulas with increasing modal nesting depth.  The result
of this construction process is the stable theory of $T$ and contains
formulas of arbitrary modal nesting depth. It was shown in
\cite{Marek89b} that the stable theory of $T$ is exactly the
collection of all modal formulas that are true in the possible world
set $W$ consisting of all models of $T$. More precisely, it holds that
the stable theory of $T$ is the set of all modal formulas $F$ such
that for the collection $W$ of models of $T$ and for each model $M\in
W$, it holds that $W,M \models F$.

Iterated Induction is a generalisation of monotone induction. It is
also related to induction over a well-founded order. The link is seen
if we split up a definition of the latter kind in an infinite number
of definitions, each defining a single ground atom, and ordering or
{\em stratifying}\footnote{This stratification corresponds to the
notion of {\em local stratification} in logic programming
\cite{Przymusinski88}.}  these definitions in a sequence compatible
with the well-ordering. For example, even numbers could be defined by
the following iterated definition:
\[ \begin{array}{lrl}
 (0)    &       0  \mbox{ is even} & \\
 (1)    &      1  \mbox{ is even} &  \mbox{if 0 is not even}\\
 (2)    &      2   \mbox{ is even} &  \mbox{if 1 is not even}\\
 & ... &  \\
(n+1)  &       n+1  \mbox{ is even} &  \mbox{if $n$ is not even}\\
   & ... & 
 \end{array}
\]
Clearly, the iterated induction described here constructs the set of
even numbers.  We can thus view an inductive definition in a
well-founded set as an iterated inductive definition consisting of a
sequence of non-inductive (recursion-free) definitions. \ignore{ This
is a common aspect of definitions in a well-founded set because
whether some element belongs to the defined set or not, depends only
on whether strictly smaller elements belong to the set.} Iterated
induction is more general than induction over a well-founded set because
positive recursion within one level may be involved (as illustrated by
the stable theory example).  

The logical study of iterated induction was started in
\cite{Kreisel63} and extended in later studies of so-called Iterated
Inductive Definitions (IID) in \cite{Feferman70}, \cite{MartinLoef71},
and \cite{Feferman81}. The IID formalism defined in
\cite{Feferman70,Feferman81} is a formalism to define sets of natural
numbers through iterated induction. To represent an iterated inductive
definition of a set $H$, one associates with each natural number an
appropriate {\em level index}, an ordinal number. This level index can
be understood as the index of the subdefinition which determines
whether the number belongs to the defined set or not.  The iterated
inductive definition is described by a finite parametrised formula
$\varphi(n,x,P,H)$, where $n$ represents a level index, $x$ is a
natural number, $P$ is a unary predicate variable with only positive
occurrences in $\varphi$ and ranging over natural numbers, and $H$ is
the defined relation represented as a binary predicate ranging over
tuples $(n,x)$ of natural numbers $x$ and their level indices $n$.
The formula $\varphi(n,x,P,H)$ encodes that $n$ is the level index of
$x$, and $x$ can be derived (using the inductive definition with level
index $n$) from the set $P$ and the restriction of $H$ to tuples with
level index $<n$. Using $\varphi$, the set $H$ is characterised by two
axioms. The first one expresses that $H$ is closed under $\varphi$:
$$
\forall n \forall x\ (\varphi(P(\sigma)/H(n,\sigma)) \ra H(n,x)).
$$
In this formula, $\varphi(P(\sigma)/H(n,\sigma))$ (where $\sigma$ is
an arbitrary term) denotes the formula obtained from $\varphi$ by
substituting $H(n,\sigma)$ for each expression $P(\sigma)$.

\ignore{
\vspace{3mm} {\tt THIS EXAMPLE IS NOT CLEAR. Formula $\phi$ has 4
arguments, it depends both on P and on H. In the sentence above, H is
substituted for P.  Did the original formula depend on on H, or on P
only????  THIS NEEDS TO BE FIXED. AS IS, IT DOES NOT ILLUSTRATE THE
POINT BECAUSE IT'S NOT CLEAR WHAT'S GOING ON.  \marc{It is as I wrote
it and as you understood it. Yes, as mentioned explicitly, the
original formula contains both P and H. And Yes, as mentioned
explicitly, in the transformed formula, all occurrences of P are
substituted by occurrences of H.  So, consequently, the resulting
formula contains only H, not P.  } } }

The second axiom is a second-order axiom expressing that for each $n$,
the subset $\{x\ | \ (n,x) \in H\}$ of $\mathbb{N}$ is the least set
of natural numbers closed under $\varphi$:
$$
\forall n \forall P\ [\forall x\ (\varphi \ra P(x)) \ra \forall
x \ ( H(n,x)\ra P(x))].
$$
As an example, let us encode the non-monotone definition of even
numbers in the IID-formalism. It is a definition by induction on the
standard order of natural numbers which means that we can take a
natural number and its level index to be identical. The formula
$\varphi$ to be inserted in the axioms above is\footnote{ The formula
doesn't contain the predicate variable $P$ because this is a
definition over a well-founded order which does not involve monotone
induction.}:
$$
(n=0\land x=0) \lor
\exists y (n=s(y) \land x=s(y) \land \neg H(y,y)\land y<n).
$$
This formula represents that $(n,x)$ can be derived if $x$ and its
level index $n$ are identical and if $x=0$ or if the
predecessor of $x$ is not even.

\ignore{
The IID-formalism was defined to investigate proof-theoretic
properties of iterated induction. It was never intended to be used,
and it is not adequate for specification and knowledge representation.
Similar to the studies of monotone induction prior to Moschovakis'
landmark work \cite{Moschovakis74}, the theory of IIDs concentrates on
iterated inductive definitions of one set (one unary relation) in the
context of natural numbers.  The encoding of the induction involves a
tedious encoding of the segment of ordinal numbers corresponding to
the level indices by some recursive total well-founded order on the
natural numbers. 
\ignore{, and a G\"{o}del numbering of tuples of natural
numbers\footnote{This G\"{o}del numbering of tuples is necessary only
when defining a n-ary relation, $n>1$. In the example, a unary
relation is defined and G\"{o}del numbering is not needed.}  {\tt FROM
THIS EXAMPLE, OR THE EXPLANATION OF THE IID ABOVE, IT IS NOT CLEAR WHY
GOEDEL NUMBERING IS NEEDED.  WHERE DOES IT APPEAR IN THE EXAMPE?????
} }
Moreover, it forces one to explicitly define and encode a suitable
level mapping in the formula $\varphi$. }

\subsection{A Preview of ID-Logic} \label{ssec-preview-ID}

In this paper, we design a logic for formalising several forms of
inductive definitions.  Just as the principle of Monotone Induction
has been formalised in FO(LFP), the principle of Inflationary
Induction has been formalized in FO(IFP), the principles of Induction
over a well-founded order and Iterated Induction are captured by our
logic.  We call it a Logic for Non-Monotone Inductive Definitions
(ID-logic).  

The logic is designed as an extention of classical logic with
definitions. A definition will be represented as a set of rules of the
form:
\[ \forall \x ( P(\x) \rul \psi), \]
where $P$ is a relational symbol defined by the definition, and $\psi$
an arbitrary first-order formula.  For example, the non-monotone
definition of even numbers will be represented by the set:
\[ \defin{   \forall x      (E(x)\rul  x=0),\\
\forall x ( E(s(x)) \rul \neg E(x))}.\] 
From a representational point of view, this syntax has some
interesting features:
\begin{itemize}
\item {\bf \ignore{Modular} Rule-based representation.} 
Formalisations of definitions in ID-logic preserve the rule-based
structure of definitions in mathematics. Stated differently, rules in
a mathematical definition can be formalised in a modular way by
definitional rules in an our logic.
\ignore{ {\tt THE TERM "MODULAR" MIGHT BE CONFUSING HERE. THE READER
MIGHT THINK THAT THIS IS WHAT WE MEAN BY MODULARITY.  DO YOU ARGEE?
Marc: Partially, see below. I think this form of modularity and the
one studied in the theorems are indeed related.  MAYBE WE SHOULD TALK
ABOUT RULE-BASED REPRESENTATION AND OMIT "MODULAR"? Marc: I deleted
``modular'' in the title, but kept it in the explanation since I want
to stress the modularity of rules.  }}

\item {\bf Uniform formalisation of different types of definitions.}
Syntax and semantics of ID-logic is designed for uniform
formalisation of non-inductive (recursion-free) definitions, positive
or monotone inductive definitions, definitions over well-founded sets
and iterated inductive definitions.

\item {\bf No explicit level mapping.} 
A model of an ID-logic definition is constructed following the
natural dependency order on defined atoms that is induced by the
rules. As a consequence, and contrary to the IID-formalism of the
previous section, there is no need to explicitly represent a level
mapping of an iterated inductive definition. \ignore{ {\tt
SHOULD WE SAY THAT THE NATURAL DEPENDENCY ORDER GIVEN BY THE RULES IS
CAPTURED BY THE SEMANTICS WE DESCRIBE?  WE DON'T HAVE TO SPECIFY THE
REDUCTION ORDER IN ORDER TO DEFINE A MODEL, IT IS DONE "AUTOMATICALLY"
BY THE WELL-FOUNDED SEMANTICS. DOES IT MAKE SENSE TO TALK ABOUT IT
HERE????  Marc: Yes, I agree.  }}

\item {\bf Simultaneous induction.} Consider for example the following
simultaneous inductive definition of even and odd numbers:
\[ \defin{   \forall x      (E(x)\rul  x=0),\\
\forall x ( E(s(x)) \rul  O(x)),\\ 
\forall x ( O(s(x)) \rul  E(x))}.\]

\item {\bf A  logic with second-order variables.}
ID-logic allows second-order variables and quantification. As an
example, consider the following sentence of ID-logic:
\[ \exists P ( \defin{ \forall x (P(x)\rul x=0),\\ 
\forall x ( P(s(x)) \rul  P(x))} \land \forall x \ P(x)).\]
This axiom, stating that the least set $P$ containing $0$ and closed
under the successor operation contains all domain elements, is an
ID-logic formalisation of the second-order induction axiom of the
natural numbers.

\ignore{ {\tt SHOULD WE ADD THIS ITEM AND GIVE SOME EXAMPLES? Marc: Done.}}

\end{itemize}

The main differences between ID-logic and the IID-formalism of Section
\ref{SSec:IID} are  its  rule-based nature and the absence 
of an explicit encoding of a level mapping. 
The rules of an inductive definition induce an implicit dependency
order on the defined atoms.  For example, in the definition of even
numbers, the rule $\forall x (E(s(x)) \rul \neg E(x))$ induces a
dependency of each atom $E[n+1]$ on the atom $E[n]$. Notice that the
transitive closure of this dependency relation corresponds with the
standard order of the natural numbers, the well-founded order over
which the set of even numbers is defined by this definition. This
suggests that the encoding of the level mapping in the IID-formalism
only adds {\em redundant} information to the definition.

In ID-logic, 
the construction of the model of a definition proceeds by following
the implicit dependency order that is induced by the rules. The
technique to do this was developed in logic programming.  In
\cite{Denecker98c,Denecker2001:TOCL}, Denecker proposed the thesis
that the well-founded semantics of logic programming
\cite{VanGelder91} provides a general and robust formalization of the
principle of iterated induction.  In Section \ref{Sec:ID-logic}, we recall
this argument and show how the construction of the well-founded model
can be seen as an iterated induction which follows the natural
dependency order induced by the rules.

\newcommand{\congr}{{\cong}}

\section{Preliminaries}

\subsection{Preliminaries from Logic}\label{Sec:Preliminaries}

We begin by fixing notation and terminology for the basic syntactic
and semantic notions related to first- and second-order logic.

We assume an infinite supply of distinct symbols,
which are classified as follows:
\begin{itemize}
\item[1.] Logical symbols:

\begin{itemize}
\item[a)] Parentheses: (,);
\item[b)] Logical connectives: $\land$, $\neg$;
\item[c)] Existential quantifier: $\exists$;
\item[d)] Binary equality symbol:  $=$ (optional);
\item[e)] Two propositional symbols: $\Tr$ and $\Fa$.
 \end{itemize}

\item[2.] Non-logical symbols: 
\begin{itemize} 
\item[a)] countably many object symbols.  Object symbols are denoted
  by low-case letters;
\item[b)] for each positive integer $n>0$, countably many $n$-ary
  function symbols of arity $n$.  Function symbols are denoted by
  low-case letters;
\item[c)] for each positive integer $n$, countably many $n$-ary
  relation symbols, also called predicate or set symbols of arity $n$.
  We use upper-case letters to denote predicates.
\end{itemize}
\end{itemize}
 As usual, we identify object symbols with 0-ary
function symbols and propositional symbols with predicate symbols of
arity 0.

\begin{remark}\label{remark-symbols}
  In most parts of this paper, we do not make a formal distinction
  between variable and constant symbols. Symbols occurring free in a
  formula can be viewed as constants; symbols in the scope of a
  quantifier can be viewed as variables. In examples, we tend to
  quantify over $x$, $y$, $X$, $Y$, and leave $c$, $g$, $f$ and $P$,
  $Q$ free and treat them as constants.
\end{remark}

We define a vocabulary as any set of non-logical symbols. We denote
vocabularies by $\tau, \ \toD,\ldots$. We shall denote the set of 
function symbols of $\tau$ by $\tfun$, and 
we use $\symb$, $\symb_1$, $\symb_2$ etc., to refer to an arbitrary
symbol of the vocabulary.
We write $\bsymb$ to denote a sequence of symbols
$(\symb_1,\symb_2,\dots)$ or, depending on the context, simply the set
of symbols $\{\symb_1,\symb_2,\dots\}$. Likewise, $\X$ denotes a
sequence or a set of relational symbols (i.e, set variables or
constants), and $\x$ is used to denote a sequence or a set of object
symbols, etc..

A {\em term}  is defined inductively as follows:
\begin{itemize}
\item[ - ] an object symbol is a term;
\item[ - ] if $t_1$, \dots, $t_n$ are terms and $f$ is an $n$-ary function
symbol, where $n\geq 1$, then $f(t_1, \dots, t_n)$ is a term.
\end{itemize} 

A formula is defined by the following induction:
\begin{itemize}
\item[ - ] if $P$ is an $n$-ary predicate constant or variable, and
  $t_1, \dots, t_n$ are terms then $P(t_1, \dots, t_n)$ is a formula,
  called an {\em atomic formula} or simply an {\em atom};
\item[ - ] if $\phi, \psi$ are formulas, then so are
$\neg\phi, \phi\land\psi$;
\item[ - ] if $x$ is an object symbol, $f$ a function symbol, $X$
  is a predicate symbol and $\phi$ is a formula, then $\exists x\ 
  \phi$, $\exists f\ \phi$ and $\exists X\ \phi$ are formulas.
\end{itemize}

A bounded occurrence of symbol $\symb$ in formula $\phi$ is an
occurrence of $\symb$ in  a subformula $\exists \symb
\psi$ of $\phi$.  A free occurrence of $\symb$ 
in $\phi$ is an unbounded occurrence.  The set of symbols
which occur free in $\phi$ is denoted $\free{\phi}$. This set
can also be  defined inductively:
\begin{itemize}
\item [ - ] If $\phi$ is atomic, say of the form $A(t_1, \dots, t_n)$
 then the set $\free{\phi}$ 
is the set of all object, relational and functional symbols  occurring in $\phi$;
\item [ - ] $\free{\neg \phi} \  :=\  \free{\phi} $ ;
\item [ - ] $\free{\phi \land \psi} \  :=\  \free{\phi} \ \cup \ \free{\psi} $;
\item [ - ] $\free{\exists \sigma \ \phi} \  :=\  \free{\phi}\setminus \{\sigma\} $. \ignore{, for a non-logical symbol $\sigma$. \sobject, relational or functional symbol .}
\end{itemize}
A relation symbol $X$ has a negative (positive) occurrence in formula
$F$ if $X$ has a free occurrence in the scope of an odd (even) number
of occurrences of the negation symbol~$\neg$.

A formula $\phi$ is a formula {\em over vocabulary $\tau$ } if its free
symbols belong to $\tau$ ($\free{\phi} \subseteq \tau$). We use ${\rm
  SO}[\tau]$ to denote the set of all formulas over $\tau$; and we use
${\rm FO}[\tau]$ to denote the set of first-order formulas over
$\tau$, that is those without quantified predicate or function
variables.

We use $(\phi \vee \psi)$, $(\phi \supset \psi)$, $(\phi \equiv \psi)$,
$\forall x\ \phi$, $\forall f\ \phi$ and $\forall X \ \phi$, in the
standard way, as abbreviations for the formulas 
$\neg (\neg \phi \land \neg \psi)$, 
$\neg (\phi \land \neg \psi)$, 
$ \neg (\phi \land \neg \psi)\land \neg ( \psi \land \neg \phi)$, 
$\neg \exists x \ (\neg \phi)$, 
$\neg \exists f \ (\neg \phi)$, 
$\neg \exists X \ (\neg \phi)$, respectively. 

\vspace{3mm}

\ignore{ MARC/ THE NOTION OF VALUE WAS NOT CCURATELY TREATED BELOW:
Having defined the basic syntactic concepts, we define the semantic
concepts.  Let $A$ be a nonempty set.  The {\em value} of an $n$-ary
relation (function) symbol $\symb$ of vocabulary $\tau$ in $A$ is an
$n$-ary relation (function) in $A$. The value of a 0-ary function
symbol, i.e., an object constant or variable, is an element of the
domain $A$. If the value of a 0-ary relation symbol $Y$ is
$\emptyset$, we say $Y$ is {\em false}; if its value is $\{()\}$ (the
singleton of the empty tuple), we say $Y$ is {\em true}.  The value of
the equality symbol is always the identity relation on $A$. The value
of $\Tr$ is $\{()\}$ (true) and the value of $\Fa$ is $\emptyset$
(false).

A {\em structure} $\valI$ for a given vocabulary $\tau$ (in short, a
{\em $\tau$-structure}) is a tuple of a domain $\domI$, which is a
non-empty set, and a mapping of each symbol $\symb$ in $\tau$ to a
value $\symb^{\valI}$ in $\domI$.  If $\symb \in \tau$ and $I$ is a
$\tau$-structure, we say that $\valI$ {\em interprets} $\symb$.  We
also use letters $J$, $K$, $L$, $M$ to denote structures. Given $I$,
$\tauI{I}$ denotes the set of symbols interpreted by $I$.
} 

Having defined the basic syntactic concepts, we define the semantic
concepts.  Let $A$ be a nonempty set.  A {\em value} for an $n$-ary
relation (function) symbol $\symb$ of vocabulary $\tau$ in $A$ is an
$n$-ary relation (function) in $A$. A value for a 0-ary function
symbol, i.e., an object constant or variable, is an element of the
domain $A$.  A value for a 0-ary relation symbol $Y$ is either
$\emptyset$ or $\{()\}$, the singleton of the empty tuple. We identify
these two values with {\em false}, respectively {\em true}.  The value
of the equality symbol is always the identity relation on $A$. The
value of $\Tr$ is $\{()\}$ (true) and the value of $\Fa$ is
$\emptyset$ (false).

A {\em structure} $\valI$ for a given vocabulary $\tau$ (in short, a
{\em $\tau$-structure}) is a tuple of a domain $\domI$, which is a
non-empty set, and a mapping of each symbol $\symb$ in $\tau$ to a
value $\symb^{\valI}$ in $\domI$.  If $\symb \in \tau$ and $I$ is a
$\tau$-structure, we say that $\valI$ {\em interprets} $\symb$.  We
also use letters $J$, $K$, $L$, $M$ to denote structures. Given $I$,
$\tauI{I}$ denotes the set of symbols interpreted by $I$.

Let us introduce notation for constructing and modifying structures
with a shared domain $A$.  
Let $I$ be a $\tau$-structure, and $\bsymb$ be a tuple of symbols not
necessarily in $\tau$.  Structure $\valI[\bsymb:\bv]$ is a $\tau \cup
\bsymb$-structure, which is the same as $I$, except symbols $\bsymb$
are interpreted by values $\bv$ in $\domI$.  Given a $\tau$-structure
$\valI$ and a sub-vocabulary $\tau'\subseteq \tau$, the restriction of
$\valI$ to the symbols of $\tau'$ is denoted $\restr{\valI}{\tau'}$.

Let $t$ be a term, and let $\valI$ be a structure interpreting each
symbol in $t$. We define the {\em denotation} $\eval{t}{\valI}$ of $t$ under
$\valI$ by the usual induction:
\begin{itemize}
\item [ - ] if $t$ is an object symbol $\symb$, then $\eval{t}{\valI}$ is $\symb^I$,   the
value of $\symb$ in $I$;
\item  [ - ] if $t=f(t_1,..,t_n)$, then $\eval{t}{\valI}:=f^\valI(\eval{t_1}{\valI}\,..,\eval{t_n}{\valI})$.
\end{itemize}

Next we define the {\em satisfaction} or {\em truth} relation
$\models$.  Let $\valI$ be a structure and let $\phi$ be a formula
such that each free symbol in $\phi$ is interpreted by $\valI$.  We
define $\valI\models\phi$ (in words, {\em $\phi$ is true in $\valI$},
or {\em $\valI$ satisfies $\phi$}) by the following standard
induction:
\begin{itemize}
\item[ - ]  $\valI\models X(t_1,..,t_n)$ if
$(\eval{t_1}{\valI},..,\eval{t_n}{\valI}) \in X^\valI$ ;
\item[ - ]$\valI \models \psi_1\land \psi_2$ if $\valI \models \psi_1$ if
$\valI \models \psi_2$;
\item[ - ]$\valI \models\neg \psi$ if $\valI \not\models \psi$;
\item[ - ]$\valI \models\exists \symb \ \psi$ if for some value $v$
of $\symb$ in the domain $\domI$ of $\valI$,
$\valI[\symb:v] \models\psi$.
\end{itemize}
Note that the truth of a formula $\phi$ is only well-defined in a
structure interpreting each free symbol of $\phi$.  
We shall denote the truth value of $\phi$ in $\valI$ by $\phi^\valI$, i.e.,
if $\valI\models\phi$ then $\phi^\valI$ is true $(\{()\})$ and
otherwise, it is false ($\emptyset$).

Sometimes, we wish to investigate the truth value of a
formula $\phi$ as a function of the values assigned to a specific
tuple of symbols $\bsymb$. We then call these symbols the {\em
parameters} of $\phi$ and denote the formula by $\phi(\bsymb)$. Let
$\valI$ be some structure and let $\bv$ be a tuple of values for
$\bsymb$ in the domain $\domI$. We often write
$\valI\models\phi[\bv]$ to denote $\valI[\bsymb:\bv]\models\phi$.

\ignore{Let $X$ be an n-ary relation symbol and $\bd$ be an n-tuple of
  elements of $dom(\valIopen)$.  We define {\em a domain atom in $A$}
  as $X[\bd]$, and we define $\At{A}{\X}$ as the set of all domain
  atoms in $dom(\valIopen)$ of symbols in $\X$.}

Let $X$ be an n-ary relation symbol and $\bd$ be an n-tuple of
elements of some domain $A$.  We define {\em a domain atom in $A$} as
$X[\bd]$. For $I$ a structure with domain $A$, the value of $X[\bd]$
in $I$ is true if $\bd\in X^I$; otherwise it is false.  For a
vocabulary $\tau$, we define $\At{A}{\tau}$ as the set of all domain
atoms in domain $A$ over relation symbols in $\tau$. \ignore{ MARC: I
DONT THINK WE USED $\X$: For a set $\X$ of relation symbols, we define
$\At{A}{\X}$ as the set of all domain atoms in domain $A$ over
relation symbols in $\X$. In general, for a given vocabulary $\tau$
with predicates $\X$, we denote $\At{A}{\X}$ as $\At{A}{\tau}$.}

Suppose we are given a structure $\valI$ with domain $\domI$, a tuple
$\x$ of $n$ variables and a first-order formula $\phi(\x)$ such that
all its free symbols not in $\x$ are interpreted by $\valI$. The
relation {\em defined} by ${\phi(\x)}$ in the structure $\valI$ is
defined as follows: 
$$ 
R:= \{ \ba \ |\ \valI \models \phi[\ba],\ \ba\in
(\domI)^n \}. 
$$ 
We call $R$ {\em first-order definable} in $\valI$.
In this paper, we study inductive and non-monotone inductive
definability.  In this context, defined relations are not, in general,
first-order definable.



\subsection{Preliminaries from Set and Lattice Theories}
\label{preliminaries-lattices}

\subsubsection{Orders, Lattices, operators and fixpoints}

A {\em pre-ordered set} is a structured set $\langle W, \leq \rangle$,
where $W$ is an arbitrary set and $\leq $ is a pre-order on $W$, i.e.,
a reflexive and transitive binary relation. As usual, $x<y$ is a
shorthand for $x\leq y \land y\not\leq x$. A {\em pre-well-founded
set} is a pre-ordered set where $\leq$ is a pre-order such that every
non-empty set $S\subseteq W$ contains a minimal element, i.e., an
element $x$ such that for each $y \in S$, if $y\leq x$ then $x\leq y$.
Equivalently, it is a set without infinite descending sequence of
elements $x_0 > x_1 > x_2 > \ldots$.

A {\em partially ordered set}, or simply {\em poset}, is an
asymmetric pre-ordered set $\langle W, \leq \rangle$, i.e., one
such that $x \leq y$ and $y\leq x$ implies $x=y$. A {\em well-founded
set} is a pre-well-founded poset.

A {\em lattice} is a poset $\langle L, \leq\rangle$ such that every
finite set $S \subseteq L$ has a least upper bound $\lub(S)$, the {\em
supremum of $S$}, and a greatest lower bound $\glb(S)$, the {\em
infimum of $S$}. A lattice $\langle L, \leq\rangle$ is {\em complete}
if {\em every} (not necessarily finite) subset of $L$ has both a
supremum and an infimum.  Consequently, a complete lattice has a least
element ($\bot$) and a greatest element ($\top$). An example of a
complete lattice is the power set lattice $\langle \Pow{A}, \subseteq
\rangle$ of some set $A$.  For any set $S$ of elements of this lattice
(i.e., for any set $S$ of subsets of $A$), its least upper bound is
the union of these elements, $\lub(S)=\cup S$.  Thus, the greatest
element $\top$ of $\langle \Pow{A}, \subseteq \rangle$ is $\cup
\Pow{A}$, which is $A$.  Similarly, $\glb(S)=\cap S$, and the least
element $\bot$ of this lattice is $\cap \Pow{A}$, which is
$\emptyset$.

Given a lattice $\langle L, \leq \rangle$, an operator $\G:L\ra L$ is
{\em monotone} with respect to $\leq$ if $x\leq y$ implies $\G(x) \leq
\G(y)$. Operator $\G$ is {\em non-monotone}, if it is not monotone. A {\em
  pre-fixpoint} of $\G$ is a lattice element $x$ such that
$\G(x)\leq x$.  The following theorem was obtained by Tarski in 1939
and is sometimes referred to as the Knaster-Tarski theorem because it
improves their earlier joint result. The theorem was published in
\cite{Tarski55}, and it is one of the basic tools to study fixpoints 
of operators on lattices.

\begin{theorem}[existence of a least fixpoint]
Every monotone operator over a complete lattice $\langle W, \leq
\rangle$ has a complete lattice of fixpoints (and hence a least
fixpoint $\lfp(\G)$ and greatest fixpoint $\gfp(\G)$).
\end{theorem}

This least fixpoint $\lfp(\G)$ is the least pre-fixpoint of $\G$ and
is the supremum of the sequence $(x^\xi)_\xi$ which is defined
inductively
$$ {x}^{\xi} := \G({x}^{<\xi}), \ \ \mbox{and}\ \ {x}^{<\xi}:=
 \lub \{ {x}^{\eta} |0\leq \eta<\xi\}.  
$$ 
Notice that ${x}^{<0}$ is, by definition, $\bot$.

An operator $\G$ is {\em anti-monotone} if $x\leq y$ implies $\G(y)
\leq \G(x)$.
\begin{proposition}\label{composition}
If $\G_1$ and $\G_2$ are anti-monotone operators, then $\G_1 \circ
\G_2$, the composition of $\G_1$ and $\G_2$, is monotone. 
\end{proposition} 
\noindent In particular, the square $\G^2 = \G \circ \G$ of an anti-monotone
operator is monotone.

An {\em oscillating pair} of an operator $\G$ is a pair $(x,y)$ such that
$\G(x)=y$ and $\G(y)=x$. An anti-monotone operator $\G$ in a complete
lattice has a maximal oscillating pair $(x,y)$, i.e., for any
oscillating pair $(x',y')$, it holds that $x \leq x'$ and $y'\leq
y$. Since $(y,x)$ is also an oscillating pair, it follows that $x\leq
y$. Moreover, since each fixpoint $z$ of $\G$ corresponds to an
oscillating pair $(z,z)$, it follows that $x\leq z\leq y$. The maximal
oscillating pair $(x,y)$ of $\G$ can be constructed by an alternating
fixpoint computation.  Define four sequences
$(x^\xi)_\xi, (x^{<\xi})_\xi, (y^\xi)_\xi, (y^{<\xi})_\xi$ by the
following transfinite induction:
\begin{itemize}
\item [ - ] $x^{<\xi} = \lub(\{x^\eta : \eta<\xi\})$,
\item [ - ] $x^\xi = \G(y^{<\xi})$,
\item [ - ] $y^{<\xi} = \glb(\{y^\eta : \eta<\xi\})$,
\item [ - ]  $y^\xi = \G(x^{<\xi})$.
\end{itemize}
Note that $x^{<0}=\bot$ and $y^{<0}=\top$. It can be shown that for
each $\xi$,  $x^{<\xi}\leq x^\xi \leq
y^\xi \leq y^{<\xi}$. The following theorem holds. 
\begin{theorem} \cite{VanGelder93} 
  The sequence $(x^\xi)_\xi$ is ascending and its supremum is
  $\lfp(\G^2)$.  The sequence $(y^\xi)_\xi$ is descending and its
  infimum is $\gfp(\G^2)$. The pair $(\lfp(\G^2),\gfp(\G^2))$ is the
  maximal oscillating pair of $\G$.
\end{theorem}

We will use the following simple lemma on lattices.

\begin{lemma}\label{lemma-fixpoints-on-lattice} 
  Let $\G_1, \G_2$ be two monotone operators in a lattice with least
  fixpoints $\lfp(\G_1)=o_1, \lfp(\G_2)=o_2$ respectively.
 
(a) if $\G_1(x) \leq \G_2(x)$ for each $x\leq o_1$ then $o_1 \leq o_2$;
 
(b) if $\G_1(x) \leq \G_2(x)$ for each $x\geq o_2$ then $o_1 \leq o_2$. 
\end{lemma} 

\begin{proof} 
(a) Define $o_i^{\xi}$ and  $o_i^{<\xi}$ by induction:
\begin{itemize} 
\item[ - ] $o_i^{<\xi}:=lub(\{o_i^{\eta}\  | \ \eta<\xi\})$,
\item[ - ]  $o_i^{\xi}:=\G_i(o_i^{<\xi})$. 
\end{itemize} 
Then $o_i$ is the limit of the increasing sequence 
$(o_i^{\xi})_{\xi}$. Moreover, for each $\xi: o_1^{\xi}\leq o_1$ and 
$o_1^{<\xi}\leq o_1$. 
 
The proof is by transfinite induction. Obviously $o_1^{0}=\G_1(\bot) 
\leq \G_2(\bot)=o_2^{0}$. Assume that for each $\eta<\xi$, 
$o_1^{\eta}\leq o_2^{\eta}$. Then also $o_1^{<\xi}\leq o_2^{<\xi}$. 
Then $o_1^{\xi} = \G_1( o_1^{<\xi}) \leq \G_2(o_1^{<\xi}) \leq 
\G_2(o_2^{<\xi}) = o_2^{\xi}$.

(b) It holds that $o_i$ is the least fixpoint and hence the least
pre-fixpoint of $\G_i$. Hence $o_i = glb(\{ x\  |\  \G_i(x) \leq x\})$.
Since $\G_1(x) \leq \G_2(x)$ for each $x\geq o_2$, it holds that if $x$
is a pre-fixpoint of $\G_2$, then $x$ is also a pre-fixpoint of $\G_1$.
Thus we have $\{ x\  | \ \G_2(x) \leq x\} \subseteq \{ x \  |  \ \G_1(x) \leq x\}$.
Since $o_1=glb(\{ x \ | \ \G_1(x) \leq x\})$,  we have $o_1\leq o_2$.
\end{proof}

\subsubsection{Lattice Homomorphisms and Congruences}
\label{Sec-lattice-homo}

\ignore{SEARCH REFERENCE.}

Let $\langle L, \leq\rangle$ be a complete lattice and let $\congr$ be
an arbitary equivalence relation (i.e. a reflexive, symmetric and
transitive relation) on $L$. For any $x\in L$, we denote its
equivalence class $\{y \in L \ | \ x\congr y\}$ by $|x|$. The
collection of equivalence classes is denoted by $L^\congr$. The
relation $\congr$ can be extended to tuples: $(x_1,\ldots, x_n) \congr
(y_1,\ldots, y_n)$ if $x_1\congr y_1$ and \ldots and $x_n\congr y_n$.
It is extended to subsets of $L$ by defining for all $S, S'\subseteq
L$: $S\congr S'$ if for each $x\in S$ there exists $x'\in S'$ such
that $x \congr x'$ and vice versa, for each $x'\in S$ there exists $x
\in S$ such that $x\congr x'$.

An equivalence relation $\congr$ on $L$ is called a {\em lattice
congruence} of $\langle L, \leq\rangle$ if for each pair $S, S'
\subseteq L$, $S\congr S'$ implies that $\lub(S) \congr \lub(S')$ and
$\glb(S) \congr \glb(S')$. We can define a binary relation $\leq$ on
$L^\congr$: for all $S, S' \in L^\congr$, define $S \leq S'$ if for some $x
\in S, y\in S': x\leq y$. It can be shown easily that if $\congr$ is
a lattice congruence, then the structure $\langle L^\congr,
\leq\rangle$ is a complete lattice.

Let $\langle L, \leq\rangle$, $\langle L', \leq'\rangle$ be two
complete lattices. A mapping $h: L \ra L'$ is called a {\em lattice
homomorphism} if it is a mapping onto (i.e., $h(L) = L'$), and for each
$S\subseteq L$, $h(\glb_\leq(S)) = \glb_{\leq'}(h(S))$ and
$h(\lub_\leq(S)) = \lub_{\leq'}(h(S))$.

The notions of lattice congruence and lattice homomorphism are
strongly related.  A homomorphism $h: L \ra L'$ induces a relation
$\congr$ on $L$ where $x \congr y$ holds if $h(x)=h(y)$, for all $x, y
\in L$. The relation $\congr$ is a lattice congruence of $\langle L,
\leq\rangle$. Moreover, $\langle L^\congr, \leq\rangle$ and $\langle
L', \leq'\rangle$ are isomorphic. Vice versa, for each lattice
congruence $\congr$, the mapping $L\ra L^\congr$ such that $ x \ra |x|$ is a
lattice homomorphism.

Let $h$ be a lattice homomorphism from $\langle L, \leq\rangle$ to
$\langle L', \leq'\rangle$ and $\congr$ the induced congruence on
$L$. We say that an operator $O: L\ra L$ preserves $\congr$ if for all
$x, y \in L$, $x\congr y$ implies $O(x) \congr O(y)$. In general, for
any operator $O: L^m\ra L^n$, we say that $O$ preserves $\congr$ if
for any pair of $\xx, \yy \in L^m$, $\xx\congr\yy$ implies $O(\xx)
\congr O(\yy)$.

If $O:L\ra L$ preserves $\congr$ then for any $x' \in L'$, for any
$x_1, y_1 \in h^{-1}(x'), h(O(x_1)) = h(O(x_2))$.  We then define the
homomorphic image $O^h: L'\ra L'$ of $O$. This operator maps $x'\in
L'$ to $y'$ iff for each $x \in h^{-1}(x')$, $y' = h(O(x))$. This
definition can be extended to operators $O:L^m \ra L^n$.

The following proposition describes relationships between $O$ and
$O^h$.
\begin{proposition}\label{PropCongruence} Let $O$ be an operator which preserves $\congr$. 
\begin{itemize}
\item[(a)] If $O$ is (anti-)monotone, then $O^h$ is (anti-)monotone.
\item[(b)]  If $O$ is monotone then $h(\lfp(O)) = \lfp(O^h)$ and
$h(\gfp(O)) = \gfp(O^h)$. 
\item[(c)] If $O$ is anti-monotone and $(x,y)$ its maximal oscillating pair
then  $(h(x),h(y))$ is the maximal  oscillating pair of $O^h$. 
\end{itemize}
\end{proposition}
\begin{proof}
The proof of item (a) is straightforward and is
omitted.

\noindent (b) The least fixpoint $\lfp(O)$ is the limit of the 
sequence $(x^\xi)_{\xi\geq 0}$ which is defined inductively
$$ 
{x}^{\xi} := O({x}^{<\xi}), \ \ \mbox{and}\ \ {x}^{<\xi}:= \lub \{
{x}^{\eta} |0\leq \eta<\xi\}.  
$$ 
The point $\lfp(O^h)$ is the limit of the sequence $(y^\xi)_\xi$
defined similarly using $O^h$ in the lattice $L'$.  By a
straightforward induction, one can show that for each ordinal $\xi$,
$h(x^\xi) = y^\xi$. Since $h$ is a lattice homomorphism,
$h(\lfp(O))=h(\lub(\{x^\xi \ | \ \xi \geq 0\})) = \lub(\{h(x^\xi) \ | \ \xi
\geq 0\}) = \lub(\{y^\xi\  |\  \xi
\geq 0\}) = \lfp(O^h)$. 
The proof that $h(\gfp(O)) = \gfp(O^h)$ is similar.\\

\noindent (c)  It is easy to show that $({O^2})^h$ is
$({O^h})^2$. Then (c) is a direct consequence of (b) and the fact that
the maximal oscillating pair of $O$ and $O^h$ are
$(\lfp(O^2),\gfp(O^2))$, respectively
$(\lfp(({O^h})^2),\gfp(({O^h})^2))$.

\end{proof}

\subsubsection{Structure lattices}

The type of lattices that play a central role in this paper are the sets
of structures that extend a given structure. For a given vocabulary
$\tau$ and structure $\baseI$ such that $\tauI{\baseI} \subseteq
\tau$, define $\vallat{\baseI}{\tau}$ as the set of $\tau$-structures
that extend $\baseI$, i.e. the set of $\tau$-structures $\valI$ such
that $\valI|_{\tauI{\baseI}} =\baseI$.

For any pair $\valI_1, \valI_2$ of $\tau$-structures, define $\valI_1
\leqt \valI_2$ if both structures have the same interpreted symbols ,
the same domain and the same values for all object and function
symbols and for each interpreted relation symbol $X$, $X^{\valI_1}
\subseteq X^{\valI_2}$.  

The structured set $\langle \vallat{\baseI}{\tau},\leqt\rangle$ is a
partial order. In general, it is not a lattice, because elements $I,
J$ giving different interpretation to a function symbol $f\in \tau
\setminus \tauI{\baseI}$ have no greatest lowerbound nor least
upperbound in $\vallat{\baseI}{\tau}$. However, if $\baseI$ interprets
all function symbols of $\tau$, that is, if $\tfun\subseteq
\tauI{\baseI}$, then $\langle \vallat{\baseI}{\tau},\leqt\rangle$ is a
complete lattice. Its least element is the structure $\bot_\baseI :=
\baseI[\X:\emptyset]$ assigning the empty relations to all symbols $X$
in $\tau\setminus \tauI{\baseI}$ and its largest element $\top_\baseI$
is the structure assigning the cartesian product $A^n$ to each $n$-ary
symbol $X \in \tau\setminus \tauI{\baseI}$.

The lattice $\langle \vallat{\baseI}{\tau}$ contains many sublattices.
In particular, for any structure $K$ extending $\baseI$ such that
$\tauI{\baseI} \subseteq \tauI{K} \subseteq \tau$, $\langle
\vallat{K}{\tau},\leqt\rangle$ is a sublattice of $\langle
\vallat{\baseI}{\tau},\leqt\rangle$. 

In this paper, the family of structure lattices and homomorphisms and
congruences on them play an important role.

\section{ID-Logic}\label{Sec:ID-logic}
\label{SecID}

In this section, we present an extension of classical logic with
non-monotone inductive definitions. This work extends previous work of
the authors \cite{Denecker:CL2000,Ternovskaia:1999}.

\subsection{Syntax}

First, we introduce the notion of a definition.  \ignore{ Let us fix
  some vocabulary $\tau$.  } We introduce a new binary connective
$\rul$, called the {\em definitional implication}.  A {\em definition $\D$} is a
set of rules of the form
\begin{equation}\label{eq-rule}
\forall\x\ (X(\bt) \rul \varphi) \ \ \mbox{where}
\end{equation} 
\begin{itemize}
\item $\x$ is a tuple of object variables,
\item $X$ is a predicate symbol (i.e., a predicate constant
or variable) of some arity $r$, 
\item$\bt$ is a tuple of terms of length $r$, \ignore{ of the vocabulary $\tau$,}
\item $\varphi$ is an arbitrary  first-order formula.\ignore{ of $\tau$.}
\ignore{ which may contain free object or relational variables. }
\end{itemize}

The definitional implication $\rul$ must be distinguished from material
implication. A rule $\forall\x\ (X(\bt) \rul \varphi)$ in a definition
does not correspond to the disjunction $\forall\x (X(\bt) \lor \neg
\varphi)$, but implies it.   
Note that in front of rules, we allow only universal quantifiers. In
the rule (\ref{eq-rule}), $X(\bt)$ is called the {\em head} and
$\varphi$ is the {\em body} of the rule.

\begin{example}\label{example-even-odd-def}
The following expression is a simultaneous definition 
of the sets of even and odd numbers on the structure
of the natural numbers with zero and the successor function:
\begin{equation}\label{eq-even-odd-def}
\left\{ \begin{array}{l}
\forall x\ (E(x) \rul x=0) ,\\
\forall x\ (E(s(x)) \rul O(x)),\\
\forall x\ (O(s(x)) \rul E(x))
\end{array}
\right\}.
\end{equation}
\end{example}

\begin{example}\label{example-TC-def}
This is the definition of the transitive closure of a directed graph $G$: 
\begin{equation}\label{eq-TC-def}
\left\{ \begin{array}{l}
\forall x\ \forall y \ (T(x,y) \rul G(x,y)),\\
\forall x\ \forall y \ (T(x,y) \rul \exists z\ (T(x,z) \land T(z,y)))
\end{array}
\right\}.
\end{equation}
\end{example}

The definitions of bound and free occurrence of a symbol in a formula
extend to the case of a rule and a definition $\D$.  A {\em defined
symbol} of $\Delta$ is a relation symbol that occurs in the head of at
least one rule of $\D$; other relation, object and function symbols
are called {\em open}. In the Example \ref{example-even-odd-def}
above, $E$ and $O$ are defined predicate symbols, and $s$ is an open
function symbol.  In the Example \ref{example-TC-def}, $T$ is a
defined predicate symbol, and $G$ is an open predicate symbol.
We call $\D$ a {\em positive} definition if no defined predicate $X$
has a negative occurrence in the body of a rule of $\D$.  The
definitions in Example \ref{example-even-odd-def} and Example
\ref{example-TC-def} are positive.

Let $\tau$ be a vocabulary interpreting all free symbols of $\D$. The
subset of defined symbols of definition $\D$ is denoted $\defp{\D}$.
The set of open symbols of $\D$ in $\tau$ is denoted $\topen{\D}$.
The sets $\defp{\D}$ and $\topen{\D}$ form a partition of $\tau$,
i.e., $\defp{\D}\cup \topen{\D} = \tau$, and $\defp{\D}\cap \topen{\D}
= \emptyset$.


Now we are ready to define the well-formed formulas of the logic.
\ignore{Let $\tau$ be a fixed vocabulary.}  A {\em well-formed
  formula} of the Logic for Non-Monotone Inductive Definitions\ignore{
  over vocabulary $\tau$}, briefly a ID-formula, is defined by the
following induction:
\begin{itemize}
\item If $X$ is an $n$-ary predicate symbol\ignore{ of $\tau$}, and $t_1,
  \dots, t_n$ are terms then $X(t_1, \dots, t_n)$ is a formula.
  \ignore{, called an {\em atomic formula} or simply an {\em atom}.}
\item If $\D$ is a definition then $\D$ is a formula.
\item If $\phi, \psi$ are formulas, then so are
$(\neg\phi)$ and $(\phi\land\psi)$.
\item  If $\phi$ is a formula, then $\exists \symb\ \phi$ is a
  formula. \ignore{WHY DID WE NOT ALLOW FUNCTION QUANTIFIERS? If $x$
    is an object variable, $X$ is a predicate variable and $\phi$ a
    formula, then $(\exists x\ \phi)$, and $(\exists X\ \phi)$ are
    formulas.}
\end{itemize}
The definitions of bound and free occurrence of a symbol in a formula
(see Section 3) extend to ID-formulas $\phi$. We shall denote the
set of symbols with free occurrences in $\phi$ by $\free{\phi}$.

A formula $\phi$ is an ID-formula over a vocabulary $\tau$ if
$\free{\phi}\subseteq \tau$.  We use ${\rm SO(ID)}[\tau]$ to denote
the set of all formulas of our logic over fixed vocabulary $\tau$.
The first-order fragment ${\rm FO(ID)}[\tau]$ is defined in the same
way, except that quantification over set and function symbols is not
allowed.

\begin{example}\label{example-even-odd-formula}
  In the structure of the natural numbers, the following formula
  expresses that  $E$ and $O$ are respectively the set of even and odd
  numbers, and that the number 2, which is representated by $s(s(0))$,
  belongs to $E$. 
\begin{equation}\label{eq-even-odd-formula}
\left\{ \begin{array}{l}
\forall x\ (E(x) \rul x=0),\\
\forall x\ (E(s(x)) \rul O(x)),\\
\forall x\ (O(s(x)) \rul E(x))
\end{array}
\right\} \land E(s(s(0))).
\end{equation}

\end{example}

\ignore{
\begin{example}\label{example-TC-formula}
The following formula expresses that the transitive closure $X$
of any graph $G$ is transitive
\begin{equation}\label{eq-TC-formula}
\forall T \forall G\ \left( 
\left\{ \begin{array}{l}
\forall x\ \forall y \ (T(x,y) \rul G(x,y)),\\
\forall x\ \forall y \forall z \ (T(x,y) \rul T(x,z) \land T(z,y))
\end{array}
\right\}\supset \forall x \forall y \forall z \ (T(x,y) \land T(y,z) \supset T(x,z)\right)
\end{equation}

\end{example}}

\begin{example}\label{Peano} 
The Peano induction axiom is: 
\[  \forall P [P(0) \land \forall n (P(n) \supset P(s(n))) \supset \forall n P(n)]. \]
This axiom can be formulated in ID-logic as: 
\begin{equation}\label{IndAx1}
\exists N \left [ \defin{ \forall x \ (N(x) \rul  x=0), \\ 
\forall x\ ( N(s(x)) \rul N(x))}
\land \forall x \ N(x)\right ] .
\end{equation}
The first conjunct in this formula defines the set variable $N$ as
the set of the natural numbers through the standard induction.  The
second conjunct expresses that each domain element is a natural
number. An equivalent alternative formalisation is: 
\begin{equation}\label{IndAx2}
\forall N \left [ \defin{ \forall x \ (N(x) \rul  x=0), \\ 
\forall x\ ( N(s(x)) \rul N(x)}
\supset \forall x \ N(x)\right ].
\end{equation}
The equivalence of axioms (\ref{IndAx1}) and (\ref{IndAx2})  follows from the fact that 
the defined set is unique. The uniqueness is guaranteed by the semantics we define next.

In the sequel, we use $T_{\mathbb{N}}$ to denote the ID-theory
consisting of axiom (\ref{IndAx1}) and the two other Peano axioms:
\[\begin{array}{c}
  \forall n \  \neg (s(n) = 0) ,\\
  \forall n \forall m\  ( s(n)=s(m) \supset n=m ).
\end{array}\]
\ignore{
Let $T_{\mathbb{N}}$ be the ID-theory obtained from this theory by
substituting the following axiom $Ax_{ind}$ for the induction axiom:
$$
\exists N \left [ \defin{ \forall x \ (N(x) \rul x=0) , \\ \forall x \ N(s(x)) \rul N(x)}
\land \forall x \ N(x)\right ]
$$
$T_{\mathbb{N}}$ is a theory in the standard vocabulary of the natural
numbers. We will prove that this theory is equivalent to the Peano
axioms with the second-order induction axiom.}
\end{example}


\subsection{Semantics} \label{semantics}

\ignore{The satisfaction relation between structures and
  ID-formulas \ignore{well-formed formula of the Logic for
    Non-Monotone Inductive Definitions} is defined trough the standard
  truth recursion (see Section \ref{{preliminaries-logic}}) augmented
  with an extra rule defining when a structure $I$ satisfies a
  definition $\D$.  }  
The exposition below is a synthesis of different approaches to the
well-founded semantics, in particular those presented in
\cite{VanGelder93,Fitting2001,Denecker2001:TOCL}. We begin by defining the
operator associated with a definition $\D$.  We shall assume that
definitions are finite sets of rules. The theory can easily be
extended to the infinite case (using infinitary logic).
\ignore{{\tt MOVE THE SENTENCE ABOVE TO 
SYNTAX ??? \marc{I prefer not}} \marginpar{???}}

\ignore{
{\tiny 
Let $\D$ be an arbitrary definition with defined relational symbols
$\X:=(X_1, \dots , X_n)$.  For each defined symbol $X$ of $\D$, we
define:
\begin{equation}
 \label{eq-varphi}
\varphi_{X}(\x) :=     \exists \y_1\ (\x=\bt_1 \land \varphi_1)
                            \lor \ \dots
                            \lor \  \exists \y_m\ (\x=\bt_m \land\varphi_m),
\end{equation}
where $\x$ is a tuple of new object variables, and $\forall \y_1\ 
(X(\bt_1) \rul \varphi_1)$, \dots , $\forall \y_m\ (X(\bt_m)
\rul \varphi_m)$ are the rules of $\D$ with $X$ in the head.}

{\tt ABOVE: We used to have a different (longer) explanation of how
the formula $\varphi_{X}(\x) $ is obtained. A definition was first 
transformed, in particular rules with the same head were combined,
 etc. remember? 
Then you replaced it with this more 
compact explanation, It is correct, I checked, but from my experience it is not
very intuitive. I gave the paper to read to my student who is 
quite experienced in reading hard papers, and she had difficulties.
Should we go back to the longer explanation? Did we keep it in the old files?

MARC: I DIDNT FIND THE OLD TEXT; BUT WHAT I PROPOSE IS THE REPLACE ABOVE PARAGRAPH BY THE NEXT WHICH IS INDEED MORE INSIGHTFUL. }
}
Any definition containing multiple rules with the same predicate in
the head can be easily transformed into a definition with only one
rule per defined predicate.
\begin{example}
The following definition of even numbers
\[
\defin{ 
\forall x\ (E(y) \rul y=0) ,\\
\forall x\ (E(s(s(x))) \rul E(x))}
\]
is equivalent to this one:
\[\defin{ 
\forall x\ (E(y) \rul y=0 \lor \exists x (y=s(s(x))\land E(x)))}.
\]
\end{example}
In general, let $\D$ be an arbitrary definition with defined
relational symbols $\X:=(X_1, \dots , X_n)$.  For each defined symbol
$X$ of $\D$, we define:
\begin{equation}
 \label{eq-varphi}
\varphi_{X}(\x) :=     \exists \y_1\ (\x=\bt_1 \land \varphi_1)
                            \lor \ \dots
                            \lor \  \exists \y_m\ (\x=\bt_m \land\varphi_m),
\end{equation}
where $\x$ is a tuple of new object variables, and $\forall \y_1\ 
(X(\bt_1) \rul \varphi_1)$, \dots , $\forall \y_m\ (X(\bt_m)
\rul \varphi_m)$ are the rules of $\D$ with $X$ in the head.
Then $\D$ is equivalent to the definition $\D'$ consisting of rules
$\forall \x (X(\x) \rul \varphi_X(\x))$. The formulas
$\varphi_{X}(\x)$ play an important role in defining the semantics of
definitions.

Let $\D$ be definition over a vocabulary $\tau$.
\begin{definition} [operator $\G_\D$] \label{DTp} 
  We introduce a total unary operator $\G_\D \ :\ {\cal I} \mapsto
  {\cal I}$ where ${\cal I}$ is the class of all $\tau$-structures.
  We have $I'=\G_\D(I)$ iff
\begin{itemize}
\item $\dom{I} = \dom{{I'}}$,
\item for each open symbol $\symb$, $\ \  \symb^{I'}=\symb^{I}$ and
\item for each defined symbol $X \in\defp{\D}$, 
$$
X^{I'}\  :=\  \{ \ba \ |\ I\models \varphi_{X}[\ba] \}, 
$$
where
$\varphi_{X_i}$ is defined by equation (\ref{eq-varphi}). 
\end{itemize}
\end{definition}

\ignore{ MARC: IN ORDER TO KEEP DISCUSSION FOCUSSED, I PROPOSE TO
DELETE  STATEMENT BELOW :
In the notation used in a fixpoint logic, the definition $\D$ would
be represented by the tuple of formulas $(\varphi_{X_1}(\x_1), \dots
,\varphi_{X_n}(\x_n))$\ignore{THIS IS NOT REALLY SO}. The operator
associated with this tuple would correspond to the above defined one,
but would operate on tuples of values $\R$ for the defined symbols
$\X$ rather than on structures.  The same operator has been defined
also in the context of logic programming where it is called the
Immediate Consequence Operator \cite{vanEmden76}.
} 
 
Let $\baseIo$ be a structure interpreting the open symbols of $\D$ in
$\tau$. Lattice $\langle \vallat{\baseIo}{\tau},\leqt\rangle$ consists
of all $\tau$-structures that extend $\baseIo$.  Operator $\G_\D$ is
an operator on this lattice.  If $\D$ is a positive definition (no
negative occurrences of defined symbols in rule bodies), then $\G_\D$
will be monotone.
\ignore{ \marc{Forward reference: Drop this? In this case, we will prove in
theorem \ref{theorem-Mon} that the model of the definition coincides
with the least fixpoint of $\G_\D$.}}  The least fixpoint is the limit
of the sequence $(I^\xi)_\xi$ which is defined inductively:
$$ 
{I}^{\xi} := \G_\D({I}^{<\xi}), \ \ \mbox{and}\ \ {I}^{<\xi}:= \bigsqcup \{ {I}^{\eta}\  |\  0 \leq \eta<\xi\}.  
$$
Notice that ${I}^{<0}$ is, by definition, the bottom element
$\bot_\baseIo := \baseIo[\X:\emptyset]$ in the lattice.
 
In general, $\G_\D$ is a non-monotone operator with no or multiple
minimal fixpoints.  Iterating the operator starting from the bottom
element may oscillate and never reach a fixpoint, or, when it does
reach a fixpoint, this fixpoint may not be the intended fixpoint.

\begin{example}\label{Ex:prop1}
Consider the following propositional definition:
$$
\D_0 := \defin{P \rul \Tr,\\ Q\rul\neg P ,\\ Q\rul Q }.
$$
Formally, structures of $\D_0$ are mappings of the symbols
$P,Q$ to 0-ary relations. We will represent such a structure in a
more traditional way as the set of the propositional symbols that are
true (i.e., that are interpreted by $\{()\}$). 

Notice that, in definition $\D_0$,$Q$ depends on $P$. In ID-logic this
definition is understood as a 2-level iterated inductive
definition $(\D_{01},\D_{02})$, where
$$
\begin{array}{l}
 \D_{01}:=\{P \rul \Tr\}, \\
 \D_{02}:=\{Q\rul\neg P\ ,\  Q\rul Q\}.
\end{array}
$$
By applying iterated induction, we obtain $\{P\}$ for the first level,
and then $\emptyset$ for the second. Consequently, the intended model of
this definition is $\{P\}$. On the other hand, if we iterate the
operator $\G_{\D_0}$ from the empty structure, we obtain immediately
the fixpoint $\{P,Q\}$.

\end{example}

The intuition underlying the semantics is to use definitions to
perform iterated induction, while following the implicit dependency
order given by the rules.  We explain how this intuition is formalised
in the well-founded semantics. We compute a converging sequence of
pairs $(I^\xi,J^\xi)_{\xi\geq 0}$ of $\tau$-structures extending
$\baseIo$. In each pair, $I^\xi$ represents a lower bound to the
intended model of $\D$ extending $\baseIo$; $J^\xi$ represents an
upper bound: domain atoms true in $I^\xi$ can be derived from the
definition; atoms false in $J^\xi$ cannot be derived; for all atoms
false in $I^\xi$ and true in $J^\xi$, it is not determined yet whether
they can be derived or not. Alternatively, a pair $I^{\xi}\sq J^{\xi}$
can be understood as a 3-valued structure defining the truth value of
part of the defined domain atoms, namely those domain atoms $A$ for
which $A^{I^\xi}=A^{J^\xi}$.  Thus, the pair $(I^\xi,J^\xi)$
represents approximate information about what can and what cannot be
derived from $\D$ in $\baseIo$.

The construction process starts with the pair
$(\bot_\baseIo,\top_\baseIo)$ of the least and largest element in the
lattice $\vallat{\baseIo}{\tau}$. This pair obviously consists of a
lower and and an upper bound of what can be derived from the
definition. Assuming we have obtained a pair $(I^\xi,J^\xi)$ of a safe
lower and upper bound, we then apply an operation which transforms
this pair into a new pair $(I^{\xi+1},J^{\xi+1})$ with an improved
lower and upper bound. By iterating this operation, a sequence
$(I^{\xi},J^{\xi})_{\xi\geq 0}$ of increasing precision is
constructed.  The sequence of lower bounds $(I^{\xi})_{\xi\geq 0}$ is
monotonically increasing and has a limit $I$ (its $\lub$); the
sequence of upper bounds $(J^{\xi})_{\xi\geq 0}$ is monotonically
decreasing and has a limit $J$ (its $\glb$) such that $I\sq J$.  The
pair of limits $(I,J)$ is the result of the construction and
represents the information that can be derived from $\D$ in the
context of the structure $\baseIo$. The definition $\D$ properly
defines its defined symbols in $\baseIo$ if $I=J$, that is, if for
each defined domain atom $A$, $A^I=A^J$. If $I=J$, then we will call
$\D$ {\em total} in $\baseIo$ and $I$ the {\em extension of $\baseIo$
defined by $\D$}. If $I\neq J$, then there will be no extension of
$\baseIo$ defined by $\D$.

We now explain how a pair $(I^{\xi},J^{\xi})$ of lower and upperbound
is refined into a new pair $(I^{\xi+1},J^{\xi+1})$. The idea is to
compute the new lower bound $I^{\xi+1}$ and upper bound $J^{\xi+1}$ by
monotone induction using the existing bounds $(I^\xi,J^\xi)$. We
cannot use $\G_\D$ for this, due to its non-monotonicity, but there is
a way.

In general, defined symbols have positive and negative occurrences in
the rule bodies $\varphi_{X}(\x)$. The negative occurrences are
responsible for the non-monotone behaviour of the operator $\G_\D$:
adding {\em more} tuples to the value of a negatively occurring
defined symbol in $\varphi_{X}(\x)$ has an anti-monotone effect on the
derived relation and may lead to the derivation of {\em fewer} tuples
$\ba$ satisfying this formula.  Thus we can eliminate the
non-monotonicity of $\G_\D$ and set up a monotone induction process
using $\D$ if we {\em fix} the value of negative occurrences of
defined symbols in rule bodies. Suppose we choose a fixed structure
$M$ to evaluate the negative occurrences of defined symbols in rule
bodies. We can then perform a monotonic derivation process
$\bot_\baseIo, K^1, K^2,\dots$ in which each $K^{i+1}$ is derived
from $\D$ by evaluating positive occurrences of defined symbols in
each $\varphi_X(\x)$ with respect to $K^i$ and negative occurrences
with respect to $M$. This process will be monotone.

We first choose $M$ to be $I^{\xi}$: negative occurrences of defined
symbols are interpreted by the lower bound of what can be derived.
Thus, during the derivation process of $\bot_\baseIo, K^1, K^2,\dots$,
we systematically underestimate the truth of negative occurrences of
defined predicates. Due to the anti-monotone effect of negative
occurrences of defined symbols on what can be derived, in each stage
$K^i$, too many atoms may be derived.  Consequently, the limit of this
derivation process yields an upper bound of what can be derived, and
we take it to be our new upper bound $J^{\xi+1}$. Second, we choose
$M$ to be $J^{\xi}$, our best upper bound so far on what can be
derived. Thus, during the derivation process $\bot_\baseIo, L^1,
L^2,\dots$, we systematically overestimate the truth of negative
occurrences of defined symbols, and in each derivation stage $K^i$,
too few atoms are derived.  Therefore, the limit of this sequence
represents a lower bound of what can be derived and we define it to be
$I^{\xi+1}$.  We have constructed our new approximating pair
$(I^{\xi+1},J^{\xi+1})$, by two monotone inductions.

It is now easy to understand in what sense the above construction
follows the natural dependency order between domain atoms, induced by
the rules of a definition. Assume that at some stage $(I^\xi,J^\xi)$,
the truth of a domain atom $A$ has not yet been fixed
(i.e. $A^{I^\xi}\neq A^{J^\xi}$), but the truth values of all atoms on
which $A$ depends negatively have been derived. In the fixpoint
computations $\bot_\baseIo= K^0, K^1, K^2,\dots$ with limit
$J^{\xi+1}$ and $\bot_\baseIo= L^0, L^1, L^2,\dots$ leading to
$I^{\xi+1}$, the structures $K^0$ and $L^0$ evidently coincide on all
atoms on which $A$ depends, and this property is preserved during the
induction, since the structures $I^\xi$ and $J^\xi$ which are used to
evaluate negative occurrences of defined symbols, coincide on all
atoms on which $A$ depends negatively. Therefore, the new lower and
upperbounds $I^{\xi+1}$ and $J^{\xi+1}$ will coincide also on the
value of $A$. Consequently, in this step the truth value of $A$ is
obtained.

\vspace{5mm}

Now, we will formalise the above concepts.  Let $\D$ be a definition
over vocabulary $\tau$ ($\free{\D}\subseteq \tau$). The basis of the
construction of the well-founded model is an operator $\TPp_{\D}$
mapping pairs of $\tau$-structures to $\tau$-structures. Given such a
pair $(I,J)$ , the operator $\TPp_{\D}$ operates like $\G_{\D}$, but
evaluates the bodies of the rules in a different way. In particular, it
evaluates positive occurrences of defined symbols in rule bodies by
$I$, and negative occurrences of defined symbols by $J$.

To formally define this operator, we simply rename the negative
occurrences in rule bodies of $\D$.  We extend the vocabulary $\tau$
with, for each defined symbol $X$, a new relation symbol $X'$ of the
same arity. The extended vocabulary $\tau\cup\X'$ will be denoted
$\tau'$.  Then in each rule body in $\D$, we substitute the symbol
$X'$ for each negative occurrence of a defined symbol $X$, thus
obtaining a new definition $\D'$. For example, given the following
definition
$$
\D:= \left\{
\begin{array}{l}
\forall x \forall y \ (P(x) \la S(x,y,z)  \wedge \neg P(y)),\\
\forall x \forall y \ (P(x) \la  \neg Q(x,y,z)\wedge  P(y))
\end{array}
\right\},
$$
we rename selected occurrences of $P$ by $P'$, as described above, and obtain
$$
\D':= \left\{ 
\begin{array}{l}
\forall x \forall y \ (P(x) \la S(x,y,z) \wedge \neg P'(y)) ,\\
\forall x \forall y \ (P(x) \la  \neg Q(x,y,z)\wedge  P(y))
\end{array}
\right\}.
$$
The definition of $\D'$ defines the same predicates as $\D$ and its
open symbols are those of $\D$ augmented with the new primed
predicates $\X'$. Moreover, a defined symbol $X$ has only positive
occurrences and a primed symbol $X'$ only negative occurrences in rule
bodies of $\D'$. Thus, $\D'$ is a positive definition over the
vocabulary $\tau'$. As described in the formula (\ref{eq-varphi}),
with each defined symbol $X$, we construct the formula $\varphi_X'$
using $\D'$ instead of $\D$.  $\varphi_X'$ can  be obtained also from
$\varphi_{X}$ by substituting $Y'$ for $Y$ in all negative occurrences
of all defined symbols $Y$ in $\varphi_{X}$.

\ignore{ I FOUND THIS REDUNDANT/

$\D'$ defines an operator $\G_{\D'}$ on $\tau'$-structures. As described
in the formula (\ref{eq-varphi}), with each defined symbol $X$, we
construct the formula $\varphi_X'$ using $\D'$ instead of $\D$.
$\varphi_X'$ can be obtained also from $\varphi_{X}$ by substituting 
$Y'$ for $Y$ in all negative occurrences of all defined symbols $Y$ in
$\varphi_{X}$. For example, for the definition above, we have:
$$
\varphi'_P(x) := \exists y \ (S(x,y,z) \wedge \neg P'(y) \vee
\neg Q(x,y,z)\wedge P(y)).
$$
}

For any pair of $\tau$-structures $I,J$ which share the same domain,
define $\Tra{I}{J}$ as the $\tau'$-structure $J[\X:\X^I,\X':\X^J]$.
This $\Tra{I}{J}$ is a $\tau'$-structure which satisfies the
following:
\begin{itemize}
\item its domain is the same as the domain of $I$ and $J$,
\item each open symbol of $\D$ is interpreted by $J$,
\item each defined symbol of $\D$ is interpreted by $I$,
\item the value of each new symbol $X'$ is $X^J$, the value of $X$ in $J$.
\end{itemize}
It is clear that for some defined symbol $X$, evaluating $\varphi_X'$
under $\Tra{I}{J}$ simulates the non-standard evaluation of
$\varphi_X$ where $J$ is ``responsible'' for the open and the negative
occurrences of the defined predicates, while $I$ is ``responsible''
for the positive ones.

Let $\D$ be a definition over some vocabulary $\tau$.
\begin{definition}[operator $\TPp_\D$] 

  We introduce a partially defined binary operator $\TPp_\D \ :\ {\cal
    I} \times {\cal I} \mapsto {\cal I} $, where ${\cal I}$ is the
  class of all $\tau$-structures.  The operator is defined on pairs of
  structures which share the same domain, and is undefined
  otherwise. We have $I'=\TPp_\D(I,J)$ iff
\begin{itemize}
\item $dom(I') = dom (J)= dom(I)$,
\item  for each open symbol $\symb$, $\symb^{I'}:=\symb^{J}$ and
\item  for each defined symbol $X \in\defp{\D}$, 
$$
X^{I'}\   :=\  \{ \ba \ |\  \Tra{I}{J}\models \varphi_{X_i}'[\ba] \},
$$
where formula $\varphi'_{X}$ is defined by equation
(\ref{eq-varphi}) applied to $\D'$.
\end{itemize}
\end{definition}
This definition is equivalent to defining $\TPp_\D(I,J)
:= \G_{\D'}(\Tra{I}{J})|_\tau$, for any pair of
$\tau$-structures $I, J$ such that $\dom{I} = \dom{J}$.

\begin{proposition}\label{prop:mon:antimon} Let $\baseIo$ be a fixed 
$\topen{\D}$-structure.  In the lattice $\vallat{\baseIo}{\tau}$, the
operator $\TPp_{\D}(I,J)$ is monotone in its first argument, and
anti-monotone in its second argument.
\end{proposition}
\begin{proof} 
   Select arbitrary $\tau$-structures $I, I', J, J'$ with the same
  domain such that $J|_\toD = J'|_\toD$, $I \sq I'$ and $J' \sq
  J$. We need to show that $\TPp_{\D}(I,J) \sq\TPp_{\D}(I',J')$.
 Let $L = \TPp_{\D}(I,J)$ and $L' = \TPp_{\D}(I',J')$. 
  
  It holds that $L$ and $L'$ have the same domain and that for each
  open symbol $\sigma\in \topen{\D}$, $\sigma^L = \sigma^{J} =
  \sigma^{J'} = \sigma^{L'}$.  So, it suffices to verify that for each
  defined symbol $X$, $X^L \subseteq X^{L'}$. Let $\ba$ be any element
  of $X^L$. It holds that $\Tra{I}{J} \models \varphi'_X[\ba]$.  The
  structure $\Tra{I'}{J'}$ assigns the same value to open symbols in
  $\varphi'_X$, greater value to the defined symbols $\X$ which occur
  positively in $\varphi'_X$,  and lesser value to the defined symbols
  $\X'$ which occur negatively in $\varphi'_X$. Consequently, it holds
  that $\Tra{I'}{J'} \models \varphi'_X[\ba]$. We find that $\ba\in
  X^{L'}$. We obtain our proposition.
\end{proof}

\ignore{Marc: We need following corollary in section 6: }

The next corollary shows a connection between operators 
$\TPp_\D$ and $\G_{\D}$.

\begin{corollary}\label{Cor:approx}
  For any $\tau$-structure $I$, it holds that $\TPp_{\D}(I,I)=
  \G_{\D}(I)$.
\end{corollary}
\begin{proof} Follows immediately from the fact that  $\Tra{I}{I}\models \varphi_{X_i}'[\ba]$ iff $I\models \varphi_{X_i}[\ba]$.
\end{proof}

\ignore{
{\sf MARC: PERHAPS THE NEXT COROLLARY IS NOT REALLY IMPORTANT ENOUGH?
but it generalises the corollary above.}

{\tt EUGENIA: I agree that it is not important for the proofs, 
but  I think it is very important for understanding
the semantics. Keep? Or delete because there is not enough space?- Eugenia}
}

The proposition has another interesting corollary.
\begin{corollary}\label{Cor:approx}
  Let $I, M, J$ be three $\tau$-extensions of $\baseI$ such that $I\sq M
  \sq J$.  Then it holds that $\TPp_{\D}(I,J)\sq \G_{\D}(M) \sq
  \TPp_{\D}(J,I)$.
\end{corollary}
\begin{proof}
Since $I \sq M\sq J$, Proposition
\ref{prop:mon:antimon} entails that $\TPp_{\D}(I,J)\sq 
\TPp_{\D}(M,M)=\G_{\D}(M) \sq \TPp_{\D}(J,I)$.
\end{proof}
This corollary shows that $\TPp_{\D}$ can be used to approximate
$\G_{\D}$ over an interval of structures. Indeed, if $(I,J)$ is
an approximation of $M$ (i.e., $M\in[I,J]$) then the corollary shows
that $(\TPp_{\D}(I,J),\TPp_{\D}(J,I))$ is an approximation of
$\G_{\D}(M)$. We shall elaborate on the approximation process in a
moment.


Let $J$ be a $\tau$-structure, and $\valJopen$ its restriction to
$\topen{\D}$.  The unary operator $\lambda I \ \TPp_{\D}(I,J)$, often
denoted by $\TPp_{\D}(\cdot,J)$, is a monotone operator in the lattice
$\vallat{\valJopen}{\tau}$; and its least fixpoint in this lattice is
computed by
$$
\lfp(\TPp_{\D}(\cdot,J)):= 
\bigsqcup_{\xi} E^{\xi}, \ \ \mbox{where}
$$
$$
E^{\xi} := \TPp_{\D}(E^{<\xi},J), \ \ \mbox{and}\ \ 
E^{<\xi}:= \bigsqcup_{\eta < \xi} E^{\eta}. 
$$
\begin{definition}[stable operator]
  Define the {\em stable operator} $ST_{\D}: {\cal I} \mapsto {\cal
    I}$ as follows:
$$
ST_{\D}(J):= \lfp(\TPp_{\D}(\cdot,J)).
$$
\end{definition}

The operator $\TPp_{\D}(I,J)$ performs one derivation step by
interpreting positive occurrences of defined symbols by $I$ and
negative occurrences by $J$. The stable operator performs a monotone
induction during which negative occurrences of defined predicates $X$
in $\D$ are interpreted by the fixed value $X^J$.

\begin{example}\label{Ex:stable}
  We illustrate the stable operator with the definition of Example
  \ref{Ex:prop1}:
$$
\D_0=\defin{P \rul \Tr,\\ Q \rul\neg P, \\ Q \rul Q}.
$$ 
This definition has no open symbols and is equivalent to the following
definition:
$$
\defin{P \la \Tr ,\\
   Q \la \neg P \lor Q  }.
$$
It is straightforward to see that in a propositional definition, the
mapping $ST_{\D_0}(J)$ for any $J$ is the least fixpoint of the
positive definition obtained by substituting each negative occurrence
of a defined symbol and each occurrence of an open symbol by its truth
value in $J$. Thus, the stable operator maps the empty structure
$\emptyset$ to the least fixpoint of the definition:
$$
\defin{P \rul \Tr ,\\ Q \rul\neg \Fa \lor Q }.
$$
This yields the structure $\{P,Q\}$.

Similarly, the stable operator maps the structure $\{P,Q\}$ to the
least fixpoint of the definition:
$$
\defin{P \la \Tr ,\\
   Q \la \neg \Tr \lor Q  }.
$$
This yields the structure $\{P\}$. Likewise, the stable operator maps
 $\{P\}$ to the least fixpoint of the same definition, and this yields
 $\{P\}$ itself.
\end{example}

\ignore{
\marc{CHECK: OLD: {\tt Let $\baseI$ be a  structure of the 
open symbols of $\D$ in $\tau$.} NEW: Let $\baseI$ be a
fixed structure such that $\tfun \subseteq \tau_{\baseI}
\subseteq \topen{\D}$.  } }

\ignore{ THE FOLLOWING PROPOSITION CONTAINS AN ERROR: MY GENERALISATION WAS WRONG 
\begin{proposition} Let $\baseIo$ be a
fixed structure such that $\tfun \subseteq \tau_{\baseI}
\subseteq \topen{\D}$. 

Operator $ST_{\D}$ is anti-monotone on $\vallat{\baseI}{\tau}$. 
\end{proposition}
\begin{proof} 
Let $I\leqt J$ be extensions of $K$. To show that $ST_{\D}$ is
anti-monotone, it suffices to show that any pre-fixpoint of
$\TPp_{\D}(\cdot ,I)$ is a pre-fixpoint of $\TPp_{\D}(\cdot ,J)$. Assume that
$\TPp_{\D}(J',I)\leqt J'$. Then by antimonotonicity of $\TPp_{\D}$ in
the second argument, $\TPp_{\D}(J',J)\leqt \TPp_{\D}(J',I)\leqt
J'$. Thus $J'$ is a pre-fixpoint of $\TPp_{\D}(\cdot ,J)$ and this entails
that the least (pre-)fixpoint of $\TPp_{\D}(\cdot ,J)$ is less than the
least (pre-)fixpoint of $\TPp_{\D}(\cdot ,I)$.
\end{proof}
}

\begin{proposition} Let $\baseIo$ be a fixed $\toD$-structure.
Operator $ST_{\D}$ is anti-monotone on $\vallat{\baseIo}{\tau}$. 
\end{proposition}
\begin{proof} 
Let $I\leqt J$ be $\tau$-extensions of $\baseIo$. To show that
$ST_{\D}$ is anti-monotone, it suffices to show that any pre-fixpoint
of $\TPp_{\D}(\cdot,I)$ is a pre-fixpoint of $\TPp_{\D}(\cdot ,J)$.
It will follow then that $ST_{\D}(J)$, the least pre-fixpoint of
$\TPp_{\D}(\cdot ,J)$, is smaller than $ST_{\D}(I)$, the least
pre-fixpoint of $\TPp_{\D}(\cdot,I)$.  


Assume that
for any $J'\in \vallat{\baseIo}{\tau}$, $\TPp_{\D}(J',I)\leqt
J'$. Then, by anti-monotonicity of $\TPp_{\D}$ in the second argument
(Proposition \ref{prop:mon:antimon}), $\TPp_{\D}(J',J)\leqt
\TPp_{\D}(J',I)\leqt J'$. Thus $J'$ is a pre-fixpoint of
$\TPp_{\D}(\cdot ,J)$ and this entails that the least (pre-)fixpoint of
$\TPp_{\D}(\cdot ,J)$ is less than the least (pre-)fixpoint of
$\TPp_{\D}(\cdot ,I)$.
\end{proof}

Fix some $\toD$-structure $\baseIo$ with domain $A$ of the open
symbols of $\D$ in $\tau$.

As is standard for anti-monotone operators on a complete lattice (see
Section \ref{preliminaries-lattices}), the operator $ST_{\D}$ gives
rise to a sequence $(I^{\xi},J^{\xi})_{\xi \geq 0}$ in
$\vallat{\baseIo}{\tau}$ defined by
$$
\begin{array}{l}
I^{\xi}:= ST_{\D}(J^{<\xi}),\ \ \mbox{where} \ \
J^{<\xi}:= \sqcap_{\eta<\xi}J^{\eta},\\
J^{\xi}:= ST_{\D}(I^{<\xi}), \ \ \mbox{where} \ \ 
I^{<\xi}:= \sqcup_{\eta<\xi}I^{\eta}.\\
\end{array}
$$
Notice that $I^{<0}$ is, by definition, the bottom element
$\bot_\baseIo$ of the lattice, i.e., the structure which assigns
$\emptyset$ to every defined symbol; and $J^{<0}$ is the top element
$\top_\baseIo$ which assigns Cartesian product $A^{r}$ to each
$r$-ary defined symbol $X$ of $\D$.

The anti-monotonicity of $\G_\D$ implies that the sequence
$(I^{\xi})_{\xi \geq 0}$ is increasing and $(J^{\xi})_{\xi \geq 0}$ is
decreasing. Moreover, for each $\xi$, $I^{\xi} \leqt J^{\xi}$. Thus,
it holds that the sequence $(I^{\xi},J^{\xi})_{\xi \geq 0}$ is indeed
a sequence of increasingly precise approximations. This sequence has a
limit $(I,J)$, which is the maximal oscillating pair of $ST_{\D}$.
Equivalently, $I$ and $J$ are fixpoints of the square $ST^2_{\D}$,
$\lfp(ST^2_{\D})$ and $\gfp(ST^2_{\D})$, respectively.

In the lattice $\vallat{\baseIo}{\tau}$, we define 
$$
\lb{\baseIo}{\D} :=\lfp(ST^2_{\D}) ,\ \  \mbox{and} \ \
\ub{\baseIo}{\D}:=\gfp(ST^2_{\D}).
$$
 We extend this notation to any structure
$L$ which interprets at least $\toD$ and define 
$$\lb{L}{\D}:=
\lb{(L|_{\toD})}{\D}, \ \  and \ \ \ub{L}{\D}:= \ub{(L|_{\toD})}{\D}.
$$
 Note
that $\lb{L}{\D}$ and $\ub{L}{\D}$ agree with $L$ on the open symbols
but not necessarily on the defined symbols.


\ignore{ I GENERALISED THE FOLLOWING DEFINITION:
\begin{definition}[total definition]        
Definition $\D$ is {\em total } in $\toD$-structure $\baseIo$ if
$\lb{\baseIo}{\D}=\ub{\baseIo}{\D}$. If $\tfun \subseteq \tauI{\baseI}
\subseteq \toD$, we say that $\D$ is total in $K$ if $\D$ is total in
each $\toD$-structure extending $K$.  If $\toD\subseteq \tauI{K}
\subseteq \tau$, then we define that $\D$ is total in $K$ if $\D$ is total in
$K|_{\toD}$. 
We say that a definition $\D$ is {\em total } if it is total in each
$\toD$-structure $\baseIo$.
\end{definition}
}

\begin{definition}[total definition]        
Definition $\D$ is {\em total } in $\toD$-structure $\baseIo$ if
$\lb{\baseIo}{\D}=\ub{\baseIo}{\D}$. If $\tauI{K} \subseteq \toD$, we
say that $\D$ is {\em total in $K$} if $\D$ is total
in each $\toD$-structure extending $K$.  If $\toD\subseteq
\tauI{K} \subseteq \tau$, then we say that $\D$ is {\em total in $K$} if
$\D$ is total in $K|_{\toD}$.  We say that a definition $\D$ is {\em
total } if it is total in each $\toD$-structure $\baseIo$.
\end{definition}

\ignore{
The above definition has a natural extension to structures of
arbitrary vocabularies $\tau'\subseteq\tau$.  We say that $\D$ is
total in a $\tau'$-structure $K$ if $\D$ is total in each
$\toD$-structure $\valIopen$ such that
$\valIopen|_{\tau'}=K|_{\toD}$. In particular, this means that for
every structure $K$ interpreting all open symbols, $\D$ is total if
$\lb{K}{\D}=\ub{K}{\D}$. If $\tau'\subseteq\toD$, then $\D$ is total
in $K$ if $\D$ is total in each $\toD$-structure extending $K$.}

The aim of a definition is to {\em define} its defined
symbols. Therefore, a natural quality requirement for a definition is
that it is total. 
\begin{definition}[$I$ satisfies $\D$]
We say that $\tau$-structure $I$  {\em satisfies
$\D$}, or equivalently, that $\D$ is {\em true in $I$} (denoted $I\models
\D$) if $\D$ is total in $I$ and $\lb{I}{\D}$ is identical to $I$.
\end{definition}

Let $I$ be any structure such that $\toD \subseteq \tauI{I}\subseteq
\tau$.

\begin{definition}[$\D$-extension of $I$]\label{D-extension}
  Let  $\D$ be total in $I$. Define the {\em
    $\D$-extension of $I$}, denoted $I^\D$, as $I^\D:=\lb{I}{\D}$ (or,
    equivalently, $I^\D:=\ub{I}{\D}$). If $\D$ is not total in $I$,
    then $I$ has no $\D$-extension.
\end{definition}
\noindent  Note that for any $\toD$-structure $\baseIo$, there is at most one
$\D$-extension extending $\baseIo$.

\begin{example}
We illustrate the iterative process described above with the
definition of Example \ref{Ex:prop1}, which is equivalent to the
following definition
$$
\defin{P \la \Tr ,\\
   Q \la \neg P \lor Q  }.
$$
The first pair in this sequence is the least precise pair that
approximates all structures:
$$
\begin{array}{rll}
              \ I^{<0} &:=& \emptyset ,\\
                 J^{<0}&:=& \{P,Q\} .
                 \end{array}
$$
To compute the new upper bound $J^0$, we apply the
stable operator on $I^{<0}$ which yields, as shown in Example
\ref{Ex:stable}, $\{P,Q\}$. To compute $I^0$, the stable operator 
is applied on $J^{<0} = \{P,Q\}$ which yields $\{P\}$. Note that at
this moment, $I^0$ and $J^0$ agree on the fact that $P$ is true. So,
after this first step, we have derived that $P$ is true.

In the next step, we obtain $I^2 = \{P\} = J^2$. We derived that $Q$
is false. The next iteration produces exactly the same pair, so, we
obtained a fixpoint with identical lower and upper bound $\{P\}$. The
definition $\D_0$ is total. Since there are no open symbols, $\{P\}$
is the unique $\D_0$-extension. It coincides with the structure that
we obtained in Example \ref{Ex:prop1} by applying iterated induction.
\end{example} 

\begin{example}\label{ExNMEven}
Consider the definition:
$$
\D_{even} := \defin{\forall x \ (E(x) \rul x=0) ,\\\forall x \ (E(s(x))\rul \neg E(x))},
$$
which  is equivalent to: 
$$
\defin{\forall x \ [E(y) \rul y=0 \lor \exists x (y=s(x) \land \neg E(x))]}.
$$
We show that in the extension of $\D_{even}$ in the natural numbers,
$E$ is interpreted by the set of even numbers.  Note that this
definition has no positive occurrences of the defined predicate
$E$. Therefore, $\TPp_{\D_{even}}(I,J) = \Gamma_{\D_{even}}(J)$, for
all $I, J$ sharing the same domain.

The well-founded model computation starts in the least precise pair
extending the natural numbers:
$$
\begin{array}{rll}
              \ E^{I^{<0}} &:=& \emptyset, \\
                E^{J^{<0}} &:=& \natnr .
                 \end{array}
$$
To compute the new upper bound $J^0$, we apply $\Gamma_{\D_{even}}$ on
$I^{<0}$. Since $I^{<0}$ satisfies the body of the rule for each
natural number, we obtain the set $\natnr$ as a new upper bound. As a
new lower bound, we derive the singleton $\{0\}$. At this point, we derived
that $0$ is even.
$$
\begin{array}{rll}
              \ E^{I^{0}} &:=& \{0\} ,\\
                E^{J^{0}} &:=& \natnr .
                 \end{array}
$$
In the next step, since the upper bound did not change, we derive the
same lower bound. When computing the new upper bound, we obtain all natural
numbers except 1. This means that we derived that 1 is not even:
$$
\begin{array}{rll}
              \ E^{I^{1}} &:=& \{0\} ,\\
                E^{J^{1}} &:=& \natnr \setminus \{1\}.
                 \end{array}
$$
In the third step, the upper bound remains unaltered. With respect to
the lower bound, we can now derive both 0 and 2:
$$
\begin{array}{rll}
              \ E^{I^{2}} &:=& \{0,2 \} ,\\
                E^{J^{2}} &:=& \natnr \setminus \{1\}.
                 \end{array}
$$
In the subsequent step, we obtain the same lower bound, but 3 is
eliminated from the upper bound. $$
\begin{array}{rll}
              \ E^{I^{3}} &:=& \{0,2 \} ,\\
                E^{J^{2}} &:=& \natnr \setminus \{1,3\}.
                 \end{array}
$$
After iterating this process $\omega$ steps, we obtain the fixpoint:
$$E^{I^{\omega}} = E^{J^{\omega}} = \{ 2n\  | \ n\in\natnr \}.$$
 \end{example} 

Now we are ready to define the satisfaction relation between
structures and well-formed formulas of the logic. 

  


\begin{definition}[$\phi$ true in structure $I$]\label{Defmodel}
 Let $\phi$ be a  ID-formula and  $\valI$ any structure such that $\free{\phi}\subseteq \tauI{\valI}$. 
  
  We define $\valI\models\phi$ (in words, {\em $\phi$ is true in
    $\valI$}, or {\em $\valI$ satisfies $\phi$}, or $\valI$ is a {\em  model}
    of $\phi$) by the following induction:
\begin{itemize}
\item  $\valI\models X(t_1,\dots,t_n)$ if
$(\eval{t_1}{\valI},\dots,\eval{t_n}{\valI}) \in X^\valI$;
\item$\valI \models \psi_1\land \psi_2$ if $\valI \models \psi_1$ and
$\valI \models \psi_2$;
\item$\valI \models\neg \psi$ if $\valI \not\models \psi$;
\item$\valI \models\exists \symb \ \psi$ if for some value $v$
of $\symb$ in the domain $\domI$ of $\valI$,
$\valI[\symb:v] \models\psi$;
\item$\valI \models \D$ if $I = \lb{I}{\D} =\ub{I}{\D}$. 
\end{itemize}

Given an ID-theory $T$ over $\tau$, a $\tau$-structure $I$ satisfies
$T$ (is a model of $T$) if $I$ satisfies each $\phi\in T$. This is
denoted by $I\models T$.
\end{definition}

\begin{example}\label{Ex:nat2} Consider the theory $T_{\mathbb{N}}$ 
of Example \ref{Peano}.  We prove that each model $I$ of
$T_{\mathbb{N}}$ is isomorphic to the structure of the natural
numbers. Let $I$ be a model of this theory. First, since $I$ satisfies
the first-order Peano axioms, the domain elements
$0^I,s(0)^I,\dots ,s^n(0)^I,\dots$ are pair-wise distinct and the set of
these domain elements constitutes a subset of $\dom{I}$, isomorphic
to the natural numbers. Therefore, it suffices to show that this set
is exactly the domain of $I$. Since $I$ satisfies the ID-axiom
replacing the induction axiom, there exists a set $S\subseteq \dom{I}$
such that $I[N:S]$ satisfies 
$$
\defin{\forall x \ (N(x) \rul x=0) , \\ \forall x \ (N(s(x)) \rul N(x))}
\land \forall x \ N(x).
$$
Since $I[N:S]$ satisfies $\forall x \ N(x)$, $S$ must be $\dom{I}$.  As
proven later in Theorem \ref{theorem-Mon}, $I[N:S]$ satisfies the
positive definition in this axiom iff $S$ is the least set containing
$0^I$ and closed under $s^I$. Hence, $\dom{I}$ is exactly the set
$\{0^I,s(0)^I,\dots ,s^n(0)^I,\dots \}$.
\end{example}

\begin{example} \label{Ex:MultipleDef}
An ID-theory can contain multiple definitions for the
  same predicate. A simple illustration is when  a natural class is
  partitioned in subclasses in different ways, depending 
  on the property used. 
For
  example, humans can be partitioned in males and females, but also in
  adults and children, etc.. This is modeled by the following formula:
  $$
  \defin{ \forall x \ (Human(x) \rul Male(x)),\\ \forall x \ (Human(x) \rul  Female(x))} \land \defin{
    \forall x \ (Human(x) \rul Adult(x)),\\ \forall x \ (Human(x) \rul Child(x))}.
  $$
  This formula implies that the class humans is the union of the
  classes males and females, and also of the classes adults and
  children. The definition  
  $$
  \defin{ \forall x \ (Human(x) \rul Male(x) \lor Female(x)),\\ 
          \forall x \ (Human(x) \rul Adult(x) \lor Child(x))}.
  $$
  is weaker, in the sense that it does not entail that humans are
  either males or females.
\end{example}

\subsection{Total Definitions}
\label{SSecTotalD}

Totality of non-monotone definitions is a fundamental property in our
theory of non-monotone induction. In particular, it indicates that a
definition is well-constructed, i.e., does not produce undefined
atoms.  

\begin{example} Consider the following definition:
$$
\D_2 := \defin{P \rul \neg P}.
$$
One verifies that the iterated induction yields the limit
$(\emptyset,\{P\})$. This definition has no model. 
\end{example}
\begin{example}
Consider the definition:
$$
\D_3 := \defin{P \rul  \neg Q,\\ Q \rul \neg P}.
$$
The iterated induction yields the limit $(\emptyset,\{P,Q\})$. The definition has no model. 
\end{example}

The next example shows that a (useful) definition which is total in one
structure, may not be total in other structures.
\begin{example} \label{ExNMEven1}
Consider the definition of Example \ref{ExNMEven}:
$$
\D_{even} := \defin{\forall x \ (E(x) \rul x=0) ,\\\forall x \ (E(s(x))\rul \neg E(x))}.
$$
Recall that the stable operator of this definition is identical to
$\Gamma_{\D_{even}}$.  In Example \ref{ExNMEven}, we showed that this
definition is total in the structure of the natural numbers.  It is
not total in many other structures, in particular in those where the
successor function contains cycles or infinite descending chains. 
For example, $\D_{even}$ is not total in the structure $\baseIo$
with domain $\{0,1\}$, and $s^{\baseIo}(0)=1,s^{\baseIo}(1)=1$.
In this structure, the maximal oscillating pair $(I,J)$ of
$\Gamma_{\D_{even}}$ interprets $E$ as follows:
$$
\begin{array}{rll}
              \ E^{I} &:=& \{0 \} ,\\
                E^{J} &:=&  \{0,1\}.
                 \end{array}
$$
The reason for this oscillation is that in this structure, atom  $E[1]$
depends on $\neg E[1]$.

Definition $\D_{even}$ is not total either in the structure
$\baseIo'$ with domain $\integers$ and $s^{\baseIo'}$ the standard
successor function on $\integers$. In this structure, the maximal
oscillating pair $(I,J)$ of $\Gamma_{\D_{even}}$ interprets $E$ as
follows:
$$
\begin{array}{rll}
              \ E^{I} &:=& \{2n \ | \ n\in \natnr \}, \\
                E^{J} &:=& \{n\  | \ n<0\} \cup \{2n | n\in \natnr \}.
                 \end{array}
$$
\end{example}

\begin{example} \label{ExNat3}
Recall the theory $T_{\mathbb{N}}$ of Example \ref{Peano},
$$
\begin{array}{c}
\exists N \left [ \defin{ \forall x \ (N(x) \rul  x=0), \\ 
\forall x\ ( N(s(x)) \rul N(x)}
\land \forall x \ N(x))\right ],\\
  \forall n \ \neg (s(n) = 0) ,\\ 
  \forall n \forall m\ ( s(n)=s(m) \ra  n=m ).
\end{array}
$$
and the definition of Example \ref{ExNMEven},
$$
\D_{even} :=  \defin{\forall x \ (E(x) \rul x=0) ,\\
			\forall x\ (E(s(x))\rul \neg E(x))}.
$$
In Example \ref{Ex:nat2}, we saw that the natural numbers are the
unique model of $T_{\mathbb{N}}$ (modulo isomorphism). In Example
\ref{ExNMEven1}, we saw that $\D_{even}$ is total in the
natural numbers. Consequently, $\D_{even}$ is total in
$T_{\mathbb{N}}$. The theory $T_{\mathbb{N}}\cup \{\D_{even}\}$ is
consistent and has one model,  the natural numbers and $E$ interpreted
by the even numbers.
\end{example} 

\ignore{The cause of the non-totality in these examples is clear. The natural
dependency order, induced by the rules, either contains cycles or
contains infinite descending chains in which atoms depend negatively
on other atoms.  {\tt ABOVE: NOT CLEAR. NEED TO EXPLAIN WHY THIS LEADS
TO NON-TOTALITY } NEW: }  What is the cause of the non-totality of a
definition?  In the above examples, the natural dependency order,
induced by the rules, contains infinite descending chains in which
atoms depend negatively on the same or other atoms. When this happens,
the stable operator oscillates between a structure in which all atoms
of the chain are false and one in which  these atoms  are true. 

When $\D$ is not total in $\baseIo$, Definition \ref{Defmodel}
states that there is no model that extends $\baseIo$. To cope with
such cases, we might adopt an alternative definition of $\D$-extension
and define the $\D$-extension of $\baseIo$ as a 3-valued
structure. With $(\lb{I}{\D},\ub{I}{\D})$, a unique three-valued
structure corresponds which coincides with $\lb{I}{\D}$ and
$\ub{I}{\D}$ on all atoms where $\lb{I}{\D}$ and $\ub{I}{\D}$ agree
and is undefined on all atoms where $\lb{I}{\D}$ and $\ub{I}{\D}$
disagree.  This is the option that has been taken in the original
well-founded semantics of logic programming.  In this paper, we will
stick to a 2-valued solution and avoid the complexities caused by
using three-valued logic.

Notice that we cannot restrict the syntax of the logic to allow total
definitions only.  Such a restriction would lead to undecidable syntax
--- there would be no procedure which would decide, for a given
formula $\phi$, whether $\phi$ is a well-formed formula of the
language.  This is because the problem of determining, for a given
definition $\D$ and structure $I_o$, whether $\D$ is total in $I_o$,
is undecidable \cite{Schlipf:1995a}. 

For important  classes of definitions, it is known that they are total.
For example, positive definitions are total in any structure. For
other types of definitions, techniques must be developed to prove 
that they are total. In this paper, we develop such techniques.

\ignore{

HERE WAS A PROPOSITION ABOUT TOTALITY OF AN OPERATOR AND UNIQUE FIXPOINTS OF
THE SQUARE OF THE STABLE OPERATOR. IT WAS NEVER USED. I PROPOSE TO DROP IT.

The following proposition illustrates some properties of total
definitions.
\begin{proposition}\label{proposition-x}
  Definition $\D$ is total in $\toD$-structure $\valIopen$ if
  operator $ST^2_{\D}$ has a unique fixpoint in the lattice of
  structures extending $\valIopen$.
\end{proposition}
\begin{proof}
  By definition of a total definition, $\lfp(ST^2_\D)=\gfp(ST^2_\D)$.
  Operator $ST^2_\D$ is a monotone operator in the lattice.  Thus, by
  the Knaster-Tarski theorem, it has a complete lattice of fixed
  points.  The greatest and the least fixpoints of a monotone operator
  on a complete lattice coincide iff all its fixed points are equal.
  Therefore, $\D$ is total in $\valIopen$ iff $ST^2_\D$ has
  a unique fixpoint in its lattice.
\end{proof}
}

\section{Reduction Relations}\label{Sec:reductions}
 

\newcommand{\dep}{\prec}
\newcommand{\sdep}{\prec \!\!\!\prec}
\newcommand{\islePa}{\cong_{\dep\Pa}}
\newcommand{\islPa}{\cong_{\sdep\Pa}}
\newcommand{\vallePa}[1]{{\cal S}_{#1}^{\dep\Pa}}

\newcommand{\homoPa}{|\cdot|_{\dep\Pa}}
\newcommand{\homoPaa}[1]{|#1|_{\dep\Pa}}
\newcommand{\homoPal}[1]{|#1|_{\sdep\Pa}}

In Section \ref{ssec-preview-ID}, we mentioned that a definition
implicitly induces a dependency relation between atoms and that the
well-founded semantics performs iterated induction along this
dependency relation, in the sense that the truth assignment to an atom
is delayed until enough information about the atoms on which it
depends has become available. This shows that the notion of dependency
relation induced by a definition is important. In this section, we
formalise this intuitive concept by the notion of {\em reduction
  relation}. Intuitively, a reduction relation $\dep$ is a binary
relation between domain atoms such that for each defined atom $\Pa$,
the truth of its defining formula $\varphi'_P[\bar{a}]$ depends only
on the truth of atoms $\Qb \dep \Pa$. In the next paragraphs we
formalise what this means.

\ignore{

An ID defines certain atoms in terms of other atoms
and in this way induces a dependency relation between atoms. This
dependenc relation is the core of the internal structure of the ID.
For example, to have a definition over a WF orde, this dependency
relation needs to be a WF (pre-)order. The WF oder construction delays
the truth assignment to an atom until enough information about the
atoms on whichit depends has become available. Also, if two structures
$\baseIo$ and $\baseJo$ of the open symbols agree on all open atoms
$\Qb$ on which a defined atom $\Pa$ depends, the models $\baseIo^\D$
and $\baseJo^\D$ will agree on the truth value of $\Pa$.  For this
reason, this section studies the dependency relation and formalises it
by the notion of a reduction relation.

}


Let $\tau$ be a vocabulary and $A$ a domain. Recall that
$\Att$ denotes the set of domain atoms over vocabulary $\tau$ in
domain $A$. Let $\dep$ be any binary relation on $\Att$. If $\Qb
\dep \Pa$, we will say that $\Pa$ depends on $\Qb$ (according to
$\dep$). The binary relation $\sdep$ is derived from $\dep$ in the
following way: $\Qb \sdep \Pa$ iff $\Qb \dep \Pa \land \Pa \not \dep
\Qb$. Intuitively, $\Pa$ depends on $\Qb$ but not vice versa.

For any domain atom $\Pa \in \Att$, for any structure $I$ with
domain $A$ such that $\tauI{I}\subseteq \tau$, define $\homoPaa{I}$ as the
structure $I[\X:\R]$ where $\X$ is the set of relation symbols in
$\tauI{I}$ and for each relation symbol $X \in \X$, its value is given
by 
$$
R := \{ \bd \ |\ I \models X[\bd]
 \ \ \mbox{and} \ \  X[\bd] \dep \Pa \}.
$$
Intuitively, $\homoPaa{I}$ {\em falsifies } all true atoms $\Qb$ on
which $\Pa$ does not depend. The operation $\homoPaa{\cdot}$ is an
idempotent operation, that is $\homoPaa{(\homoPaa{I})}=\homoPaa{I}$.

For any pair $I, J$ of structures with domain $A$, we define $I
\islePa J$ if $\homoPaa{I} = \homoPaa{J}$.  When $I \islePa J$, then
$I$ and $J$ interpret the same symbols, assign the same value to all
function symbols, and assign the same value to all domain atoms $\Qb
\dep \Pa$. We extend this relation to tuples and define $(I,J) \islePa
(I',J')$ if $I \islePa I'$ and $J \islePa J'$. Intuitively, $(I,J)
\islePa (I',J')$ means that $(I,J)$ and $(I',J')$ are identical on all
atoms on which $\Pa$ depends.

Recall from Section \ref{Sec:ID-logic} that for any defined symbol
$P$ of $\D$, $\varphi'_P(\x)$ is obtained by renaming negative
occurrences of defined symbols in $\varphi_P(\x)$.  For each pair of
$\tau$-structures $I, J$ with domain $A$, the associated
$\tau'$-structure $\Tra{I}{J}$ is the structure $J[\X:\X^I,\X':\X^J]$
where $\X$ is the collection of defined symbols of $\D$.

Assume a definition $\D$ over $\tau$ and a structure $\baseI$ with
domain $A$ such that $\tauI{\baseI}\subseteq \toD$. 

\begin{definition}[reduction relation] \label{DefReduction}
  A binary relation $\dep$ on $\Att$ is a {\em reduction
    relation} (or briefly, a reduction) of $\D$ in $\baseI$ if for
    each domain atom $\Pa$ with $P$ a defined symbol, for all
    $\tau$-structures $I, J, I', J'$ extending $\baseI$, if $(I,J)
    \islePa (I',J')$ then $\Tra{I}{J} \models \varphi'_P[\bar{a}]
    \mbox{ iff } \Tra{I'}{J'} \models \varphi'_P[\bar{a}]$.
\end{definition}

\begin{example}\label{D3}
For the following definition
$$
\D := \defin{ \forall x\ (E(x)\la x=0),\\ 
        \forall x\ (E(s(x)) \la O(x)),\\ 
       \forall x \ ( O(s(x))\la E(x)) } ,
$$
a reduction $\dep$ in the structure of the natural numbers is the
relation represented by the set of tuples:
$$
  \{ (E[n] , O[n+1]), (O[n], E[n+1]) \ |\  n\in \natnr\}. 
$$
Also its transitive closure $\dep^*$ is a reduction. 

\noindent
It can be easily verified that, e.g. for an atom $E[n+1]$, if structures
$I, J$ agree on the atom $O[n] \dep E[n+1]$, then $\Gamma_{\D}(I)$ and
$\Gamma_{\D}(J)$ will agree on the value of $E[n+1]$.
\end{example}

\begin{example} 
Reduction relations are context dependent. Consider the following
propositional definition:
$$
\D := \defin{  P \rul Q\land R,\\ Q\rul P\land \neg R}.
$$
The relation $\dep_1 := \{ (Q,P),(R,Q),(R,P) \}$ is a reduction
relation of $\D$ in the $\toD$-structure $\{ R\}$ but not in the
$\toD$-structure $\emptyset$. Vice versa, the relation $\dep_2 := \{
(P,Q),(R,Q),(R,P) \}$ is a reduction relation of $\D$ in $\emptyset$
but not in $\{ R \}$.
\end{example}

Let $\dep$ be a reduction of $\D$ in $\baseI$. 
\begin{proposition} \label{PropSupReduction} If $\baseI'$ extends $\baseI$,
then $\dep$ is a reduction of $\D$ in $\baseI'$.  
\end{proposition}

\begin{proposition} \label{PropSupReductiona} Any superset $\dep'$
  of $\dep$ is a reduction relation of $\D$ in $\baseI$.
\end{proposition}
In particular, the transitive closure and the reflexive transitive
closure of a reduction are reductions. This stems from the fact that
$I\cong_{\dep'\Pa}J$ implies $I\islePa J$.

As shown by this proposition, a definition $\D$ may have many
reduction relations in $\baseI$. The total binary relation $\dep_t =
\Att \times \Att$ is always a reduction relation. Since $I
\cong_{\dep_t\Pa} J$ iff $I=J$, the relation $\dep_t$ trivially satisfies
Definition \ref{DefReduction}. It can be seen here that a reduction
relation in general overestimates the dependencies between domain
atoms in a definition.  Only the least reduction relation
of a definition reflects the true dependencies. However, as shown in
the next example, some definitions do not have a least reduction
relation.

\begin{example}
Consider the following definition in the context of the natural numbers:
$$
\D := \defin{  P \rul \exists n \forall m (m>n \supset Q(m)) }.
$$
The predicate $Q$ is open in this definition. This definition
defines $P$ to be true if there exists a number $n$ such that $Q$
contains at least all natural numbers larger than $n$. It can easily be
verified that for each $n\in \natnr$, the relation
$$
\dep_n = \{ (Q(m),P) | m > n\} 
$$
is a reduction relation of $\D$ in $\natnr$.  The intersection of
these relations is $\emptyset$, and this is not a reduction relation
of $\D$.
\end{example}

We defined the operator $\TPp_\D$ as a map from pairs $I, J$ of
$\tau$-structures with shared domain to the interpretation $J'$
extending $J|_{\toD}$ such that for each defined atom $\Pa$,
$\Pa^{J'}$ is true iff $\varphi'_P[\bar{a}]$ is true in
$\Tra{I}{J}$. We have the following proposition.

\begin{proposition}\label{PropReductionBasic} 
Let $\dep$ be a reduction relation of $\D$ in $\baseI$, $\Pa$ a domain
atom and let $I, I', J, J'$ be $\tau$-structures extending $\baseI$
such that $(I,J) \islePa (I',J')$.\\
  (a) If $P$ is defined then $\Pa^{\TPp_\D(I,J)}= \Pa^{\TPp_\D(I',J')}.$  \\
  (b) If $\dep$ is transitive, then $\TPp_\D(I,J) \islePa  \TPp_\D(I',J')$.
\end{proposition}

The condition of item (b) that $\dep$ should be transitive is not very
restrictive since the transitive closure of a reduction is a reduction
as well.

\begin{proof}
(a) Since $P$ is a defined predicate of $\D$, $\Pa^{\TPp_\D(I,J)}$ is
the truth value of $\varphi'_P[\bar{a}]$ in $\Tra{I}{J}$ and likewise
$\Pa^{\TPp_\D(I',J')}$ is the truth value of $\varphi'_P[\bar{a}]$ in
$\Tra{I'}{J'}$. Since $\dep$ is a reduction relation, the truth value
of this formula is the same in $\Tra{I}{J}$ as in  $\Tra{I'}{J'}$.

\noindent
(b) Assume that $\dep$ is transitive.  Let $\Qb$ be  an arbitrary
domain atom such that $\Qb \dep \Pa$. If $Q$ is an open predicate of
$\D$ then $\Qb^{\TPp_\D(I,J)} = \Qb^J = \Qb^{J'} =
\Qb^{\TPp_\D(I',J')}$.  Let $Q$ be a defined predicate of $\D$. By
transitivity of $\dep$, the set of atoms on which $\Qb$ depends is a
subset of the set of atoms on which $\Pa$ depends.  This, and the fact
that $(I,J) \islePa (I',J')$ implies that $(I,J) =_{\dep \Qb}
(I',J')$. By application of (a) we obtain that $\Qb^{\TPp_\D(I,J)} =
\Qb^{\TPp_\D(I'J')}$. 

\end{proof}

Under the condition that $\dep$ is a transitive reduction, item (b) of
this proposititon states that $\TPp_\D$ preserves $\islePa$, for each
domain atom $\Pa$. This is a key property.  The reduction relation
$\dep$ defines a collection of lattice congruences $\islePa$ in
$\vallat{\baseI}{\tau}$, one for each domain atom $\Pa$.  The operator
$\TPp_\D$ is the basic operator in the well-founded model
construction. The fact that it preserves the congruences $\islePa$
``propagates'' to the stable operator $ST_{\D}$ and to the
construction of the well-founded model. This leads to the main theorem
of this section.

Let $\D$ be total in $\baseI$ and $\dep$ a transitive reduction
relation of $\D$ in $\baseI$.
\begin{theorem} \label{TheoReductions}
  For  $\toD$-structures $\baseIo, J_{\rm o}$ extending
  $\baseI$, $\baseIo \islePa J_{\rm o}$ implies $\baseIo^\D \islePa
  J_{\rm o}^\D$.
\end{theorem}
In other words, the value of a defined atom $\Pa$ depends only on the open atoms
on which $\Pa$ depends according to reduction relation $\dep$. 

Consider the lattice $\vallat{\baseI}{\tau}$ which consists of
$\tau$-structures extending $\baseI$. To prove the theorem, we will
show that $\homoPaa{\cdot}$ is a lattice homomorphism and $\islePa$
the corresponding lattice congruence (confer Section
\ref{Sec-lattice-homo}).  Since $\TPp_\D$ preserves $\islePa$, by
application of the basic Proposition \ref{PropCongruence}, it
will be easy to show that also the stable operator $ST_{\D}$ preserves
$\islePa$ and that the statement of the theorem holds.


Let $\dep$ be a transitive reduction of $\D$ in $\baseI$ and assume
that $\taufun \subseteq \tauI{\baseI} \subseteq \tau$.  Then
$\langle\vallat{\baseI}{\tau},\sq\rangle$ is a complete lattice. For
any domain atom $\Pa$, we define the collection $\vallePa{\baseI}$ as
the image of $\vallat{\baseI}{\tau}$ under the mapping $\homoPa$.


\begin{proposition}\label{PropHomoPA} 
The structure $\langle \vallePa{\baseI}, \sq\rangle$ is a complete
lattice. The mapping $\homoPa:\vallat{\baseI}{\tau} \ra
\vallePa{\baseI}$ is a lattice homomorphism and its induced lattice
congruence is $\islePa$. The operator $\TPp_\D$ preserves $\islePa$.
If $\baseI$ is idempotent (i.e., $\homoPaa{\baseI}=\baseI$), the
homomorphic image of $\TPp_\D$ on $\vallePa{\baseI}$ is the operator
$\homoPaa{\TPp_\D(\cdot,\cdot)}$.
\end{proposition}
\begin{proof}
The proposition is straightforward. We prove only the last item. If
$\baseI$ is idempotent, then it is easy to see that
$\vallePa{\baseI}\subseteq\vallat{\baseI}{\tau}$.  Since $\TPp_\D$
preserves $\islePa$, it has a homomorphic image on $\vallePa{\baseI}$,
say $T$.  For all $I, J \in \vallePa{\baseI}$,
\[ \begin{array}{lll} 
T(I,J) & = T(\homoPaa{I},\homoPaa{J}) & \text{( by  idempotence of } \homoPa)\\
        & = \homoPaa{\TPp_\D(I,J)}  & \text{ (by definition of $T$) }
\end{array}\]
\end{proof}

\begin{remark}\label{Rem-idempotent}

A condition in Proposition \ref{PropHomoPA} is that $\baseI$ is
idempotent for $\homoPa$. This condition can always be
satisfied. Given a reduction $\dep$ of $\D$ in $\baseI$, there exists
an equivalent reduction $\dep'$ of $\D$ in $\baseI$ such that for all
$\Pa$, $\baseI$ is idempotent for $\homoPa$. 

Define $\dep' := \dep
\cup \{ ~ (\Qb,Q'[\bar{c}]) ~|~ Q \in \tauI{\baseI} \land
Q'\in\tau\setminus\tauI{\baseI}\}$. In $\dep'$, each domain atom
depends on the same atoms as in $\dep$ but also on all domain atoms
interpreted by $\baseI$. It is easy to see that for all $\Pa \in
\Att$:\\ 
(a) $|\baseI|_{\dep'\Pa}=\baseI$, and \\
(b) for all extensions $I, J$ of $\baseI$, $I \cong_{\dep'\Pa} J$ iff $I\islePa
J$.\\ 
Consequently, if $\dep$ is a reduction of $\D$ in $\baseI$, then
$\dep'$ is an equivalent reduction of $\D$ in $\baseI$ in which
$\baseI$ is idempotent for $\homoPa$, for all $\Pa\in\Att$.
\end{remark}

Unless explicitly stated otherwise, we assume that $\tfun \subseteq
\tauI{\baseI} \subseteq \toD$,  that $\dep$
is transitive and that $\baseI$ is idempotent for $\homoPa$, for all $\Pa\in\Att$.

\ignore{
{\footnotesize 
\marc{PROOF WITHOUT A POINT OR CONTENT / DROP IT??}
\begin{proof}
(a)  The relation $\sq$ is reflexive, anti-symmetric and transitive and
  for each set $S\subseteq \vallePa{\baseI}$, its least upper bound
  $\bigsqcup S$ belongs to $\vallePa{\baseI}$, hence it is the least
  upper bound of $S$ in $\vallePa{\baseI}$. Similarly, $\bigsqcap
  S$ is the greatest lower bound of $S$ in $\vallePa{\baseI}$.
  Hence, $\langle \vallePa{\baseI},\sq\rangle$ is a complete
  lattice.
  
(b) It is straightforward to show that $\homoPa$ is a lattice
homomorphism. It is easy to see that for any $\tau$-structures $I, J$
extending $\baseI$, $\homoPaa{I}=\homoPaa{J}$ iff $I \islePa
J$.

(c) The statement (c) is an alternative formulation of Proposition
\ref{PropReductionBasic}(b).
\end{proof}
}}

\begin{proposition}\label{PropReductionExtension} For each domain atom
$\Pa$, the operator $ST_{\D}$ preserves $\islePa$ in
$\vallat{\baseI}{\tau}$. Its homomorphic image on $\vallePa{\baseI}$
is $\homoPaa{ST_\D(\cdot)}$.  Moreover, for all
$I\in\vallat{\baseI}{\tau}$, $\homoPaa{ST_\D(I)}$ is
$\lfp(\homoPaa{\TPp_\D(\cdot,I)})$ in the lattice $\vallePa{I|_\toD}$.
\end{proposition}

\begin{proof}
Let $I$ be an element of $\vallat{\baseI}{\tau}$. The structure
$ST_\D(I)$ is the least fixpoint of the operator $\TPp_\D(\cdot,I)$ in
the sublattice $\vallat{I|_\toD}{\tau}\subseteq\vallat{\baseI}{\tau}$.
By Proposition \ref{PropHomoPA}, the operator preserves $\islePa$ and,
since we assume that $\homoPaa{\baseI}=\baseI$, its homomorphic image
in the lattice $\vallePa{I|_\toD}$ is
$\homoPaa{\TPp_\D(\cdot,\homoPaa{I})}$.  By Proposition
\ref{PropCongruence}(b), taking the homomorphic image and the least
fixpoint of this operator commute. Consequently, we obtain that
$$\begin{array}{lll}
\homoPaa{ST_\D(I)} & = \homoPaa{\lfp(\TPp_\D(\cdot,I))}\\ & = 
\lfp(\homoPaa{\TPp_\D(\cdot,\homoPaa{I})})\\ & =  \lfp(\homoPaa{\TPp_\D(\cdot,I)}) & (\text{ since } I \islePa \homoPaa{I}) \end{array}$$ in the image lattice
$\vallePa{I|_\toD}$.

\ignore{
Assume $I, J\in \vallat{\baseI}{\tau}$. The structure $ST_\D(I) =
\lfp(\TPp_\D(\cdot,I))$ in the lattice $\vallat{{I|_\toD}}{\tau}$ and,
by Lemma \ref{lemma-extended-lattice}, in
$\vallat{\baseI}{\tau}$. Applying Proposition \ref{PropCongruence}(b)
and Proposition \ref{PropHomoPA}, we obtain that$\homoPaa{ST_\D(I)} =
\lfp(\homoPaa{\TPp_\D(\cdot,\homoPaa{I})})$ in the image lattice
$\vallePa{\baseI}$. }

\noindent Assume $I, J\in \vallat{\baseI}{\tau}$ such that $I\islePa J$. Then $I|_{\toD} \islePa
J|_{\toD}$ and the lattices $\vallePa{I|_{\toD}}$ and
$\vallePa{J|_{\toD}}$ are identical and the operators
$\homoPaa{\TPp_\D(\cdot,I)}$ and $\homoPaa{\TPp_\D(\cdot,J)}$ on this
lattice are identical. We obtain that
$\homoPaa{ST_\D(I)}=\homoPaa{ST_\D(J) }$ or that $ST_\D(I)\islePa
ST_\D(J)$.

\end{proof}

Now we found that $ST_{\D}$ preserves $\islePa$ for each domain atom
$\Pa$, and we can repeat the argument for the construction of the
well-founded model.

\begin{proposition}\label{PropReductionWF}\label{LemReductionWF} 
 For each domain atom
$\Pa$, for all $\baseIo, \baseJo \in 
\vallat{\baseI}{\toD}$, if $\baseIo\islePa \baseJo$ then
$\lb{\baseIo}{\D} \islePa\lb{\baseJo}{\D}$ and $\ub{\baseIo}{\D}
\islePa \ub{\baseJo}{\D}$.  Moreover, 
$\homoPaa{\lb{\baseIo}{\D}}$ is\\ $\lfp((\homoPaa{ST_{\D}(\cdot)})^2)$, and 
$\homoPaa{\ub{\baseIo}{\D}}$ is $\gfp((\homoPaa{ST_{\D}(\cdot)})^2)$
in the lattice $\vallePa{\baseIo}$.
\end{proposition}
The proof of this proposition is entirely similar to the proof of
Proposition \ref{PropReductionExtension} and is omitted.  The
proposition entails that if $\D$ is total in $\baseIo$ and in
$\baseJo$ and $\baseIo\islePa \baseJo$, then ${\baseIo}^{\D} \islePa
{\baseJo}^{\D}$. This proves Theorem~\ref{TheoReductions}.

\section{Modularity}\label{Sec:modularity}
 

In this section, we split a definition $\D$ into subdefinitions $\{
\D_1, \D_2, \dots, \D_n\}$. We study under what conditions we can 
guarantee that for structure $I$,
$$ 
I\models \D \ \ \ \text{iff} \ \ \ I\models \D_1 \land \D_2 \land
\dots \land \D_n.
$$ 
This is the subject of the Modularity theorem. 
\ignore{IGNORE FOR THE MOMENT:
We also investigate under what conditions, a sub-definition of $\D$  
can be replaced by another definition whithout changing the original  
meaning. }
 
 \ignore{
The Modularity theorem is our main result here.  Problem-free
combining and decomposing of definitions is crucial while axiomatizing
a complex system.  For example, one may write two cycle-free modules
of the system, which, when combined, produce a cyclic dependency
between its syntactic components. Such a dependency may cause a change
in the intended meaning of the original definitions. However not every
syntactic cycle is problematic. If the condition of the Modularity
theorem are satisfied, one can guarantee that the composition does not
change the intended meaning of the original component definitions.
The ability to decompose definitions without side effects is useful
for analyzing complex definitions --- some properties of large
defintions are implied by properties of subdefinitions.  Thus, the
Modularity theorem is also an important tool for simplifying logical
formulas with definitions. 
} 

The Modularity theorem is our main result here. The theorem tells us
when we can understand a large definition as a conjunction of
component definitions. Frequently, these component definitions have a
simpler form --- they may be positive definitions or non-recursive
definition. Therefore, the ability to decompose definitions without
side effects is useful for analyzing large definitions --- some
properties of large defintions are implied by properties of
subdefinitions. Thus, the Modularity theorem is an important tool for
simplifying logical formulas with definitions. 

From a knowledge representation perspective, problem-free combining
and decomposing of definitions is crucial while axiomatizing a complex
system.  For example, one may write two cycle-free modules of the
system, which, when combined, produce a cyclic dependency between its
syntactic components. Such a dependency may cause a change in the
intended meaning of the original definitions. However not every
syntactic cycle is problematic. If the condition of the Modularity
theorem are satisfied, one can guarantee that the composition does not
change the intended meaning of the original component definitions.

\subsection{Partition of  Definitions}\label{subsection-decomposition}

Everywhere in this section, we fix a definition $\D$ over some
vocabulary $\tau$.

\begin{definition}[partition of definitions]\label{def-partition} 
A {\em partition of definition $\D$} is a set \\ $\{\D_1,\dots,\D_n\}$,
$1<n$, such that $\D = \D_1 \cup \dots \cup \D_n$, and if defined
symbol $P$ appears in the head of a rule of $\D_i$, $1\leq i\leq n$,
then all rules of $\D$ with $P$ in the head belong to $\D_i$. 
\end{definition}
If $\{\D_1,\dots, \D_n\}$ is a partition of $\D$, then $\cup_i
\tau_{\D_i}^{\rm d} = \tau_{\D}^{\rm d}$, and $\tau_{\D_i}^{\rm d}
\cap \tau_{\D_j}^{\rm d} = \emptyset$ whenever $i\not= j$.
Notice that each $\D_i$ has some ``new'' open symbols.  For instance, if
$P$ is defined in $\D$, but not in $\D_i$, then it is a new open
symbol of $\D_i$. Of course, it holds that $\tau=\toD \cup \td = \toi
\cup \tdi$, $1\leq i \leq n$.

The following theorem demonstrates that a model of a definition, is,
at the same time, a model of each of its sub-definitions. As a side
effect, we demonstrate that the totality of the large definition
implies the totality of its sub-definitions.
 
\begin{theorem}[decomposition]\label{theorem-decomposition}
Let $\D$ be a definition over $\tau$ with partition\\ $(\D_1,\dots,\D_n)$. 
Let $I$ be a $\tau$-structure.
If $I\models \D$ then $I\models \D_1\land\ldots\land\D_n$.
\end{theorem}
Before proving this theorem, let us consider some of its implications.

\noindent Let $I$ be a model of $\D$. 
The theorem says that for each $i$, $1\leq
i\leq n$, $\D_i$ is total in the restriction $I|_{\topen{\D_i}}$ of
$I$ to the open symbols of $\D_i$ and moreover that $I$ is the
$\D_i$-extension of $I|_{\topen{\D_i}}$.  Using the notations of
Definition \ref{D-extension}, this means that $I = \lb{I}{\D_i} =
\ub{I}{\D_i}$. We obtain the following corollary.
 
\begin{corollary}\label{corollary-partition} 
  If $I \models \D$, then for each $i$, $1\leq i\leq n$,
  $\D_i$ is total in $I|_{\topen{\D_i}}$.
\end{corollary}

\noindent The following example shows that the inverse direction of 
Theorem \ref{theorem-decomposition} does not hold in general.
\begin{example} \label{ExNonModular}
  Let $\D$, $\D_1$, $\D_2$ be the following definitions.
$$
\D:=\left\{
\begin{array}{l}
P\la Q,\\
Q\la P
\end{array}
\right\},
$$
$$
\D_1:=\left\{
\begin{array}{l}
P\la Q
\end{array}
\right\},
$$
$$
\D_2:=\left\{
\begin{array}{l}
Q\la  P
\end{array}
\right\}.
$$
Definition $\D$ is total, and its unique model is $\emptyset$ in which
both $P$ and $Q$ are false.  According to Theorem
\ref{theorem-decomposition},
$\emptyset$ satisfies $\D_1$ and $\D_2$. 
Note that $\{P,Q\}$ is not a model of $\D$ and yet, it satisfies
$\D_1$ and $\D_2$. Indeed, $\{P,Q\}$ is the $\D_1$-extension of the
$\topen{\D_1}$-structure $\{Q\}$ and the $\D_2$-extension of the
$\topen{\D_2}$-structure $\{P\}$.
\end{example}


To prove the theorem, we shall use two lemmas.  The stable operator
maps a structure $I$ to the least fixpoint of $\TPp_\D(\cdot,I)$ in
the lattice $\vallat{I|_{\toD}}{\tau}$. The first lemma shows that the
image of $I$ is also the least fixpoint of this operator in much
larger lattices.

\begin{lemma}\label{lemma-extended-lattice} Let $\D$ be a definition 
over $\tau$ and $I$ a $\tau$-structure. Let $\tauo$ be any vocabulary
such that $\taufun \subseteq \tauo \subseteq \toD$. The least
fixpoint of $\TPp_\D(\cdot,I)$ in the lattice
$\vallat{I|_{\tauo}}{\tau}$ is $ST_\D(I)$.
\end{lemma} 

\begin{proof} From Proposition \ref{prop:mon:antimon}, it follows  easily 
that the operator $\TPp_{\D}(\cdot,I)$ is a well-defined,
monotone operator in the lattice $\vallat{I|_{\tauo}}{\tau}$. Each
structure in the image of $\TPp_{\D}(\cdot,I)$ belongs to
$\vallat{I|_{\toD}}{\tau} \subseteq \vallat{I|_{\tauo}}{\tau}$.
Thus, the least fixpoint of $\TPp_{\D}(\cdot,I)$ in
$\vallat{I|_{\tauo}}{\tau}$ belongs to $\vallat{I|_{\toD}}{\tau}$ and
must be $ST_\D(I)$, the least fixpoint  of $\TPp_{\D}(\cdot,I)$ in
$\vallat{I|_{\toD}}{\tau}$. \end{proof}

\begin{lemma}\label{lemma-ind-step}
Let $M$ be a fixpoint of the stable operator $ST_{\D}$ extending
$\toD$-structure $\baseIo$. Let $K, L$ be $\tau$-structures
extending $\baseIo$ and let $L|_\toi=M|_\toi$. \\ 
(a) If $K \sq L$ and $K \sq M$ then $ST_{\D_i}(L)\sq ST_{\D}(K) $.\\ 
(b) If $L \sq K$ and $M  \sq K$ then $ST_{\D}(K) \sq ST_{\D_i}(L)$.
\end{lemma}
 
\begin{proof} 

Denote $\Moi:=M|_\toi$.  Note
that since $\D_i$ is exactly the set of rules of $\D$ defining the
predicates of $\tdi$, it holds for each $I, J$ extending $\baseIo$
that 
\begin{equation}\label{eqtdi}
\TPp_{\D}(I,J)|_{\tdi} = \TPp_{\D_i}(I,J)|_{\tdi}.
\end{equation}  
Also, by definition of $\TPp_{\D_i}$, it holds that
\begin{equation}\label{eqtoi}
\TPp_{\D_i}(I,L)|_{\toi} = L|_{\toi}= \Moi.
\end{equation} 
Let $K':=ST_{\D}(K)$ and $L':=ST_{\D_i}(L)$.  The structure $L'$ is the least
fixpoint of the monotone operator $\TPp_{\D_i}(\cdot,L)$ in the
lattice $\vallat{\Moi}{\tau}$ and, using Lemma
\ref{lemma-extended-lattice}, also in the larger
lattice $\vallat{\baseIo}{\tau}$. The structure $K'$ is the least
fixpoint of the monotone operator $\TPp_{\D}(\cdot,K)$ in the same
lattice $\vallat{\baseIo}{\tau}$.

(a) Let $K\sq L$ and $K\sq M$.  We show that $L'\sq K'$.  Our goal is
to show that for each $I\in \vallat{\baseIo}{\tau}$ such that $K'\sq
I$,
\begin{equation}\label{fff}
\TPp_{\D_i}(I,L) \sq \TPp_{\D}(I,K).
\end{equation}
Then by applying Lemma \ref{lemma-fixpoints-on-lattice}(b) in the lattice  $\vallat{\baseIo}{\tau}$, we will obtain that
$$
\lfp(\TPp_{\D_i}(\cdot,L)) \sq \lfp(\TPp_{\D}(\cdot,K)),
$$
or equivalently that $L'\sq K'$. We prove (\ref{fff}) separately for
open and defined symbols of $\D_i$.

\noindent Because $K \sq M$ and by anti-monotonicity of $ST_{\D}$, 
it holds that $M = ST_{\D}(M) \sq ST_{\D}(K) = K'$. Combined with
(\ref{eqtoi}), this yields $\TPp_{\D_i}(I,L)|_{\toi} = M|_{\toi} \sq
K'|_{\toi}$. By monotonicity of $\TPp_{\D}$ in its first argument, we
have that if $K'\sq I$ then $K' = \TPp_{\D}(K',K) \sq \TPp_{\D}(I,K)$.
We conclude that
\begin{equation}\label{ffff}
\TPp_{\D_i}(I,L)|_{\toi} \sq K'|_{\toi}\sq \TPp_{\D}(I,K)|_{\toi}.
\end{equation}
\noindent We need to show the same for the defined symbols.
Since $K\sq L$, anti-monotonicity of $\TPp_{\D}$ in the second
argument implies that $\TPp_{\D}(I,L) \sq \TPp_{\D}(I,K)$.  Using
(\ref{eqtdi}), we obtain that
\begin{equation}\label{fffff}
\TPp_{\D_i}(I,L)|_{\tdi}=\TPp_{\D}(I,L)|_{\tdi} \sq \TPp_{\D}(I,K)|_{\tdi} .
\end{equation}
Statements (\ref{ffff}) and (\ref{fffff}) give us (\ref{fff}), and we
conclude that $L' \sq K'$.\\

(b) Assume that $M\sq K$ and $L\sq K$.  We prove that for each $I\sq
K'$, $\TPp_{\D}(I,K) \sq \TPp_{\D_i}(I,L)$. Then we can apply Lemma
\ref{lemma-fixpoints-on-lattice}(a) which will yield the desired
result that $K' \sq L'$. The proof is  similar to that of (a).

\noindent First, by anti-mononticity of $ST_\D$, we have that $K'\sq M$.  
By monotonicity of $\TPp_{\D}$ in its first argument and
anti-monotonicity in its second argument, it holds for each $I\sq K'\sq
M$ that $\TPp_{\D}(I,K) \sq \TPp_{\D}(I,M) \sq \TPp_{\D}(M,M) = M$.
Using (\ref{eqtoi}), we obtain that 
\begin{equation}\label{eqg}
\TPp_{\D}(I,K)|_{\toi} \sq M|_{\toi} = \TPp_{\D_i}(I,L)|_{\toi} .
\end{equation}

\noindent Second,  by anti-monotonicity in the second argument it holds that  
$\TPp_{\D}(I,K) \sq \TPp_{\D}(I,L)$. Using  (\ref{eqtdi}), we find  
\begin{equation}\label{eqgg}
\TPp_{\D}(I,K)|_{\tdi} \sq \TPp_{\D}(I,L)|_{\tdi} =
\TPp_{\D_i}(I,L)|_{\tdi} .\end{equation}
Statements (\ref{eqg}) and (\ref{eqgg}) yield $K' \sq L'$.

\end{proof} 


\begin{proof}(of Theorem \ref{theorem-decomposition})
Let $M$ be the $\D$-extension of the $\toD$-structure $\baseIo$ and
denote $\Moi := M|_{\toi}$.  Consider the sequences of
$\tau$-extensions of $\baseIo$, the increasing sequence
$(I^{\xi})_{\xi \geq 0}$, and the decreasing sequence $(J^{\xi})_{\xi
\geq 0}$. The sequences are determined by operator $ST_{\D}$ in
the lattice $\vallat{\baseIo}{\tau}$.  Since $\D$ is total in
$\baseIo$, structure $M$ is the limit of both sequences.  Likewise,
for each $i$, consider two sequences, the increasing sequence
$(I_i^{\xi})_{\xi \geq 0}$ and the decreasing sequence
$(J_i^{\xi})_{\xi \geq 0}$, determined by the operator $ST_{\D_i}$ in
the sublattice $\vallat{\Moi}{\tau}$.  They converge to $\lb{\Moi}{\D_i}$
and $\ub{\Moi}{\D_i}$, respectively.  Recall that
  $$
 \begin{array}{c}
\ I^{\xi}:=ST_{\D}(J^{<\xi} ) \mbox{, where } 
                J^{<\xi}:= \sqcap_{\eta<\xi}J^\eta \\
J^{\xi}:=ST_{\D}(I^{<\xi}) \mbox{, where }  I^{<\xi}:= \sqcup_{\eta<\xi}I^\eta\\
\mbox{ and }\\
\ I_i^{\xi}:=ST_{\D_i}(J_i^{<\xi} )\mbox{, where } 
                J_i^{<\xi}:= \sqcap_{\eta<\xi}J_i^\eta  \\   
J_i^{\xi}:=ST_{\D_i}(I_i^{<\xi})\mbox{, where }  
                I_i^{<\xi}:= \sqcup_{\eta<\xi}I_i^\eta.
\end{array}
$$
We will prove that for each $\xi$,
\begin{equation}\label{eq-transf-ind}
I^{\xi} \sq  I_i^{\xi}  \sq J_i^{\xi} \sq J^{\xi}.
\end{equation}
This property allows us to conclude that, since the outer sequences
$(I^{\xi})_{\xi \geq 0}$ and $(J^{\xi})_{\xi \geq 0}$ converge to $M$,
the inner sequences $(I_i^{\xi})_{\xi \geq 0}$ and $(J_i^{\xi})_{\xi
\geq 0}$ converge to $M$ as well.  Therefore, we obtain that $\D_i$ is
total in $I|_{\toi}$ and that structure $M$ is the unique
$\D_i$-extension of $I|_{\toi}$. Since $i$ was arbitrary, we obtain
that $I\models \D_1\land \dots \land \D_n$.

\noindent 
For each $\xi$, we will prove statement (\ref{eq-transf-ind}) and the
following statement:
\begin{equation}\label{eq2}
I^{<\xi} \sq  I_i^{<\xi}  \sq J_i^{<\xi} \sq J^{<\xi}
\end{equation}
The two statements are proven by simultaneous induction on $\xi$.

\noindent First we prove the base case of statement (\ref{eq2}):
\begin{equation}\label{eq-decomp-base-case}
I^{<0} \sq  I_i^{<0}  \sq J_i^{<0}  \sq J^{<0}.
\end{equation}
This is  equivalent to 
$$
\bot_\baseIo \sq \bot_{\Moi} \sq \top_{\Moi} \sq 
\top_\baseIo
$$
which is straightforward. 

\noindent Second, assume that for arbitrary $\xi\geq 0$, 
statement (\ref{eq2}) holds.  We prove that then (\ref
{eq-transf-ind}) holds for $\xi$ as well.  Notice that, as a special
case, we obtain the base case for (\ref {eq-transf-ind}).

\noindent  By the construction of the sequences, we have for each $\xi$ that
$I_i^{<\xi} \sq J_i^{<\xi}$. By anti-monotonicity of $ST_{\D_i}$, we
obtain the middle inequality of (\ref {eq-transf-ind}). Similarly, we
have that $I^{<\xi}\sq M \sq J^{<\xi}$.  This, together with the
induction hypothesis, entails that the conditions of Lemma
\ref{lemma-ind-step} (a)+(b) are satisfied. Application of the lemma
yields that $I^{\xi}
\sq I_i^{\xi}$ and $J_i^{\xi} \sq J^{\xi}.$

\noindent Third, assume that for each $\eta <\xi$,
$$ 
I^{\eta} \sq I_i^{\eta} \sq J_i^{\eta} \sq J^{\eta} .
$$ 
 The standard fixpoint construction of anti-monotone operators
guarantees the inner inequality of (\ref{eq2}): 
$$
I_i^{<\xi} \sq J_i^{<\xi}
$$ 
By taking the union and intersection of  appropriate sets over all
$\eta <\xi$, we obtain the  two outer inequalities of (\ref{eq2}):
$$
I^{<\xi} \sq I_i^{<\xi} \mbox{ and } J_i^{<\xi} \sq J^{<\xi}.
$$

\noindent This proves the statements (\ref{eq-transf-ind}) and
(\ref{eq2}) for every $\xi$ and concludes the proof of the theorem.
\end{proof}


\subsection{Reduction Partitions}

Theorem \ref{theorem-decomposition} gives one direction of the
Modularity theorem.  Now our goal is to come up with some condition on
the partition of $\D$ so that both directions of the Modularity
theorem hold.  Recall from Example \ref{ExNonModular} that the other
direction does not hold in general. 

\begin{example} Recall the definitions in  Example \ref{ExNonModular}:  

$$
\D:=\left\{
\begin{array}{l}
P\la Q,\\
Q\la P
\end{array}
\right\} \ \ \ 
\D_1:=\left\{
\begin{array}{l}
P\la Q
\end{array}
\right\} \ \ \ 
\D_2:=\left\{
\begin{array}{l}
Q\la  P
\end{array}
\right\}.
$$
The structure $\{P,Q\}$ satisfies $\D_1 \land \D_2$ but not
$\D$. 
\end{example}

In the example, splitting the definition breaks the circular {\em
  dependency} between $P$ and $Q$. This causes the broken equivalence
between $\D$ and $\D_1 \land \D_2$. The example suggests that
splitting a definition will be equivalence preserving if the splitting
does not break circular dependencies between atoms. Below we will
formalise this notion using the notion of {\em reduction relation}
defined in Section \ref{Sec:reductions}. 

The following proposition formulates a simple and useful property of
(possibly non-transitive) reductions in the context of a partition.

\begin{proposition} \label{PropReductionSubdefinition}
  Let $\{\D_1,\dots,\D_n\}$ be a partition of $\D$. \\ 

(a) A relation $\dep$ is a reduction relation of $\D$ in $\baseI$ iff
  for each $i$, $1\leq i \leq n$, $\dep$ is a reduction relation of
  $\D_i$ in $\baseI$.\\ 
(b) Let for each $i$, $1\leq i \leq n$, $\dep_i$ be reduction relation
  of $\D_i$ in $\baseI$. Then $\dep_1\cup \dots\cup\dep_n$ is a
  reduction relation of $\D$ in $\baseI$.
\end{proposition}
\begin{proof} (a) 
For each defined predicate $P$, there is exactly one $i$ such that $P$
is defined in $\D_i$ and the formulas $\varphi_P^{\D}$ defining $P$ in
$\D$ and $\varphi_P^{\D_i}$ defining $P$ in $\D_i$ are identical. It
is obvious then that $\dep$ is a reduction of $\D$ in $\baseI$ iff
$\dep$ is a reduction of $\D_i$ in $\baseI$, for each $i$, $1\leq i
\leq n$.

(b) If for each $i \in \{1, \dots , n\}$, $\dep_i$ is a reduction of
$\D_i$ in $\baseI$ then by Proposition \ref{PropSupReduction},
$\dep_1\cup \dots\cup\dep_n$ is a reduction relation of each $\D_i$.
By (a), $\dep_1\cup \dots\cup\dep_n$ is a reduction relation of
$\D$ in $\baseI$.
\end{proof}

Recall that a pre-well-order is a reflexive and transitive relation
such that every non-empty subset contains a minimal element.

\ignore{
\begin{definition}[reduction pre-well-order]
  A reduction relation $\dep$ of $\D$ in $\baseI$ is a {\em reduction
    pre-well-order} if it is reflexive and transitive order such that
  every non-empty subset contains a minimal element.
\end{definition} 
}
\ignore{
Note that every definition $\D$ has a trivial reduction pre-well-order in
every $\toD$-structure. Indeed, the total binary relation on $\Att$
trivially satisfies the conditions of a reduction pre-well-order.
}
 
The following definition is crucial for the right-to-left direction of
the Modularity theorem. Let $\baseI$ be a structure such that
$\tauI{\baseI} \subseteq \toD$.
\begin{definition}[reduction partition]\label{Def-reduction-partition}
  Call partition $\{\D_1,\dots, \D_n\}$ of definition $\D$ a {\em
    reduction partition} of $\D$ in $\baseI$ if there is a reduction
    pre-well-order $\dep$ of $\D$ in $\baseI$ and if $\Qb \dep \Pa$
    and $\Pa \dep \Qb$, then $P$ and $Q$ are both open predicates of
    $\D$ or they are defined in the same $\D_i$.
\end{definition} 
Equivalently, if $P$ and $Q$ are not defined in the same $\D_i$, then
$\Qb \sdep \Pa$ iff $\Qb \dep \Pa$.  The intuition underlying this
definition is that in a reduction partition, if an atom defined in one
module depends on an atom defined in another module, then the latter
atom is strictly less in the reduction ordering and hence does not
depend on the first atom.

In a first step towards proving the second half of the modularity
theorem, we prove that if $\{\D_1, \dots,\D_n\}$ is a reduction
partition of $\D$ in a $\toD$-structure $\baseIo$, then the conjunction
$\D_1\land\dots\land \D_n$ has at most one model extending
$\baseIo$.
\begin{theorem}
\label{theorem-uniqueness} If $\D$ has a 
  reduction partition $\{\D_1,\dots , \D_n\}$ in a
  $\topen{\D}$-structure $\baseIo$ then $\D_1\land\dots\land \D_n$
  has at most one model extending $\baseIo$.
\end{theorem}

\begin{proof}
  
  Let $M$ and $M'$ be models of $\D_1\land\dots\land \D_n$.  Assume
  towards contradiction that $M$ and $M'$ differ. Let us select a
  minimal atom $\Pa$ in the reduction pre-well-order $\dep$ such that
  $M$ and $M'$ disagree on $\Pa$. Assume that $\Pa$ is defined in
  $\D_i$.  Since $\dep$ is reflexive and $M$ and $M'$ disagree on
  $\Pa$, $M \not \islePa M'$. On the other hand, because $\Pa$ is
  minimal, it holds that $M \islPa M'$.  It follows that $M|_{\toi}
  \islPa M'|_{\toi}$. Moreover, if $Q\in \toi$ then for each atom
  $Q[b]$, $Q[b]\dep \Pa$ iff $Q[b]\sdep \Pa$. Hence, we have
  $M|_{\toi} \islePa M'|_{\toi}$. By Proposition
  \ref{PropReductionSubdefinition}(a), $\dep$ is a transitive
  reduction relation  of $\D_i$ in $\baseIo$. The condition of
  Proposition \ref{PropReductionWF} holds and we can infer that $M =
  (M|_{\toi})^{\D_i} \islePa (M'|_{\toi})^{\D_i} = M'$. We obtain
  $M\islePa M'$, a contradiction.
\end{proof}



\begin{example} \label{ExNM1} 
Consider the definitions from Example \ref{ExNonModular}:
$$\D:= \defin{ P\la Q,\\ Q\la P } ,\ \  \D_1:= \defin{ P\la Q } , \ \ 
\D_2:= \defin{ Q\la  P }.
$$
Each reduction of the definition $\D$ includes tuples $(P,Q)$ and
$(Q,P)$. Hence, the partition $(\D_1,\D_2)$ is not a reduction
partition. The formula $\D_1\land \D_2$ has multiple models
$\emptyset$ and $\{P,Q\}$.
\end{example} 

\begin{example} \label{ExNMevenodd}
Consider the partition of the definition from Example \ref{D3}:
$$
\D_1:= \defin{\forall x\ (E(x)\la x=0)\ ,\\ 
        \forall x\ (E(s(x)) \la O(x)) },\ \ 
\D_2:=  \defin{ \forall x \ ( O(s(x))\la E(x)) }.
$$ 
The transitive reflexive closure $\dep^{**}$ of the reduction of $\D$
presented in Example \ref{D3} is a well-founded partial order. It
holds that $E[n] \dep^{**} O[m]$ and $O[n] \dep^{**} E[m]$ iff $n<m$.
Consequently, $\{\D_1,\D_2\}$ is a reduction partition. The
conjunction of $\D_1$ and $\D_2$ has one model.
\end{example} 
\begin{example}\label{ExBadDef}
Consider the following definitions:
$$
\D := \defin{ P\la \neg P,\\  Q\la \neg P } , \ \ \D_1 := \defin{ P\la \neg P } ,\ \ \D_2 := \defin{ Q\la \neg P } .
$$
The reflexive closure of the relation $\{(P,Q)\}$ is a well-order.
Clearly, the partition $(\D_1,\D_2)$ is a reduction partition. The
definitions $\D$ and $\D_1$ and the conjunction $\D_1 \land \D_2$ are
all inconsistent.
\end{example}

\ignore{ 
\begin{example}
$$
\D_1 := \defin{P\la \neg P,\\
 Q\la \neg P}.
$$
It has the following Definition $\D_1$ from Example \ref{D1}
has the following reduction partition:
$\{\{P\la \neg P\},\{Q\la \neg P\}\}$. The reflexive, transitive closure 
of $\{(P,Q)\}$ is the the corresponding reduction pre-well-order. 
\end{example}

\begin{example}
$\{\{P\la Q\},\{Q\la P\}\}$ is not a reduction partition  of $\D_2$
from Example \ref{D2} because $P$ depends on $Q$ and $Q$ depends 
on $P$. This definition does not have such a partition. 
Note that $\{P,Q\}$ is not a model of the definition, but it is 
a model of each of the component definitions. 
\end{example}

\begin{example}\label{D3a}
In the structure $( \mathbb{N}, s )$, 
the following two definitions

\[      \D_4 := \left \{\begin{array}{l} 
\forall x \ (E(x)\la x=0),\\ 
      \forall x \ (  E(s(x)) \la O(x))\end{array}\right\}, 
        \] 
\[\D_5 := \left \{\begin{array}{l} 
 \forall x \ (  O(s(x))\la E(x))
\end{array}\right\} 
        \]
 form a  reduction partition  $\{\D_4,\D_5\}$  of
$\D_3$ from Example \ref{D3}. The corresponding reduction
 ordering  is the one defined in Example \ref{D3}. 
\end{example}

}

\ignore{
\subsection{A Sufficient Condition for Totality 
of Complex Definitions} }

In the next step towards proving the second half of the modularity
theorem, we prove the totality of a well-behaved definition $\D$ with
a  reduction partition. 
\begin{definition} 
A partition $\{\D_1,\dots,\D_n\}$ of definition $\D$ is total in a
structure $\baseI$ ($\tauI{\baseI}\subseteq \toD$) if each
$\D_i$, $1\leq i \leq n$, is total in $\baseI$.
\end{definition}
 
 
 \begin{theorem}[totality]\label{theorem-totality} If $\D$ has a total
  reduction partition $\{\D_1,\dots , \D_n\}$ in a structure
  $\baseI$ ($\tauI{\baseI}\subseteq\toD$) then $\D$ is total in ${\baseI}$.
  \end{theorem}
  
\noindent Thus, one way to prove that $\D$ is total in $\baseI$ is
to prove that it has a reduction partition, and that each definition
$\D_i$ in the partition is total in ${\baseI}$. \\

To prove this theorem, we  need the following  lemma.  
\begin{lemma}\label{lemma-totality-theorem-1} 
Let $\{\D_1,\dots,\D_n\}$ be a partition of $\D$ and let $\dep$ a
reduction of $\D$ in $\toD$-structure $\baseIo$ such that $\homoPaa{\baseIo}=\baseIo$ for all $\Pa\in\Att$. Let $M, M'$ be
$\tau$-structures extending $\baseIo$ such that for 
some domain atom $\Pa$, $(M,M')$ is an
oscillating pair of $\homoPaa{ST_\D(\cdot)}$ and
$M|_\toi=M'|_\toi=\Moi$. Then for all $I\in \vallePa{\Moi}$, it holds
that: \\ 
(a) if $I\sq M$, then $M'\sq \homoPaa{ST_{\D_i}(I)}$;\\ 
(b) if $M'\sq I$, then $\homoPaa{ST_{\D_i}(I)}\sq M$.
\end{lemma}
The lemma has a  similar proof as Lemma \ref{lemma-ind-step}.

\begin{proof}  
By Proposition \ref{PropReductionExtension}, the structure $M'$ is the
least fixpoint of $\homoPaa{\TPp_{\D}(\cdot,M)}$ and $M$ the least
fixpoint of $\homoPaa{\TPp_{\D}(\cdot,M')}$ in the lattice
$\vallePa{\baseIo}$. Since $M, M' \in \vallePa{\Moi}$, they are also
the least fixpoints of these operators in the sublattice
$\vallePa{\Moi}$. Let $I\in \vallePa{\Moi}$ and denote $I':=
\homoPaa{ST_{\D_i}(I)}$. 
By Proposition~\ref{PropReductionSubdefinition}, $\dep$ is a reduction
of $\D_i$. By Proposition~\ref{PropReductionExtension}, $I'$ is the
least fixpoint of $\homoPaa{\TPp_{\D_i}(\cdot,I)}$ in
$\vallePa{\Moi}$.

(a) Let $I \sq M$.  We prove that $M'\sq I'$. Our goal is to show that
for each $J\in \vallePa{\Moi}$ such that $J\sq M'$,
\begin{equation}\label{f1}
\homoPaa{\TPp_{\D}(J,M)}\sq\homoPaa{\TPp_{\D_i}(J,I)}.
\end{equation}
Then  by Lemma \ref{lemma-fixpoints-on-lattice}(a), it will follow 
that
$$
M' = \lfp (\homoPaa{\TPp_{\D}(\cdot,M)}) \sq 
\lfp(\homoPaa{\TPp_{\D_i}(\cdot,I)}) =I'.
$$
We prove (\ref{f1}) separately for open and defined symbols of $\D_i$.\\

\noindent First, let $J\sq M'$. By monotonicity in the first argument,  
$\homoPaa{\TPp_{\D}(J,M)} \sq \homoPaa{\TPp_{\D}(M',M)} = M'$.  Also,
$\TPp_{\D_i}(J,I)|_{\toi} = \Moi = \homoPaa{\Moi}$. Combining these
statements, we obtain:
\begin{equation}\label{f10}
\homoPaa{\TPp_{\D}(J,M)}|_{\toi} \sq M'|_{\toi} =  \Moi = \homoPaa{\TPp_{\D_i}(J,I)}|_{\toi}.
\end{equation}

\noindent Second, $\TPp_{\D}$ is anti-monotone in its second
argument, which implies $\TPp_{\D}(J,M) \sq
\TPp_{\D}(J,I)$, for each $J\in \vallePa{\Moi}$.  
Since the operators $\TPp_{\D}$ and
$\TPp_{\D_i}$ coincide on the defined symbols of $\D_i$, it
follows that:
$$
\TPp_{\D}(J,I)|_{\tdi}=\TPp_{\D_i}(J,I)|_{\tdi}.
$$
By combining these statements, we conclude that:
$$
\TPp_{\D}(J,M)|_{\tdi} \sq \TPp_{\D}(J,I)|_{\tdi} =  
\TPp_{\D_i}(J,I)|_{\tdi}.
$$
After projection with  $\homoPa$, we obtain:
\begin{equation}\label{f11}
\homoPaa{\TPp_{\D}(J,M)}|_{\tdi} \sq  
\homoPaa{\TPp_{\D_i}(J,I)}|_{\tdi}.
\end{equation}
The combination of (\ref{f10}) and (\ref{f11}) yields statement
(\ref{f1}).

(b) Let $M'\sq I$. We show that for each $J\in \vallePa{\Moi}$ such
that $M\sq J$, $$\homoPaa{\TPp_{\D_i}(J,I)} \sq
\homoPaa{\TPp_{\D}(J,M')}.$$ Then we can apply Lemma
\ref{lemma-fixpoints-on-lattice}(b) to prove  (b).

\noindent For the open predicates, if $M\sq J$ then by the same 
kind of reasoning as in (a),
$$
\homoPaa{\TPp_{\D_i}(J,I)}|_{\toi} = \Moi = 
\homoPaa{\TPp_{\D}(M,M')}|_{\toi}\sq \homoPaa{\TPp_{\D}(J,M')}|_{\toi}.
$$
For the defined predicates, for each $J$ it holds that
$$\TPp_{\D_i}(J,I)|_{\tdi}= \TPp_{\D}(J,I)|_{\tdi} 
\sq \TPp_{\D}(J,M')|_{\tdi}.$$ 
Combining both results, we obtain that 
$\homoPaa{\TPp_{\D_i}(J,I)}\sq\homoPaa{\TPp_{\D}(J,M')}$.
\end{proof} 

\begin{proof} of Theorem \ref{theorem-totality}.\\
  Let $\baseIo$ be an arbitrary $\toD$-extension of $\baseI$. Then,
  $\{\D_1,\dots , \D_n\}$ is a total partition of $\D$ in $\baseIo$
  and, by Proposition \ref{PropSupReduction}, $\{\D_1,\dots , \D_n\}$
  is a reduction partition of $\D$ in $\baseIo$.  Let $\dep$ be a
  pre-well-ordered reduction of $\D$ in $\baseIo$ satisfying the
  condition of Definition~\ref{Def-reduction-partition}.  We assume
  idempotence of $\baseIo$, i.e. $\homoPaa{\baseIo}=\baseIo$ for all
  $\Pa\in\Att$. This assumption can always be made: see Remark
  \ref{Rem-idempotent}.

  Assume, towards a contradiction, that $\D$ is not total in
  $\baseIo$. The assumption implies that $\lb{\baseIo}{\D} \not =
  \ub{\baseIo}{\D}$.  Let $\Pa$, where $P \in \tdD$, be a minimal atom
  in the reduction ordering $\dep$ such that $\lb{\baseIo}{\D} \not
  \models \Pa$ and $\ub{\baseIo}{\D}\models \Pa$.  By reflexivity of
  $\dep$ and our choice of $\Pa$, it holds that
$$ 
\lb{\baseIo}{\D} \not\islePa \ub{\baseIo}{\D}, 
$$
$$ 
\lb{\baseIo}{\D} \islPa \ub{\baseIo}{\D}. 
$$
Because we have a reduction partition, for each atom $\Qb$ not defined
in $\D_i$, $\Qb \dep \Pa$ iff $\Qb \sdep \Pa$. Therefore,
$$ 
\lb{\baseIo}{\D}|_\toi \islePa \ub{\baseIo}{\D}|_\toi. 
$$
Define $M := \homoPaa{\lb{\baseIo}{\D}}$ and $M' :=
\homoPaa{\ub{\baseIo}{\D}}$ and let $\Moi := M|_\toi = M'|_\toi = \homoPaa{(\lb{\baseIo}{\D}|_\toi)}$. Since $\baseIo$ is idempotent for $\homoPa$, $\Moi$ is an extension of $\baseIo$. The structures $M$ and $M'$ are  different  
in $\Pa$.

\noindent On the one hand, since  $\D_i$ is total in $\baseIo$ and $\Moi$ is an extension of $\baseIo$, it holds that
$$\lb{\Moi}{\D_i}=\ub{\Moi}{\D_i}.$$ On the other hand, we will prove
the following:
\begin{equation}\label{eqtot2} \homoPaa{\lb{\Moi}{\D_i}} \sq M \sq M' \sq
\homoPaa{\ub{\Moi}{\D_i}}.\end{equation} 
Since $M\neq M'$, we will obtain the contradiction.

\noindent The proof of (\ref{eqtot2}) is by induction. 
By Proposition \ref{LemReductionWF}, the following equations hold in
the lattice $\vallePa{\Moi}$:
$$
\homoPaa{\lb{\Moi}{\D_i}} = \lfp((\homoPaa{ST_{\D_i}(\cdot)})^2) \text{ and } \homoPaa{\ub{\Moi}{\D_i}} =  \gfp((\homoPaa{ST_{\D_i}(\cdot)})^2).
$$
Consider the sequences $(\I_i^{\xi})_{\xi \geq 0}$ and
$(\J_i^{\xi})_{\xi \geq 0}$ determined by the operator
$\homoPaa{ST_{\D_i}(\cdot)}$ in the lattice $\vallePa{\Moi}$.
We shall demonstrate that the following holds: for every $\xi$,
\begin{equation} \label{statement}
\I_i^{\xi} \sq M \sq M' \sq \J_i^{\xi}.
\end{equation} 
\begin{equation} \label{statement1}
\I_i^{<\xi} \sq  M \sq M' \sq  \J_i^{<\xi}. 
\end{equation}

\noindent Statements (\ref{statement}) and (\ref{statement1}) are proven by
simultaneous transfinite induction on $\xi$.

\noindent First, we establish the base case of statement (\ref{statement1}): 
$$ 
I_i^{<0} \sq M \sq M' \sq J_i^{<0}.
$$ 
This is straightforward since $M, M' \in \vallePa{\Moi}$ and $I_i^{<0}$ and
$J_i^{<0}$ are the bottom and top element, respectively, of this
lattice.

\noindent Second, we  show that, for arbitrary $\xi$, if
(\ref{statement1}) holds then (\ref{statement}) holds.  Let us assume
that the statement (\ref{statement1}) holds for $\xi$. Since
$I_i^{<\xi}\ \sq M$, Lemma \ref{lemma-totality-theorem-1}(a) implies
that $M' \sq \homoPaa{ST_{\D_i}(I_i^{<\xi})} = J^{\xi}$.  Since $M'
\sq \J_i^{<\xi}$, then by Lemma \ref{lemma-totality-theorem-1}(b),
$\I_i^{\xi} = \homoPaa{ST_{\D_i}(J_i^{<\xi})} \sq M$.

\noindent Third, it remains to be proven that if for all $\eta<\xi$, 
it holds that $I_i^{\eta} \sq M \sq M' \sq \J_i^{\eta}$, then
(\ref{statement1}) holds for $\xi$. This is straightforward.

\noindent This completes the proof of the theorem.
\end{proof}

\begin{example} \label{ExNMevenodd2} 
As seen in Example \ref{ExNMevenodd}, the following partition is a
reduction partition of $\D_1\cup\D_2$ in the natural numbers:
$$
\D_1:= \defin{\forall x \ (E(x)\la x=0) \ ,\\ 
        \forall x\ (E(s(x)) \la O(x)) },\ \ \ 
\D_2:=  \defin{ \forall x \ ( O(s(x))\la E(x)) }.
$$ 
Both subdefinitions are non-recursive and positive. Consequently, both
are total. So, the conditions of Theorem \ref{theorem-totality}
hold. This definition is total in the natural numbers and has a unique
model.
\end{example}

\begin{corollary}[Consistency]
\label{corollary-composition}If $\D$ has a total
  reduction partition $\{\D_1,\dots , \D_n\}$ in a
  $\topen{\D}$-structure $\baseIo$, then $\D$ and
  $\D_1\land\dots\land\D_n$ are consistent and have a model
  extending $\baseIo$.
\end{corollary}

\begin{example}\label{ExBadDef1}
Recall the definition of Example \ref{ExBadDef}:
$$
\D := \defin{ P\la \neg P,\\  Q\la \neg P } , \ \ \D_1 := \defin{ P\la \neg P }
 ,\ \ \D_2 := \defin{ Q\la \neg P } .
$$
Although this definition has a  reduction partition, it is not
consistent. Corollary \ref{corollary-composition} does not hold
because $\D_1$ is not total in any structure.
\end{example}

Now, we are in a position to prove the second direction of the
modularity theorem. Let $\tauo\subseteq \toD$. 
\begin{theorem}
\label{theorem-composition} If $\D$ has a total
  reduction partition $\{\D_1,\dots , \D_n\}$ in the
  $\tauo$-structure $\baseI$, then for any $\tau$-structure $M$
  extending $\baseI$, if $M \models \D_1\land\dots\land\D_n$ then $M
  \models \D$.
\end{theorem}

\begin{proof}
Assume $M$ extends $\baseI$ and $M \models \D_1\land\dots\land\D_n$
and let $\baseIo = M|_{\toD}$.  Since $\baseIo$ is an extension of
$\baseI$, $\{\D_1,\dots , \D_n\}$ is a total reduction partition of
$\D$ in $\baseIo$. The conditions of Theorem
\ref{theorem-totality} are satisfied. Consequently, $\D$ is total in
$\baseIo$. The structure $\baseIo^\D$ is a model of $\D$ and, by
Theorem
\ref{theorem-decomposition}, of $\D_1\land \dots\land\D_n$. Since by
Theorem \ref{theorem-uniqueness}, $M$ is the unique model of
$\D_1\land \dots\land\D_n$ extending $\baseIo$,
$M$ and $\baseIo^\D$ are identical.
\end{proof}

\begin{theorem}[modularity]\label{theorem-modularity}
If $\{\D_1,\dots , \D_n\}$ is a total reduction partition of $\D$ in
$\tauo$-structure $\baseI$, then for any $\tau$-structure $M$
extending $\baseI$,
$$M\models \D \ \ \text {iff} \ \  M\models \D_1\land \dots \land \D_n.
$$
\end{theorem}
\begin{proof}
  Combine theorems \ref{theorem-decomposition} and
  \ref{theorem-composition}.
\end{proof}

Another immediate consequence is the following corollary.
\begin{corollary}\label{corollary-modularity-theorem}
  Let $T_o$ be a theory over $\tauo$ such that for any $\tauo$-model
  $M_o$ of $T_o$, $\{\D_1,\dots,\D_n\}$ is a total reduction partition
  of $\D$ in $M_o$.
  
  Then $T_o\land \D$ and $T_o\land \D_1\land \dots \land \D_n$ are
  logically equivalent.
\end{corollary}

\begin{example} \label{ExNMevenodd3} 
As seen in Example \ref{ExNMevenodd2}, the following partition is a
total reduction partition of $\D_1\cup\D_2$ in the natural numbers:
$$
\D_1:= \defin{\forall x \ (E(x)\la x=0) \ ,\\ 
        \forall x\ (E(s(x)) \la O(x)) },\ \ \ 
\D_2:=  \defin{ \forall x \ ( O(s(x))\la E(x)) }.
$$
In ID-logic, the natural numbers are formalised by
$T_{\mathbb{N}}$ (Examples \ref{Peano} and \ref{Ex:nat2}).  By
Corollary \ref{corollary-modularity-theorem}, the theories
$T_{\mathbb{N}} \cup \{\D_1\cup\D_2\}$ and $T_{\mathbb{N}} \cup
\{\D_1\land \D_2\}$ are equivalent.
\end{example}

\section{Some Familiar Types of Definitions}\label{Sec:familiar-types}

This section reconsiders the four different types of informal
inductive definitions discussed in section
\ref{Sec:forms-of-induction}: non-recursive definitions, positive
definitions, definitions over well-founded sets and iterated inductive
definitions. We demonstrate that these types of definitions can be
correctly and uniformly represented in ID-logic. To this end, we
define four formal subclasses of definitions of ID-logic that naturally
correspond to the four informal types of inductive definitions and
prove theorems to  show that the well-founded semantics correctly
formalises the meaning of these types of definitions.\\

\subsection{Non-Recursive Definitions.}
A first case is that of non-recursive definitions. A definition $\D$
is non-recursive if the bodies of the rules do not contain defined
predicates.

\begin{definition}[completion of $\D$] Define the {\em completion of
$\D$}, denoted $comp(\D)$, as the conjunction, for each defined symbol
$X$ of $\D$, of formulas
$$ 
\forall \x (X(\x) \lra \varphi_X[\x]).
$$ 
\end{definition}
The equivalence $\forall \x (X(\x) \lra \varphi_X[\x])$ is sometimes
referred at as the completed definition of $X$. 

\begin{theorem}\label{theorem-Hier} Let $\D$ be a non-recursive
  definition over $\tau$.  Then $\D$ is total and a $\tau$-structure
$I$ satisfies $\D$ iff $I$ satisfies $comp(\D)$.
\end{theorem}
\begin{proof} It is straightforward to show that if $\D$ 
is non-recursive, then for each $\toD$-structure $\baseIo$, the
operator $\TPp_{\D}$ is constant in the lattice
$\vallat{\baseIo}{\tau}$ and it maps each pair of $\tau$-structures to
the unique structure $I$ such that, for each defined symbol $X$,
$$
X^I = \{\bd \  |  \ \baseIo \models\varphi_X[\bd]\}.
$$ 
This $I$ is the unique model of $\D$ and the unique model of
$comp(\D)$ in $\vallat{\baseIo}{\tau}$.
\end{proof}

\subsection{Positive Definitions.} 
Let $\D$ be a positive definition, defining the symbols $\PP$. Let
$\X$ be a set of new predicate symbols such that for each defined
symbol $P_i$, $X_i$ and $P_i$ have the same arity.  Define the
following formula
$$
\posind{\D} := \bigwedge \D \land
               \forall \X( \bigwedge \D[\PP/\X] 
                                        \supset (\PP \subseteq \X))
.
$$
Here, $\bigwedge \D$ is the conjunction of formulas obtained
by replacing definitional rules with material implications in $\D$;
$\D[\PP/\X]$ is the definition obtained by substiting $X_i$ for each
defined symbol $P_i$ and $\PP \subseteq \X$ is a shorthand for the
formula $(\forall \xx P_1(\xx) \supset X_1(\xx)) \land \dots \land (\forall
\xx P_n(\xx) \supset X_n(\xx))$. The formula $\posind{\D}$ is
the standard second-order formula to express that predicates
$\PP$ satisfy the positive inductive definition $\D$. 

\newcommand{\Circ}[1]{Circ(#1)}
Define also $$
\Circ{\D;\PP} := 
             \bigwedge \D \land \forall \X( \bigwedge \D[\PP/\X] \land \X \subseteq \PP )
                                        \supset \PP \subseteq \X)
.
$$
This formula is the standard circumscription of $\bigwedge \D$ with
respect to the defined predicates $\PP$
\cite{Lifschitz:circumscription}.

\begin{theorem}\label{theorem-Mon}
Let $\D$ be a positive definition over $\tau$.  Then $\D$ is total and
for all $\tau$-structures $I$, the following are equivalent:

(a) $I$ is a model of $\D$;

(b) $I$ is the least fixpoint of $\G_\D$ in the lattice
$\vallat{\valIopen}{\tau}$;

(c)  $I$ is a model of $\posind{\D}$;

(d)  $I$ is a model of $\Circ{\D;\PP}$.
\end{theorem}
\begin{proof}
In case $\D$ is a positive definition, defined symbols have no
negative occurrences, so $\D$ and $\D'$ are identical. Consequently,
for any pair of structures $I, J$ in the lattice
$\vallat{\baseIo}{\tau}$, it holds that $\TPp_{\D}(I,J)=\G_\D(I)$
which does not depend on $J$.  Thus, the stable operator
$ST_\D$ is a constant operator in this lattice and maps any structure
$J$ to the least fixpoint of $\G_\D$. Thus, it follows that
$\lb{\baseIo}{\D}$ and $\ub{\baseIo}{\D}$ are identical to the
least fixpoint of $\G_\D$ in $\vallat{\baseIo}{\tau}$.
This proves the equivalence of (a) and (b). 

\noindent The equivalence of (b) and (c) in case of a positive definition
is well-known (see e.g. \cite{Aczel77}). Finally, the axiom
$\posind{\D}$ expresses that $\PP$ should be the least relations
satisfying $\bigwedge\D$, while $\Circ{\D;\PP}$ expresses
that $\PP$ should be minimal relations satisfying $\bigwedge\D$. Both
axioms are equivalent, since there is a set of least relations
satisfying $\bigwedge\D$, and it is the unique set of minimal
relations satisfying this formula.
\end{proof}

The theorem is significant since it shows that for positive
definitions, the semantics defined here coincides with standard monotone
induction. It implies  that if $I\models \D$ then $I$ is the least
structure extending $\baseIo$ that  satisfies the rules of $\D$
viewed as a set of first-order implications.

\begin{example} \label{Peano3}
Consider the formulation of the induction axiom 
in ID-logic in Example \ref{Peano}:
$$
\exists N \left [ 
\defin{ \forall x \ (N(x) \rul  x=0), \\ \forall x\ ( N(s(x)) \rul N(x))}
\land \forall x\ N(x))\right ].
$$
By Theorem \ref{theorem-Mon}, it is equivalent to the second-order axiom
\ignore{
{\tt THIS EXAMLE IS THE ONLY PLACE WHERE RIGHT-TO-LEFT 
MATERIAL IMPLICATION IS USED. I COULD NOT BECOME COMFORTABLE WITH
IT. TO ME, DEFINITIONAL IMPLICATION GOES RIGHT-TO-LEFT, AND 
MATERIAL IMPLICATION GOES LEFT-TO-RIGHT. IT IS PERMANENTLY 
WIRED IN MY BRAIN. CAN WE REVERSE THESE IMPLICATIONS HERE? 

MARC: DONT BE SILLY!!! :-) Eugenia: SILLY YORSELF!!! :-) }}
 
$$
\exists N \left [
\begin{array}{l} 
\forall x \ (N(x) \subset  x=0) \land \\
\forall x\ ( N(s(x)) \subset N(x)) \land \\
\forall X \ [ \forall x \ (X(x) \subset  x=0) \land 
\forall x\ ( X(s(x)) \subset X(x)) \supset \forall x (N(x) \supset X(x)) ] \land\\
\forall x \ N(x)
\end{array}
\right ].
$$
We show that this formula is logically equivalent with the standard
induction axiom. The first two conjuncts follow from the last and may
be deleted.  Using the last conjunct, the third conjunct can be simplified as follows:
 $$
\exists N \left [
\begin{array}{l} 
\forall X \ [ \forall x \ (X(x) \subset  x=0) \land 
\forall x\ ( X(s(x)) \subset X(x)) \supset \forall x \ X(x) ] \land\\
\forall x \ N(x))
\end{array}
\right ].
$$

Notice that the first element of the conjunction does not depend of $N$,
so the outer existential quantifier can be moved inwards,
and  the tautological $\exists N \ \forall x\ N(x)$ can be removed.
We obtain the standard induction axiom:
$$
\forall X \ [ \forall x \ (X(x) \subset  x=0) \land 
\forall x\ ( X(s(x)) \subset X(x)) \supset \forall x \ X(x) ].
$$

\end{example}

\subsection{Iterated Inductive Definitions}

Recall from Section \ref{Sec:forms-of-induction} that an iterated
inductive definition constructs an set as the limit of a sequence of
constructive steps, each of which itself is a monotone induction.
Here, we formalise that intuition, and make a connection between this
new ``formalism'' and the representation of iterated inductive
definitions in ID-logic.

Let $(\D_1,\dots ,\D_n)$ be a finite sequence of positive definitions over
a vocabulary $\tau$ such that:
\begin{itemize}
\item all definitions define disjunct sets of relation symbols, i.e., $\tdi \cap \tdj = \emptyset$
for $i \not = j$;
\item if a relation symbol is defined in some $\D_i$, then it does not
  occur as an open symbol in $\D_j$, for any $j<i$.
\end{itemize}
We call such a sequence an {\em iterated inductive definititon} and we
interpret it as a simple, finite case of an iterated inductive
definition.

Let $\X$ be the set $\defp{\D_1} \cup \dots \cup \defp{\D_n}$, i.e.,
the collection of all symbols defined in at least one definition
$\D_i$, $1\leq i \leq n$, and let $\tauo$ be the vocabulary
$\tau\setminus\X$. Select an arbitrary $\tauo$-structure $\baseIo$.

We define ${\baseIo}^{(\D_1,\dots ,\D_n)}$ by induction on $i$:
${\baseIo}^{()} := \baseIo$ and for each $i$, $1\leq i \leq n$,
${\baseIo}^{(\D_1,\dots ,\D_i)} := 
({\baseIo}^{(\D_1,\dots ,\D_{i-1})})^{\D_i}$.  Note that by Theorem
\ref{theorem-Mon},  $\baseIo^{(\D_1,\dots ,\D_i)}$ is the least 
fixpoint of the positive definition $\D_i$ extending
$\baseIo^{(\D_1,\dots ,\D_{i-1})}$. The above definition models
precisely the process of iterated induction as explained in
Section 2. We say that the $\tau$-structure
${\baseIo}^{(\D_1,\dots ,\D_n)}$ is the {\em structure defined by the
iterated inductive definition $(\D_1,\dots ,\D_n)$ in $\baseIo$}.

Consider the iterated inductive definititon $(\D_1,\dots ,\D_n)$ and the
new definition $\D=\D_1\cup \ldots \cup \D_n$. It is obvious that
$\toD$ is equal to $\tauo$.

\begin{theorem}[iterated induction]
\label{TheoIteratedInduction}
Let $(\D_1,\dots ,\D_n)$ be an iterated inductive definititon over
vocabulary $\tau$.  Definition $\D := \D_1\cup \ldots \cup \D_n$ is a
total definition, and for all $\tau$-structures $I$ extending a
$\tauo$-structure $\valIopen$, the following are equivalent:

(a) $I$ is a model of $\D$;

(b) $I$ is the structure defined by $(\D_1,\dots ,\D_n)$ in $\valIopen$, i.e., 
$I=\valIopen^{(\D_1,\dots ,\D_n)}$;

(c) $I$ satisfies $\posind{\D_1} \land \ldots \land \posind{\D_n}$;

\end{theorem}
\noindent The theorem's significance is that it shows that the 
semantics of the logic correctly formalises this type of finite
iterated inductive definitions.

Define for each $i$, $0\leq i \leq n$,
$\tau^i := \tauo \cup \tdef{\D_1} \cup \ldots \cup \tdef{\D_{i}}$. It
is easy to see that $\tau^0 = \tauo$ and $\tau^n = \tau$. Also, it
holds that $\D_i$ is a definition over the vocabulary $\tau^i$, all
open symbols in $\D_i$ belong to $\tau^{i-1}$ and, for any
$\tauo$-structure $\baseIo$, ${\baseIo}^{(\D_1,\dots ,\D_i)}$ is a
$\tau^i$-structure.

To prove the theorem, we need the following modularity
lemma.
\begin{lemma}\label{lem-stratified} Let $(\D_1,\dots ,\D_n)$ 
be an iterated inductive definititon and let $\D := \D_1\cup \ldots
\cup \D_n$.
The definition $\D$ and the conjunction $\D_1\land
\dots \land \D_n$  of definitions are logically equivalent.
\end{lemma}
\begin{proof} 

\noindent Consider an arbitrary $\toD$-structure $\baseIo$ 
with domain $A$.  We will now prove that $\{\D_1,\dots,\D_n\}$ is a
total reduction partition of $\D$ in $\baseIo$. Then, by application
of Theorem \ref{theorem-modularity}, we obtain the lemma.

\noindent First, by Theorem \ref{theorem-Mon}, each $\D_i$ is 
total in $\baseIo$. Consequently, the partition $\{\D_1,\dots,\D_n\}$
is total in $\baseIo$.

\noindent Second, define the following partial order in $\At{\X}{A}$: 
for arbitrary domain atoms $\Pa$, $\Qb$, define $\Qb \dep \Pa$ iff $P$
is defined in $\D_i$ for some $i$, $1\leq i \leq n$, and $Q$  is defined in $\D_j$ for some $j$, $1\leq j \leq i$.

\noindent The relation $\dep$ is clearly a pre-well-order. 
By definition of $\dep$, it holds that if $P$ and $Q$ are not defined
in the same $\D_i$ and $\Qb \dep\Pa$, then $\Qb\sdep\Pa$.  We show
that for each $i$, $1\leq i\leq n$, $\dep$ is a reduction of $\D_i$ in
$\baseIo$.  

\noindent Let $P$ be a defined predicate of $\D_i$. For each domain
atom $\Pa$, for all $\tau$-structures $I, J\in
\vallat{\baseIo}{\tau}$, $I\islePa J$ holds iff $I|_{\tau^i} =
J|_{\tau^i}$. Since $\varphi_P$ contains only symbols of $\tau^i$, if
$(I,J)\islePa (I',J')$ then $\Tra{I}{J}
\models \varphi_P[\bar{a}] \mbox{ iff } \Tra{I'}{J'} \models
\varphi_P[\bar{a}]$.  

\noindent Since $\dep$ is a reduction of each $\D_i$, $1\leq i \leq n$,
Proposition \ref{PropReductionSubdefinition}(a) guarantees that $\dep$
is a reduction of $\D$. 

\noindent Combining the above results, we conclude that 
$\{\D_1,\dots,\D_n\}$ is a total reduction partition of $\D$ in
$\baseIo$.
\end{proof}

\begin{proof} of Theorem \ref{TheoIteratedInduction}.

\noindent Let $I$ be a $\tau$-structure extending $\baseIo$.
The following equivalences hold:\\

\noindent \begin{tabular}{lll}
$I$ is a model of $\D$ 
& iff $I$ is a model of $\D_1\land\dots\land\D_n$ 
	& (Lemma \ref{lem-stratified})\\ 
& iff $I$ satisfies $\posind{\D_1} \land \ldots \land \posind{\D_n}$ 
	& (Theorem \ref{theorem-Mon})\\ 
\end{tabular}\\

\noindent What remains to be shown is that $$I\models
\D_1\land\dots\land\D_n \mbox{ iff }I=\baseIo^{(\D_1,\dots ,\D_n)}.$$
Let $I$ be any $\tau$-structure extending $\baseIo$.  We show that
for each $i$, $0\leq i \leq n$, $I\models\D_1 \land \ldots \land
\D_{i} \mbox{ iff } I|_{\tau^i} =
\baseIo^{(\D_1,\ldots,\D_{i})}$. Then for the case $i=n$,  
we obtain that $I\models \D_1 \land \ldots \land \D_{n} $ iff
$I|_{\tau^n} = \baseIo^{(\D_1,\ldots,\D_n)}$ which, since $\tau^n =
\tau$, means that $I$ and $\baseIo^{(\D_1,\ldots,\D_n)}$ are
identical. \\

\noindent The proof is by  induction. In the base case ($i=0$), 
the property is trivially satisfied.  Assume that the property 
holds for $i-1$. We prove that the equivalence holds for $i$.

\noindent To prove one direction, assume that 
$I \models \D_1 \land \ldots \land \D_{i}$.  Since $I\models\D_i$,
it holds that $I = (I|_\toi)^{\D_i}$. Since all open symbols occurring
in $\D_i$ belong to $\tau^{i-1}$, it is easy to see that $I|_{\tau^i}
= (I|_{\tau^{i-1}})^{\D_i}$. Also,  $I \models \D_1 \land \ldots
\land \D_{i-1}$ and hence, by application of  the induction hypothesis, 
$I|_{\tau^{i-1}} = \baseIo^{(\D_1,\ldots,\D_{i-1})}$. We conclude that
$
I|_{\tau^i} = (I|_{\tau^{i-1}})^{\D_i} =
(\baseIo^{(\D_1,\ldots,\D_{i-1})})^{\D_i} =
\baseIo^{(\D_1,\ldots,\D_{i})}$.

\noindent For the other direction, assume that $I|_{\tau^i} =
\baseIo^{(\D_1,\ldots,\D_{i})}$.  Since
$\baseIo^{(\D_1,\ldots,\D_{i})} =
(\baseIo^{(\D_1,\ldots,\D_{i-1})})^{\D_i}$, $I$ extends
$\baseIo^{(\D_1,\ldots,\D_{i-1})}$ and $I\models \D_i$. By the
induction hypothesis, $I$ satisfies also $\D_1 \land \ldots \land
\D_{i-1}$. We obtain that $I\models \D_1 \land \ldots \land
\D_{i}$.
\end{proof}

\subsection{Definitions over Well-Founded Order.} 
We now present a formalisation of the informal concept of a definition
over a well-founded order (see section 2) in the framework of
ID-logic.  Let $\D$ be a definition over $\tau$ and $\baseI$ a
structure such that $\tauI{\baseI}\subseteq \toD$.

\begin{definition}[strict reduction relation]
A reduction relation $\dep$ of $\D$ in $\baseI$ is {\em strict} if it
is a strict well-founded partial order (i.e., an anti-symmetric,
transitive binary relation without infinite descending chains).
\end{definition}
\noindent Hence, a strict reduction $\dep$ has no cycles. If $\D$ allows a strict
reduction then there are no atoms that depend on themselves.

\begin{definition}[definition over a well-founded order] 
We say that $\D$ is a  definition over the (strict) well-founded order
$\dep$ in $\baseI$ if $\dep$ is a strict
reduction relation of $\D$ in $\baseI$.
\end{definition}

\ignore{

{\tt I'M NOT VERY HAPPY WITH THE TERMINOLOGY HERE.
STRICTLY SPEAKING, IT SHOULD BE "def over strict WF partial order",
BUT IT'S TOO LONG. 
I DON'T HAVE A BETTER SOLUTION THOUGH.
ALSO, i'm NOT SURE THAT THE SPECIAL NOTION OF STRICT REDUCTION
SHOULD BE INTRODUCED. WE COULD 
SIMPLY SAY THAT REDUCTION IS A STRICT WF PARTIAL ORDER.
WHAT DO YOU THINK? 
MARC: I SEE THE PROBLEM BUT I DONT KNOW HOW TO SOLVE IT }
}

\begin{theorem}[completion]\label{theorem-completion}

Suppose $\dep$ is a strict reduction relation of  $\D$ 
in $\baseI$.
The definition $\D$ is total in $\baseI$ and for any $\tau$-structure
$I$ extending $\baseI$, $I \models \D$ iff $I \models comp(\D)$.
\end{theorem}

\begin{proof}

Fix an arbitrary $\toD$-structure $\baseIo$ extending $\baseI$. We
will show that the equality $\lb{\baseIo}{\D} = \ub{\baseIo}{\D} =
\baseIo^\D$ holds, and moreover that for any $\tau$-structure $I$
extending $\baseIo$, $I\models comp(\D)$ iff
$I=\baseIo^\D$. Since $\baseIo$ is arbitrary, we will obtain the
proof of the theorem.

We start by showing that there is at most one pair $(I,J)$ in
$\vallat{\baseIo}{\tau}$ satisfying $\TPp_\D(I,J)=I$ and
$\TPp_\D(J,I)=J$, moreover if such a pair exists then $I=J$.

\noindent Suppose that there are two such pairs; i.e., there exist 
$I, J, I', J' \in \vallat{\baseIo}{\tau}$ such that $\TPp_\D(I,J)=I$,
$\TPp_\D(J,I)=J$, $\TPp_\D(I',J')=I'$ and $\TPp_\D(J',I')=J'$.  Let
$\Pa$ be a minimal atom such that $\Pa^I\neq \Pa^{I'}$ or $\Pa^J
\neq \Pa^{J'}$. Since $\dep$ is irreflexive, it holds that $I\islePa
I'$ and $J\islePa J'$.  Hence by Proposition \ref{PropReductionBasic},
$$\Pa^I=\Pa^{\TPp_\D(I,J)} = \Pa^{\TPp_\D(I',J')} = \Pa^{I'}$$
and $$\Pa^J=\Pa^{\TPp_\D(J,I)} = \Pa^{\TPp_\D(J',I')} = \Pa^{J'}.$$
We obtain a contradiction.

\noindent It follows that there can be at most one pair $(I,J)$ 
satisfying this condition. Moreover, if such a pair, say $(I,J)$,
exists then also the symmetric pair $(J,I)$  satisfies the
condition and consequently, $I$ and $J$ have to be identical.

Now, the proof of totality follows easily. The pair
$(\lb{\baseIo}{\D},\ub{\baseIo}{\D})$ is the maximal oscillating pair
of the stable operator. Every oscillating pair $(I,J)$ of the stable
operator satisfies $\TPp_\D(I,J)=I$ and $\TPp_\D(J,I)=J$.  By the
previous paragraph, it follows that $\lb{\baseIo}{\D} =
\ub{\baseIo}{\D} = \baseIo^\D$.  

We also just proved that $\baseIo^\D$ is the unique structure that
extends $\baseIo$ and satisfies the fixpoint equation
$\TPp_\D(I,I)=I$. We derive  for all $I$ extending $\baseIo$:
\begin{center}\begin{tabular}{ll}
$I = \baseIo^\D$ & iff  $I =\TPp_\D(I,I)$ \\
& iff $I = \G_{\D}(I)$ (Corollary \ref{Cor:approx})\\
& iff for each defined domain atom
$\Pa$, $\Pa^I = \varphi_P[\bar{a}]^{I}$ \\
& iff $I \models comp(\D)$.
\end{tabular}\end{center}
\end{proof}

We obtain the following corollary. 

\begin{corollary}\label{CorDefWF}
  Suppose a definition $\D$ over $\tau$ and a theory $T_{\rm o}$ over $\tauo
\subseteq \toD$ such that for any model $\baseI$ of $T_{\rm o}$, $\D$ is a
definition over some well-founded order $\dep$ in $\baseI$.
Then $T_{\rm o}\land \D$ and $T_{\rm o}\land comp(\D)$ are
  logically equivalent.
\end{corollary}

\begin{example} \label{ExNMEven3} Consider the definition $\D$ of 
Example \ref{ExNMEven}:
$$
\D_{even} := \defin{\forall x\ (E(x)\la x=0) \ ,\\ 
        \forall x\ (E(s(x)) \la \neg  E(x)) }.
$$
The transitive closure of the reduction $\{ (E[n],E[n+1]) \ | \ n\in
\natnr\}$ is a strict reduction of $\D_{even}$ in the natural numbers.
Consequently, in the context of the natural numbers, this definition
can be expressed in first-order logic, by $comp(\D_{even})$. 

Notice also that $\posind{\D_{even}}$ is inconsistent in the natural
numbers. Indeed, the sets $\{0,2,4,6,\dots \}$ and $\{0,1,3,5,\dots
\}$ are both minimal sets containing 0 and containing n+1 if n is not
contained. Consequently, there is no least such set.
\end{example}

\begin{example} In this example, we illustrate  
how an ID-theory can be transformed into an equivalent second-order
theory using the techniques that were developped in this paper.

Consider the  ID-theory  $T =  T_{\mathbb{N}} \cup \{\D \}$ where $T_{\mathbb{N}}$ was defined in Example \ref{Peano} and $\D$ in Example \ref{D3}:$$
\D := \defin{ \forall x\ (E(x)\la x=0),\\ 
        \forall x\ (E(s(x)) \la O(x)),\\ 
       \forall x \  (O(s(x))\la E(x)) } .
$$
In Example \ref{Peano3}, we showed that the ID-logic induction axiom in
$T_{\mathbb{N}}$ can be translated into the standard induction axiom
and that the unique model of this theory is the set of natural
numbers.

To translate $\D$ to classical logic, one can pick among several alternatives.
\begin{enumerate}
\item Since $\D$ is a positive definition, by Theorem \ref{theorem-Mon}, 
it can be translated into a second-order induction axiom. 
\item Alternatively, we observe that  $\D$  has a strict
reduction in the natural numbers. This is the transitive closure of
the relation $$\{ (E[n],O[n+1]), (O[n],E[n+1])\  | \ n\in
\natnr\}.$$ Now we can use Theorem \ref{theorem-completion} to
translate $\D$ to the first-order theory $comp(\D)$.

\item  In Example \ref{ExNMevenodd}, it was shown that $\D$ has a reduction partition in the natural numbers
$$
\D_1:= \defin{\forall x\ (E(x)\la x=0) ,\\ 
        \forall x\ (E(s(x)) \la O(x)) },\ \ \ 
\D_2:=  \defin{ \forall x \  (O(s(x))\la E(x) )}.
$$ 
Consequently, we can substitute $\D_1 \land \D_2$ for $\D$. Both
definitions are non-recursive and, by Theorem \ref{theorem-Hier}, they
are equivalent with $comp(\D_1) \land comp(\D_2)$. 
\end{enumerate}
After applying transformations (2) and (3), we obtain the same theory,
namely the first-order theory $comp(\D)$ augmented with the
second-order induction axiom (and Peano's disequality axioms).

\ignore{
 According to Theorem \ref{theorem-modularity},
$D3$ and $\D_4\land \D_5$ are equivalent in the structure of the
natural numbers. Consequently,  Corollary \ref{corollary-equivalent4} 
entails that 
\begin{equation} 
T_{\mathbb{N}} \land \D_3
\end{equation} 
is logically equivalent to  
$$ 
T_{\mathbb{N}} \land \D_4\land \D_5.
$$  
Further on, $\D_4$ and $\D_5$ are non-recursive definitions. By
application of Theorem \ref{theorem-Hier}, we  obtain  that the latter theory is
equivalent to: $$ 
T_{\mathbb{N}} \land comp(\D_4)\land comp(\D_5).
$$ 
This is a first-order theory augmented with the second-order induction axiom. }
\end{example}






\section{Conclusion}

\label{section-conclusions}
\vspace{-2mm}

Recently, we argued
\cite{Denecker:CL2000,Denecker/Bruynooghe/Marek:TOCL2001} that 
non-monotone forms of inductive definitions such as iterated inductive
definitions and definitions over well-orders, can play a unifying role
in logic, AI and knowledge representation, connecting remote areas
such as non-monotonic reasoning, logic programming, description
logics, deductive databases and fixpoint logics. In this paper, we
further substantiated this claim by defining a more general logic
integrating classical logic and monotone and non-monotone inductive
definitions and investigating its relations to first- and second-order
logic and studying its modularity properties.

The main technical theorems here are the Modularity theorem and the
theorems translating certain classes of ID-formulas into
classical logic formalisations. Problem-free composition is crucial
while axiomatizing a complex system. Because definitions in our logic
are non-monotone, composing or decomposing definitions is in general
not equivalence preserving. However, the conditions we have presented
allow one to separate problem-free (de)compositions from those causing
change in meaning. We have shown that the Modularity theorem is useful
also for analyzing complex definitions --- some properties of large
definitions are implied by properties of sub-definition.  The
Modularity theorem is also an important tool for simplifying logical
formulas with definitions by translating them into formulas of
classical logic.

\ignore{I DONT FULLY AGREE WITH THIS PARAGRAPH: The
lattice-theoretic techniques we develop in order to prove the
Modularity theorem are completely new, and have no counterpart
elsewhere.  We believe that with these techniques in hand, similar
modularity results will follow for other logics with similar fixed
point constructions (e.g. autoepistemic, default logic, etc.).}

In \cite{Denecker/Ternovska:KR2004}, we have applied our logic to what
has always been the most important test domain of knowledge
representation --- temporal reasoning. We presented an inductive
situation calculus, a formalisation of the situation calculus with
ramification as an inductive definition, defining fluents and
causality predicates by simultaneous induction in the well-ordered set
of situations. An important aspect of our formalisation is that
causation rules can be represented in a modular way by rules in an
inductive definition. We applied the Modularity theorem to demonstrate
its equivalence with a situation calculus axiomatization based on
completion and circumscription.  

\ignore{
In a forthcoming paper, we present a
general solution to the ramification problem for the situation
calculus within the ID-logic.

I"M NOT SURE WHICH PAPER YOU HAVE IN MIND. EUGENIA}

\nocite{Ternovskaia:1998a,Ternovskaia:1998b}

\bibliography{marclib,eugenia}

\bibliographystyle{acmtrans}

\begin{received}
\end{received}

\end{document}